\documentclass[runningheads]{llncs}

\usepackage{eccv}

\usepackage[table]{xcolor}
\usepackage{booktabs}
\usepackage{tabularx}

\usepackage[T1]{fontenc}
\usepackage{lmodern}

\newcommand{\tablefontsize}{\fontsize{8.5}{9.5}\selectfont}

\usepackage{tabularx}
\usepackage{multirow}
\usepackage[per-mode=symbol, group-digits=integer, group-minimum-digits = 3, detect-all=true, input-symbols={()}, retain-explicit-plus]{siunitx}

\usepackage[percent]{overpic} 
\usepackage{fp} 
\usepackage{graphicx}
\usepackage{tikz}
\usetikzlibrary{positioning}
\usetikzlibrary{shapes.geometric, shapes, arrows, arrows.meta, bending, calc,matrix,fit, positioning}
\usetikzlibrary{backgrounds}
\usetikzlibrary{spy}

\usepackage{xcolor}
\usepackage{array}
\usepackage{float}
\usepackage{pgfplots}
\usepackage{bbding}
\usepackage{dsfont}
\usepackage{fontawesome5}
\usepackage{calc}
\pgfplotsset{compat=1.17}
\usepackage[outline]{contour} 
\contourlength{0.02em}
\usepackage[table]{xcolor}
\usepackage{makecell}
\usepackage{adjustbox}
\usepackage{pifont}
\usepackage[scaled=0.97]{newtxtt}
\usepackage{utfsym}
\usepackage[super]{nth}
\newcommand{\fade}[1]{\textcolor{black!55}{#1}}

\newcommand*{\inparagraph}[1]{\smallskip\noindent\textbf{#1}\hspace{0.4em}}

\usepackage{eccvabbrv}

\usepackage{graphicx}
\usepackage{siunitx}
\usepackage{caption}
\usepackage{tikz,pgfplots}
\pgfplotsset{compat=1.18}

\usepackage[pagebackref,breaklinks,colorlinks,citecolor=eccvblue]{hyperref}

\usepackage{orcidlink}
\usepackage{tabularx}

\usepackage{booktabs}
\usepackage{multirow}
\usepackage{rotating}
\usepackage{adjustbox}
\usepackage{bbm}
\def\eg{\emph{e.g}\onedot} 

\def\ie{\emph{i.e}\onedot} 

\usepackage{mathtools}
\usepackage{textcomp}
\usepackage{gensymb}
\usepackage{bm}
\usepackage[dvipsnames]{xcolor}
\usepackage{enumitem}

\newcommand{\myparagraph}[1]{\vspace{4pt}\noindent\textbf{#1}}
\newcommand\ours{FoundYou}
\newcommand\sam{\textsc{SAM 2}}

\begin{document}

\title{FoundYou: A Unified Model for Personalized Segmentation and Retrieval} 

\titlerunning{FoundYou: A Unified Model for Personalized Segmentation and Retrieval}

\author{Gabriele Trivigno\textsuperscript{*}\orcidlink{0000-0002-5220-1838} \quad
Marcos Alfaro\textsuperscript{*}\inst{1,2}\orcidlink{0009-0008-8213-557X} \quad
Claudia Cuttano\textsuperscript{*}\inst{1}\orcidlink{0009-0004-9672-507X}
\\
Gabriele Berton\orcidlink{0000-0002-0128-0396} \quad
Luis Payá\inst{2,3}\orcidlink{0000-0002-3045-4316} \quad
Carlo Masone\inst{1}\orcidlink{0000-0002-1609-9338}}

\authorrunning{G. Trivigno et al.}

\institute{
$^1$ Politecnico di Torino \quad
$^2$ Miguel Hernández University of Elche \\
$^3$ Valencian Graduate School and Research Network of Artificial Intelligence \\
\url{https://ga1i13o.github.io/FoundYou/}
}

\maketitle
\begingroup
\renewcommand{\thefootnote}{\fnsymbol{footnote}}
\footnotetext[1]{Equal contribution.}
\endgroup
\begin{abstract}
Personalized segmentation and personalized retrieval both aim to identify the same physical object across different images. While the former localizes the object within a target image, the latter retrieves images where it appears. Despite this shared instance-level objective, the two tasks have largely evolved separately and are addressed with distinct solutions. In this work, we introduce \ours, a unified framework built on the observation that Segment Anything 2 (SAM 2), trained to preserve object identity across video frames, inherently captures instance-level cues. We leverage this property to match objects across independent images, enabling segmentation and retrieval to emerge as two outcomes of the same instance alignment process. This unified view unlocks new capabilities beyond traditional benchmarks, including few-shot personalized retrieval and promptable personalized segmentation with flexible prompts. Extensive experiments show consistent gains over unified and task-specific methods, including +18.4 mIoU on PerMIS and +17.8 mAP on ILIAS. Performance scales with additional references and remains robust to weaker prompts. Beyond personalization, FoundYou achieves state-of-the-art results on category-level retrieval benchmarks. Notably, our approach keeps the \sam{}-\textit{small} model entirely frozen and adds only 5.9 M trainable parameters, yielding a 52 M-parameter model that is over 75× faster and 20× smaller than the only prior unified solution.
\end{abstract}

\begin{figure}[t]
    \centering
    \includegraphics[width=\textwidth]{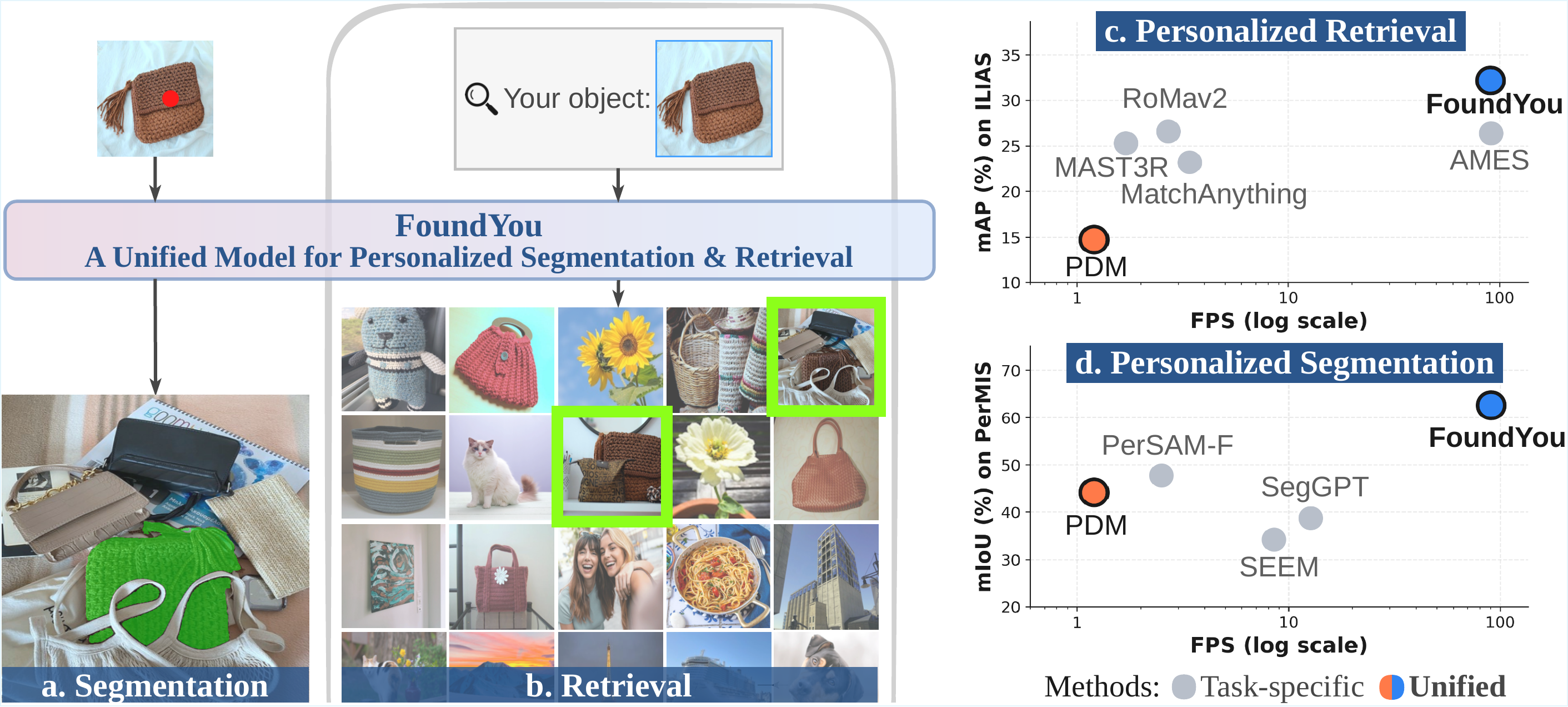}
    \caption{
\textbf{FoundYou: a Unified Model for Personalized Segmentation \& Retrieval.} Given a visual prompt of a user’s handbag, FoundYou segments the \textit{same physical instance} in a new image (\textbf{a}) and retrieves it from a gallery by ranking images via instance-level similarity (\textbf{b}). We compare performance vs.\ speed against the only prior unified solution, PDM~\cite{Samuel:2024:Waldo}, and the strongest task-specific methods: retrieval performance is shown in (\textbf{c}) as mAP vs.\ FPS, and segmentation in (\textbf{d}) as mIoU vs.\ FPS. FoundYou improves over PDM by +17.8 mAP and +18.4 mIoU while being over $75\times$ faster (note: x-axis in log scale). It outperforms also task-specific methods, while being the fastest in segmentation and matching the speed of the fastest retrieval competitor.
}
    \label{fig:teaser}
    \vspace{-0.3cm}
\end{figure}
   
\section{Introduction}

Suppose you provide a model with an example of \textit{your handbag} and ask it to segment the same handbag in completely different photos, in unrelated scenes that may contain other similar items (\cf \cref{fig:teaser}-a). Personalized segmentation \cite{Zhang:2023:PerSAM} formalizes this setting: given a visual reference identifying an object, the goal is to segment the same physical object in new images. Unlike few-shot segmentation \cite{Cuttano:2025:Sansa, Meng:2024:SEGiC, Zhu:2024:Unleashing}, the objective is not to recognize a category (\eg, \textit{any handbag}), but to preserve identity, \ie distinguish a particular instance from visually similar ones. Now consider a related objective: instead of segmenting your \textit{handbag} in a given image, you want to retrieve all the photos in a personal collection where it appears (\cf \cref{fig:teaser}-b). Personalized retrieval \cite{Samuel:2024:Waldo} captures this setting: given the same reference, the task becomes ranking images according to whether they depict that exact instance. 
Albeit producing different outputs, both tasks fundamentally require instance-level matching, from which a segmentation mask or image-level score is derived. This capability is central to applications, from robotics and embodied perception \cite{Sakaguchi:2024:Object, Kwon:2025:Embodied, Barsellotti:2024:Navigation} to cultural heritage digitization \cite{Sipiran:2021:Heritage, Ypsilantis:2021:Met}, personal media organization \cite{Kordopatis:2025:ILIAS, Alaluf:2024:MyVLM}, and creative workflows \cite{Nitzan:2022:MyStyle, Cohen:2022:Unicorn}.

Despite this shared objective, personalized segmentation and retrieval have evolved largely independently. Segmentation-focused approaches, such as PerSAM \cite{Zhang:2023:PerSAM}, solve the task via correspondence followed by mask prediction with SAM \cite{Kirillov:2023:SAM}, often relying on external models such as DINOv2 \cite{Liu:2023:Matcher, Oquab:2023:Dinov2} or diffusion encoders \cite{Tang:2025:Towards, Rombach:2022:SD}. Conversely, retrieval-specific methods learn instance-aware representations \cite{Suma:2024:Ames, Tan:2021:RRT, Cao:2020:Delg}, but do not provide dense localization. PDM \cite{Samuel:2024:Waldo} represents the closest attempt to bridge the two tasks. However, its reliance on Stable Diffusion leads to prohibitively slow inference (\cf \cref{fig:teaser}-c,d). Additionally, PDM relies on features extracted at different denoising timesteps for segmentation and retrieval, and still relies on prompting SAM for mask generation.

In this work, we take a different perspective by observing that Segment Anything 2 (\sam{}) \cite{Ravi:2024:Sam2}, trained to track objects across video frames, implicitly learns fine grained instance-aware representations. Building on this, we propose \ours{}, a unified framework for personalized segmentation and retrieval that repurposes \sam{} memory attention mechanism: instead of propagating identity across temporally adjacent frames, we use it to match a specific instance across \textit{independent} images. This shift, however, entails two fundamental challenges. First, because \sam{} is trained on natural videos, its representations inherit a temporal continuity bias: inter-frame variations are small and scene context remains stable \cite{Cuttano:2025:Sansa, Sun:2026:3AM}. In personalized segmentation and retrieval, however, matching occurs across unrelated images without positional or contextual continuity. To address this, we introduce lightweight feature adaptation layers trained for identity discrimination across distinct images, together with a self-distillation objective that preserves spatial structure of \sam{} representations, ensuring high-quality segmentation masks.
Second, \sam{} lacks a mechanism to assign an image-level similarity score. We therefore introduce a lightweight retrieval head that aggregates dense correspondences from memory attention into a global score, while reusing the original mask decoder for segmentation. Segmentation and retrieval thus emerge from the same instance alignment process.

Since instance matching is realized through memory attention over a set of reference features, \ours{} naturally supports multiple references at inference time. We therefore introduce the task of \textit{few-shot personalized retrieval}, reflecting practical cases where users provide several examples of the same object, for example, taking multiple photos of a \textit{handbag} to search for it more reliably. Additionally, \ours{} supports promptable personalized segmentation with lightweight annotations (\eg, points or boxes), removing the need for full masks.\\
Experiments show that \ours{} consistently outperforms both \emph{i)} the only prior unified solution and \emph{ii)} task-specific methods (\cf \cref{fig:teaser}-c,d). Compared to PDM \cite{Samuel:2024:Waldo}, we improve by +18.4 mIoU on PerMIS (segmentation) and +17.8 mAP on ILIAS (retrieval), while also surpassing \textit{specialized} approaches such as PerSAM-F \cite{Zhang:2023:PerSAM} (+14.8 mIoU) and AMES \cite{Suma:2024:Ames} (+6.1 mAP). Performance scales with additional references and remains robust to weaker annotations such as point prompts. Beyond personalization, \ours{} sets a new state-of-the-art on \textit{category-level} retrieval across $6$ domains. Notably, we keep \sam{} frozen and introduce only \textbf{\SI{5.9}{M} trainable parameters}, resulting in a compact \SI{52}{M}-parameter model 75× faster and 20× smaller than prior unified methods \cite{Samuel:2024:Waldo}.

\section{Related Work}

\myparagraph{Segment Anything 2 for downstream tasks.}  \sam{} \cite{Ravi:2024:Sam2} introduces a unified architecture for image and video segmentation, pretrained to segment objects and propagate masks across frames. Building on this, several works explore \sam{} for downstream tasks. A first line of research relies on external models such as DINOv2 \cite{Oquab:2023:Dinov2} or BEiT \cite{Wang:2023:Beit} to derive task-specific prompts, enabling few-shot \cite{Xu:2025:Unlocking, nie:2026:boosting}, open-vocabulary \cite{Xiao:2025:Openworldsam} and referring segmentation \cite{Rong:2025:Mpgsam}, also in remote sensing applications \cite{Rong:2025:Rs2sam}. Related to ours, a second line of work operates directly on \sam{}. SAMWISE \cite{Cuttano:2025:Samwise} adapts \sam{} features through motion- and language-aware conditioning, while GeoSAM2 \cite{Deng:2025:Geosam2} introduces geometric structure for 3D segmentation. SANSA \cite{Cuttano:2025:Sansa} and FS-SAM2 \cite{Forni:2025:fs-sam2} are close in spirit to our work. Both leverage \sam{} to guide segmentation in new images. However, their objective is to generalize matching to broader semantic categories. Conversely, our goal is to \textit{reinforce instance-level discrimination}. Rather than abstracting instance-specific cues, we aim to sharpen them, enabling reliable matching of the same physical object across independent images.

\myparagraph{Personalized Segmentation} aims to segment user-designated concepts in new images. Recent approaches address this task by first \textit{localizing} the instance in a target image via feature matching with a reference, and then \textit{segmenting} it by prompting SAM \cite{Kirillov:2023:SAM}. A key challenge lies in the representation used for matching. Early approaches such as PerSAM \cite{Zhang:2023:PerSAM} rely on SAM features, which lack sufficient semantic structure for reliable correspondence. Matcher \cite{Liu:2023:Matcher} addresses this by using DINOv2 \cite{Oquab:2023:Dinov2}, whose strong semantic representations improve matching but tend to collapse multiple similar instances of the same semantic category. To recover instance-level cues, PDM \cite{Samuel:2024:Waldo} explores diffusion-based features, while \cite{Tang:2025:Towards} further combines them with DINOv2. Despite improved localization, these diffusion-based pipelines remain computationally expensive, require careful timestep selection, and still rely on SAM for mask generation. In this work, we propose to leverage the instance-aware representations learned by SAM 2 \cite{Ravi:2024:Sam2} to jointly perform feature matching and segmentation within a unified framework.

\myparagraph{Instance Retrieval.}
Early retrieval works primarily focus on class-level discrimination~\cite{Sanakoyeu:2019:Divide, ElNouby:2021:Training, Musgrave:2020:Metric, Wang:2019:MultiSim}, while instance-level retrieval has been explored in domain-specific settings such as landmarks~\cite{Weyand:2020:GLDv2, Cao:2020:Delg}, place recognition~\cite{Arandjelovic:2018:Netvlad, Kim:2017:CRN, Berton:2023:JIST, Trivigno:2023:Divide}, and product or fashion collections~\cite{Ge:2019:Deepfashion2, Bai:2020:Products10k}. ILIAS~\cite{Kordopatis:2025:ILIAS} extends this setting to diverse object types, exposing challenges such as clutter and visually similar distractors, where global embeddings often lack sufficient instance-level discrimination~\cite{Cao:2020:Delg, Suma:2024:Ames}. To address this, prior works refine retrieval candidates using local descriptors and geometric verification~\cite{Trivigno:2024:Unreasonable, Suma:2024:Ames, Zhu:2023:R2Former}, including image-matching models such as~\cite{Sun:2021:LOFTR, Edstedt:2024:Roma, He:2025:MatchAnything}. While effective for matching covisible scene structures, improving performance in landmark-style retrieval  \cite{Sferrazza:2025:Match, Panek:2022:Meshloc}, these approaches prioritize geometric consistency and contextual alignment rather than object-centric identity. Specifically designed for re-ranking, AMES~\cite{Suma:2024:Ames} learns a transformer on top of frozen DINOv2 features, but remains grounded in representations that emphasize high-level semantics. In contrast, \sam{} \cite{Ravi:2024:Sam2}, pretrained for object tracking in videos, inherently learns instance-aware representations. We leverage this property with a lightweight retrieval head, achieving strong instance-level discrimination while retaining segmentation capability. Furthermore, we introduce a few-shot retrieval setting, where multiple reference examples are provided at test time, naturally supported by our framework without retraining.

\begin{figure*}[t]
        \centering
    \includegraphics[width=\linewidth]{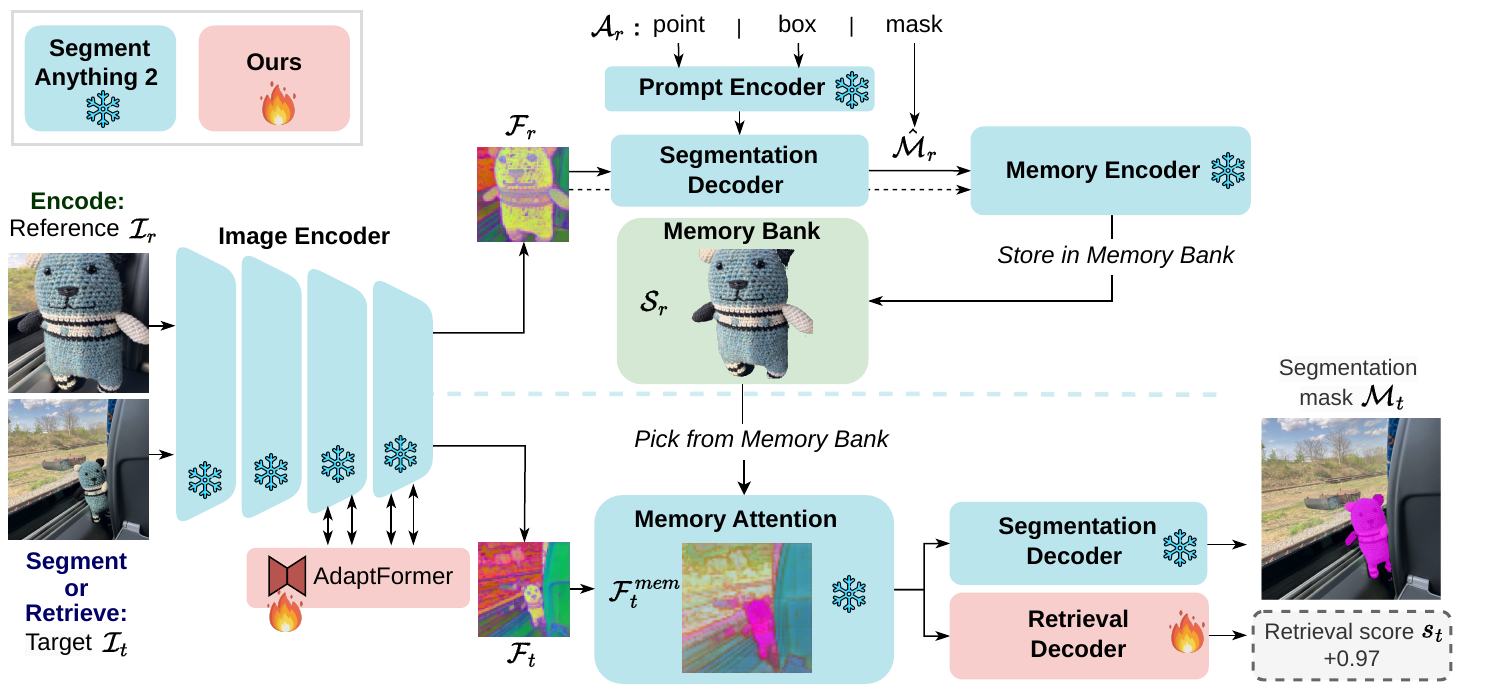}
    \caption{\textbf{Overview of FoundYou.} 
    We build on a frozen \sam{}, reinterpreting its memory mechanism to match instances across independent images rather than across temporally adjacent video frames. Given an object specified in the reference image $I_r$, the Memory Encoder stores its representation $\mathcal{S}_r$ in the Memory Bank. Target features $\mathcal{F}_t$ are matched with $\mathcal{S}_r$ through Memory Attention, producing dense memory features $\mathcal{F}_t^{mem}$, visualized via 3D PCA. From these, the frozen Segmentation Decoder produces a mask, while our lightweight Retrieval Decoder aggregates them into an image-level score, enabling personalized segmentation and retrieval within a single framework.
    }
    \label{fig:method}
\end{figure*} 

\section{Method}
\label{sec:method}
\myparagraph{Task Definition.}
A user provides a \emph{visual prompt} $V_r = (\mathcal{I}_r, \mathcal{A}_r)$ where $\mathcal{I}_r \in \mathbb{R}^{H \times W \times 3}$ is a reference image,
and $\mathcal{A}_r$ specifies the target instance within the image.
$\mathcal{A}_r$ can be a binary mask (\eg for personalized segmentation \cite{Zhang:2023:PerSAM, Samuel:2024:Waldo, Zhang:2024:GF-SAM}), a bounding box (\eg for personalized retrieval \cite{Kordopatis:2025:ILIAS, Radenovic:2018:ROP, Philbin:2008:Paris6k}), or a point. When the reference image contains a single object and no box is provided, the box coincides with the image boundaries.
Given a target image $\mathcal{I}_t \in \mathbb{R}^{H \times W \times 3}$, the goal of \textbf{personalized segmentation} is to predict a binary mask 
$\mathcal{M}_t \in \{0,1\}^{H \times W}$ corresponding to the \emph{same physical instance} indicated by $V_r$. 
We define the segmentation function as $\mathcal{M}_t = f_{\text{seg}}(V_r, \mathcal{I}_t)$. In \textbf{personalized retrieval}, given a gallery $\mathcal{G}=\{\mathcal{I}_i\}_{i=1}^N$, 
the objective is to rank database images according to \emph{instance-level relevance}, 
\ie, whether they depict the same physical object instance as specified by the visual prompt $V_r$. 
We define a similarity function $s_i = f_{\text{ret}}(V_r, \mathcal{I}_i) \in \mathbb{R}$.
which induces a ranking over $\mathcal{G}$ based on decreasing similarity. 

\myparagraph{Overview}.
An overview of our method is in \cref{fig:method}. In this work, we observe that \sam{}, unlike other vision foundation models, is \textit{explicitly} optimized to preserve object identity during mask propagation across video frames, implicitly learning instance-aware features. However, such representations are learned under a spatio-temporal continuity prior \cite{Cuttano:2025:Sansa, Sun:2026:3AM}, where identity is maintained within the same scene and under limited positional variation. Personalized segmentation and retrieval, by contrast, require discriminating and matching the same physical instance across \textit{independent} images, where scene context and spatial configuration may vary arbitrarily. Furthermore, \sam{} lacks a mechanism to assign image-level similarity scores between the reference instance and candidate images, which is necessary to rank results in personalized retrieval. To bridge this gap, we introduce a lightweight adaptation to remove the implicit tracking prior in \sam{} features, while preserving and enhancing instance awareness (\cref{sec:personalize}). The adapted representations establish dense correspondences between a visual prompt and a target image without relying on shared scene context. To produce outputs (\cref{sec:decoders}):
\emph{i)} for segmentation, we reuse the frozen \sam{} mask decoder to segment the matched instance;
\emph{ii)} for retrieval, we introduce a lightweight decoder that aggregates dense correspondences into an image-level score.
We train the adapters and retrieval head on pairs of positive and negative images, and add a self-distillation loss to preserve the spatial structure in the adapted features, required for accurate segmentation (\cref{sec:losses}). This formulation casts personalized segmentation and retrieval as two outcomes of the same instance-level alignment process, enabling both tasks within a single framework.

\subsection{Personalizing SAM 2}
\label{sec:personalize}

\sam{}~\cite{Ravi:2024:Sam2} is designed to preserve object identity across frames by encoding a reference instance into a \texttt{Memory Bank} and matching new frames through attention. 
We reinterpret this mechanism such that it can work \textit{across independent images}. The reference image $\mathcal{I}_r$ is processed by the \texttt{Image Encoder}, producing features $\mathcal{F}_r \in \mathbb{R}^{\frac{H}{16} \times \frac{W}{16} \times D}$. To construct a representation of the reference instance, the \texttt{Memory Encoder} fuses visual features with a reference mask $\hat{\mathcal{M}}_r$, which is either: \emph{i)} provided, in personalized segmentation, \ie, $\hat{\mathcal{M}_r} = \mathcal{A}_r$ or \emph{ii)} predicted from a bounding box in retrieval, \ie, $\hat{\mathcal{M}}_r = \texttt{Mask Decoder}(\mathcal{F}_r, \mathcal{A}_r)$. Formally, the \texttt{Memory Encoder} computes the memory representation $\mathcal{S}_r \in
\mathbb{R}^{\frac{H}{16} \times \frac{W}{16} \times D}$ as:

\begin{equation}
\label{eq:mem_enc}
\mathcal{S}_r =
\mathrm{conv}(\mathcal{F}_r) +
\mathrm{conv}(\hat{M}_r)
\end{equation}
This representation $\mathcal{S}_r$ is stored in the \texttt{Memory Bank} (see \cref{fig:method})\footnote{If multiple reference images are available, their memory representations are concatenated; with slight abuse of notation, we denote the result as $\mathcal{S}_r$. See Supplementary.}. Given a target image $\mathcal{I}_t$, the \texttt{Image Encoder} extract features $\mathcal{F}_t \in \mathbb{R}^{\frac{H}{16} \times \frac{W}{16} \times D}$. 
The target features are then matched to the representation $\mathcal{S}_r$ through \texttt{Memory Attention}:
\begin{equation}
\label{eq:mem_att}
\mathcal{F}_{t}^{\text{mem}} =
\mathrm{MHCA}\!\left(
Q(\mathcal{F}_t),
K(\mathcal{S}_r),
V(\mathcal{S}_r)
\right)
\end{equation}
where MHCA is a Multi-Head Cross Attention. \\
This produces memory-conditioned features $\mathcal{F}_{t}^{\text{mem}} \in
\mathbb{R}^{\frac{H}{16} \times \frac{W}{16} \times D}$, which encode dense correspondences between the target image and the reference instance.

However, \sam{} is trained to preserve instance identity across consecutive video frames, where appearance and position change smoothly over time. In our setting, identity must instead be matched across independent images, where no such continuity exists. To bridge this gap, we introduce a lightweight feature adaptation through AdaptFormer \cite{Chen:2022:Adaptformer} blocks inserted in the last two layers of the \texttt{Image Encoder}. Given down- and up- projection matrices $\mathbf{W}_{\text{down}} \in \mathbb{R}^{d \times \tilde{d}}, \mathbf{W}_{\text{up}} \in \mathbb{R}^{\tilde{d} \times d}$, an AdaptFormer block operates token-wise: 
\begin{equation}
A(x) = \sigma(x \cdot \mathbf{W}_{\text{down}}) \cdot \mathbf{W}_{\text{up}},
\end{equation}
where $\sigma$ is a ReLU and $\tilde{d} < d$ is the bottleneck dimensionality.
The adapted features are summed in a residual fashion in the backbone transformer blocks: 
\begin{equation}
\begin{aligned}
x_{\text{self}} &= \mathrm{MHSA}(x), \\
x' =  x_{\text{self}} \  +  \ & \mathrm{MLP}(x_{\text{self}}) + A(x_{\text{self}}),
\end{aligned}
\end{equation}
where MHSA is a Multi-Head Self Attention. \\
The backbone weights are frozen and we only train projections $\mathbf{W}_{down}$ and $\mathbf{W}_{up}$.

\subsection{Task-Specific Decoders}
\label{sec:decoders}

The resulting memory-conditioned features $\mathcal{F}_t^{\textrm{mem}}$ are then consumed by task-specific heads: \emph{i)} a Segmentation Decoder, for personalized segmentation or \emph{ii)} a Retrieval Decoder, for personalized retrieval.

\myparagraph{Segmentation Decoder.}
For segmentation, we directly reuse the frozen \sam{} mask decoder. 
A learned output token and the memory-conditioned features $\mathcal{F}_t^{\text{mem}}$ iteratively interact through bi-directional cross attention. 
The features are then progressively upsampled and fused with higher-resolution features from the \texttt{Image Encoder} through a top-down pathway. 
The final mask $\mathcal{M}_t \in \mathbb{R}^{H \times W}$ is obtained by computing a point-wise similarity between the refined features and the output token, followed by thresholding.

\myparagraph{Retrieval Decoder.}
Among the outputs of its decoder, \sam{} introduces an \textit{occlusion score}, predicted from the memory-conditioned features, to determine whether the tracked object is present in a frame. This mechanism is defined under temporal continuity, and does not transfer directly to independent images (\cf ablation in \cref{sec:ablation}). Nevertheless, because this score is computed directly from the memory-conditioned features, it implies that these representations already encode information about \textit{instance presence}. Building on this observation, we propose a Retrieval Decoder that aggregates the $\mathcal{F}_t^{\text{mem}}$ features into an image-level similarity score \textit{across independent images}. Specifically, we derive a simplified variant of the original decoder:
\emph{i)} we remove upsampling layers, FPN fusion and tokens required for mask prediction;
\emph{ii)} we introduce a learnable retrieval token;
\emph{iii)} we train this token to interact with the memory features through cross attention, aggregating spatial correspondences into a compact global representation.
Formally, let $\mathcal{F}_{t}^{\text{mem}}$ be reshaped into a sequence 
$\mathbf{F}^{(0)} \in \mathbb{R}^{L \times D}$ with $L=\frac{H}{16}\cdot\frac{W}{16}$, and let 
$\mathbf{z}^{(0)} \in \mathbb{R}^{1\times D}$ be a learnable retrieval token. We perform two rounds of bidirectional interaction between token and memory features:

\begin{equation}
\begin{aligned}
\vspace{-0.1cm}
\label{eq:retrieval_dec}
& \mathbf{F}^{(k)} = 
\mathrm{MHCA}(\mathbf{F}^{(k-1)}, \mathbf{z}^{(k-1)}),\\
\mathbf{z}^{(k)} =& \,  
\mathrm{MLP}
\big(
\mathrm{MHCA}(\mathbf{z}^{(k-1)}, \mathbf{F}^{(k)})
\big)\quad k \in \{1,2\}.
\vspace{-0.1cm}
\end{aligned}
\end{equation}

\noindent The final retrieval score is obtained as
\begin{equation}
\label{eq:score}
s_t =
\mathrm{sigmoid}
\big(
\mathrm{MLP}_{D\rightarrow 1}
(\mathbf{z}^{(2)})
\big),
\end{equation}
where $s \in [0,1]$ estimates the probability that the reference instance is present. Qualitative examples shown in \cref{fig:rerank_grid_rank_above} illustrate how scores reflect instance similarity, assigning high confidence to true matches and low confidence to outliers.

\subsection{Personalization Loss}
\label{sec:losses}
Training is formulated as a binary instance-matching problem. For a given reference image $\mathcal{I}_r$, we construct a candidate set $\mathcal{C}_r = \mathcal{P}_r \cup \mathcal{N}_r$, where $\mathcal{P}_r$ (positives) depicts the same physical instance and $\mathcal{N}_r$ (negatives) contains different instances. The reference is encoded into the \texttt{Memory Bank}, and each candidate image $I_j \in \mathcal{C}_r$ is processed independently producing a retrieval score $s_j \in [0,1]$. We assign binary labels  and optimize a binary cross-entropy loss:
\begin{equation}
\mathcal{L}_{\text{ret}} =
-\sum_{I_j \in \mathcal{C}_r}
\left[
y_j \log(s_j)
+
(1-y_j)\log(1-s_j)
\right].
\end{equation}

\noindent Since \sam{} already encodes strong instance-aware representations, randomly sampled negatives would result in trivial pairs (\eg, a coffee mug versus a car), which provide limited supervision and slow convergence \cite{Wu:2017:Sampling, Izquierdo:2024:Clique}. The primary challenge instead lies in distinguishing between visually similar objects within the same semantic category, as shown in \cref{fig:rerank_grid_rank_above}, \eg two \textit{different coffee mugs}.
A common strategy in retrieval is to mine hard negatives during training \cite{Arandjelovic:2018:Netvlad, Wu:2017:Sampling, Mereu:2022:SeqVLAD, Berton:2022:Benchmark}. However, online mining introduces significant computational overhead, as it must be repeated periodically while the model evolves \cite{Dutto:2024:federated_VPR, Warburg:2020:Mapillary, Revaud:2019:APLoss}.
Conversely, we leverage the semantic grouping properties of a frozen DINOv2, \ie $\phi(\cdot)$ to obtain semantically related examples (\eg, other visually similar mug for a reference coffee mug).
This decouples the mining procedure from the model itself, and can thus be carried out offline only once.
We compute global embeddings and select, for each reference $\mathcal{I}_r$, the most similar non-matching images:

\begin{equation}
\mathcal{N}_r =
\operatorname*{argmax\,top K}_{\mathcal{I}_k \notin \mathcal{P}_r}
\left(
\phi(\mathcal{I}_r)^\top \cdot \phi(\mathcal{I}_k)
\right).
\end{equation}
This results in candidate sets emphasizing intra-class instance discrimination.

\newcommand{\scorebadge}[1]{%
\begingroup
\setlength{\fboxsep}{1.2pt}%
\colorbox{black}{\textcolor{white}{\bfseries #1}}%
\endgroup
}

\newcommand{\imgW}{2.80cm}
\newcommand{\imgH}{1.9cm}

\newcommand{\colsepW}{1mm}
\newcommand{\legendW}{0.60cm}
\newcommand{\legendColW}{\legendW}
\newcommand{\gapQtoPone}{-1.0mm}   
\newcommand{\gapPonePtwo}{-1.0mm}  
\newcommand{\gapPtoNone}{0.0mm}    
\newcommand{\gapNoneNtwo}{-1.0mm}  
\newcommand{\gapNtoSeg}{0.0mm}     

\newcommand{\colLineGap}{.3mm}
\newcommand{\colLine}{%
  \hspace{\colLineGap}%
  {\color{black!18}\vrule width 0.35pt}%
  \hspace{\colLineGap}%
}

\colorlet{labelQueryBg}{blue!8}
\colorlet{labelPosBg}{green!8}
\colorlet{labelNegBg}{red!7}
\colorlet{labelSegBg}{orange!10}

\newcommand{\labelPadX}{0.8mm}
\newcommand{\labelPadY}{0.8mm}

\newcommand{\basetile}[3]{%
\begin{tikzpicture}[baseline=(box.center)]
  \node[minimum width=\imgW, minimum height=\imgW, inner sep=0pt] (box) {};
  \node at (box.center)
    {\includegraphics[width=\imgW,height=\imgW,keepaspectratio]{\detokenize{#3}}};

  \ifnum#1=1
    \draw[line width=2.4pt, green!70!black]
      (box.south west) rectangle (box.north east);
  \fi

  \if\relax\detokenize{#2}\relax\else
    \pgfmathsetmacro{\sig}{1/(1+exp(-(#2)))}%
    \node[anchor=north west,
          font=\bfseries\Large,
          fill=white, fill opacity=0.80,
          text opacity=1,
          rounded corners=1.2pt,
          draw=black!20, line width=0.3pt,
          inner xsep=3.2pt, inner ysep=1.2pt]
      at ($(box.north west)+(1.2pt,-1.2pt)$)
      {\pgfmathprintnumber[fixed,precision=2]{\sig}};
  \fi
\end{tikzpicture}%
}

\newcommand{\querytile}[1]{\basetile{0}{}{#1}}
\newcommand{\negtile}[2]{\basetile{0}{#1}{#2}}
\newcommand{\postileimg}[2]{\basetile{1}{#1}{#2}}

\newcommand{\masktile}[1]{%
\begin{tikzpicture}[baseline=(box.center)]
  \node[minimum width=\imgW, minimum height=\imgW, inner sep=0pt] (box) {};
  \node at (box.center)
    {\includegraphics[width=\imgW,height=\imgW]{\detokenize{#1}}};
\end{tikzpicture}%
}

\newcommand{\querystack}[1]{%
\begin{tabular}{@{}c@{}}
\querytile{#1}
\end{tabular}%
}
\newcommand{\posstack}[2]{%
\begin{tabular}{@{}c@{}}
\postileimg{#1}{#2}
\end{tabular}%
}
\newcommand{\negstack}[2]{%
\begin{tabular}{@{}c@{}}
\negtile{#1}{#2}
\end{tabular}%
}
\newcommand{\segstack}[1]{%
\begin{tabular}{@{}c@{}}
\masktile{#1}
\end{tabular}%
}

\newcommand{\rotlegendcellHC}[3]{%
  \parbox[c][#1][c]{\legendW}{%
    \centering
    \rotatebox[origin=c]{90}{%
      \colorbox{#2}{%
        \hspace{\labelPadX}%
        \raisebox{0pt}[\dimexpr\labelPadY+\height\relax][\dimexpr\labelPadY+\depth\relax]{%
          \shortstack[c]{#3}%
        }%
        \hspace{\labelPadX}%
      }%
    }%
  }%
}
\newcommand{\legendoverlayC}[3]{%
  \makebox[0pt][r]{\rotlegendcellHC{#1}{#2}{#3}}%
}

\newlength{\legendHQuery}
\newlength{\legendHPos}
\newlength{\legendHNeg}
\newlength{\legendHSeg}
\newlength{\legendVShift}

\newcommand{\setupLegendHeights}{%
  \setlength{\legendVShift}{9.8cm}
  \setlength{\legendHQuery}{\imgW}
  \setlength{\legendHPos}{2\imgW}
  \setlength{\legendHNeg}{2\imgW}
  \setlength{\legendHSeg}{\imgW}
}

\newcommand{\inlinebadge}[1]{%
\begingroup
\setlength{\fboxsep}{2pt}%
\fcolorbox{black!30}{white}{\bfseries #1}%
\endgroup
}

\begin{figure*}[t]
\raggedleft
\setlength{\tabcolsep}{0pt}
\renewcommand{\arraystretch}{0.95}
\setlength{\extrarowheight}{0pt}
\setupLegendHeights

\resizebox{.97\textwidth}{!}{%
\begin{tabular}{@{}p{\legendColW}@{\hspace{\colsepW}}c@{\colLine}c@{\colLine}c@{\colLine}c@{\colLine}c@{\colLine}c@{\colLine}c@{}}

\legendoverlayC{\legendHQuery}{labelQueryBg}{\bfseries\Large Reference \\ \bfseries\Large Image } &
\querystack{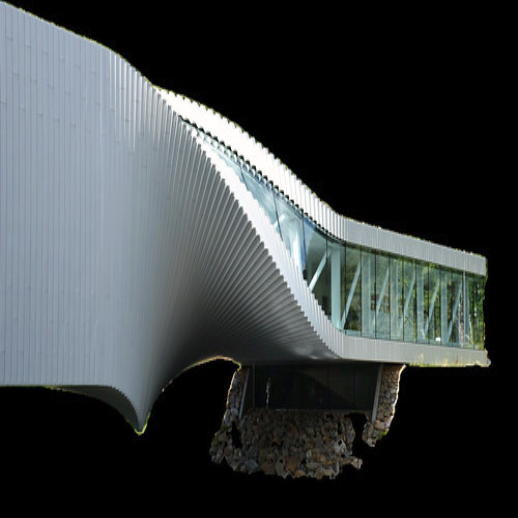} &
\querystack{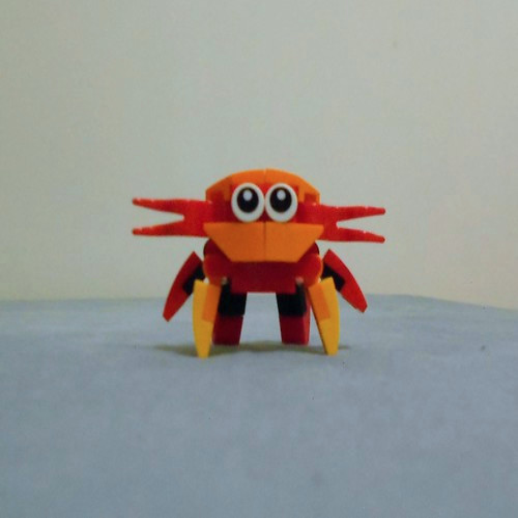} &
\querystack{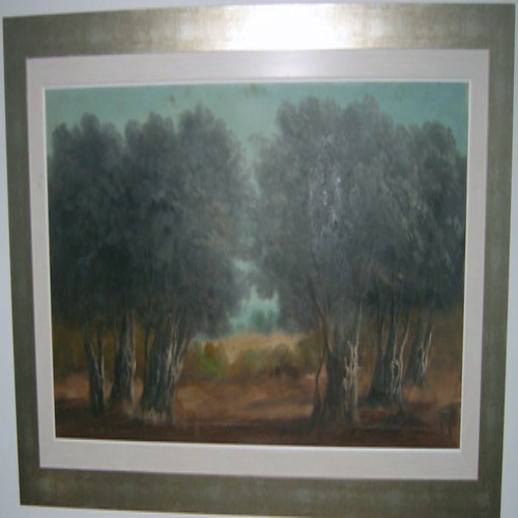} &
\querystack{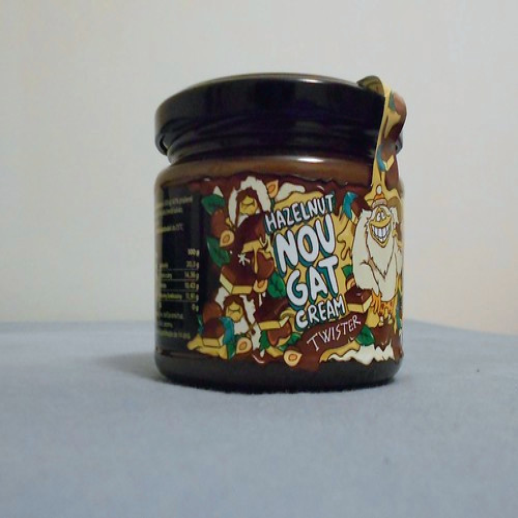} &
\querystack{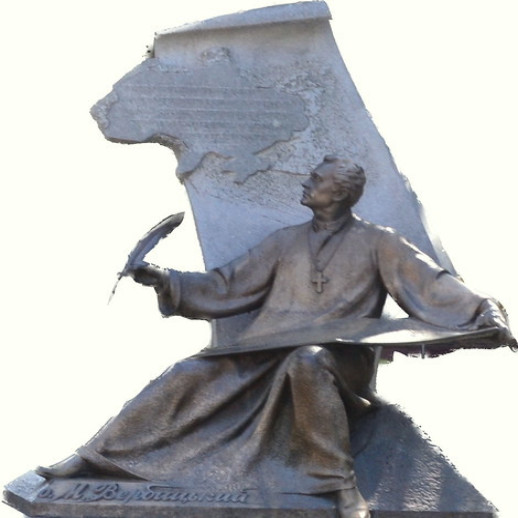} &
\querystack{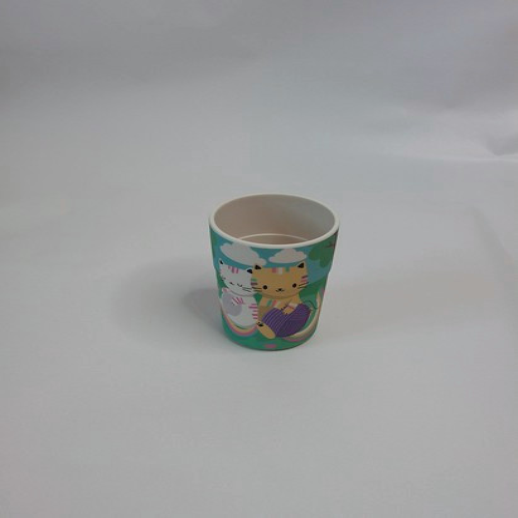} &
\querystack{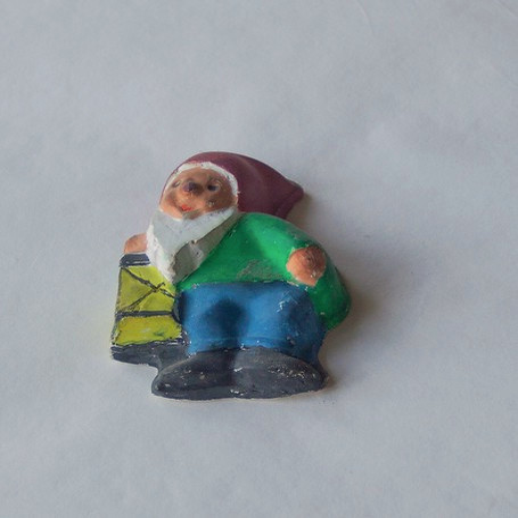} \\
[\gapQtoPone]

\multirow{2}{*}[\legendVShift]{%
\legendoverlayC{\legendHPos}{labelPosBg}{\bfseries\Large Top-2 \\ \bfseries\Large retrieved }%
} &
\posstack{2.49}{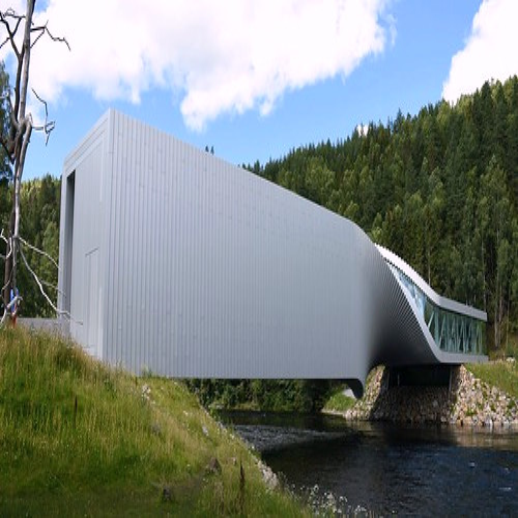} &
\posstack{1.82}{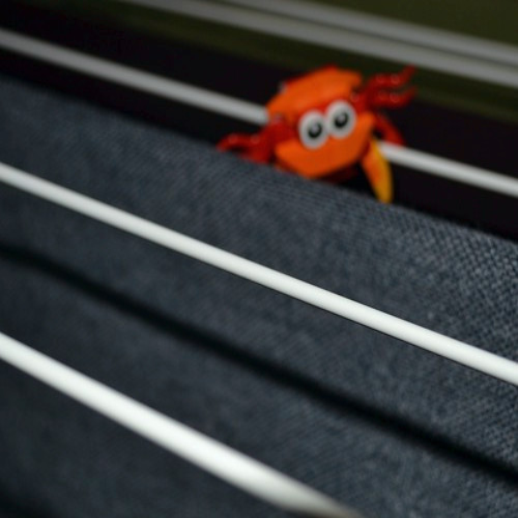} &
\posstack{3.42}{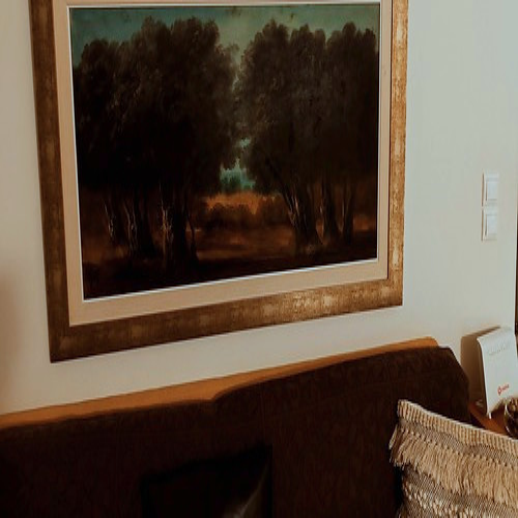} &
\posstack{0.73}{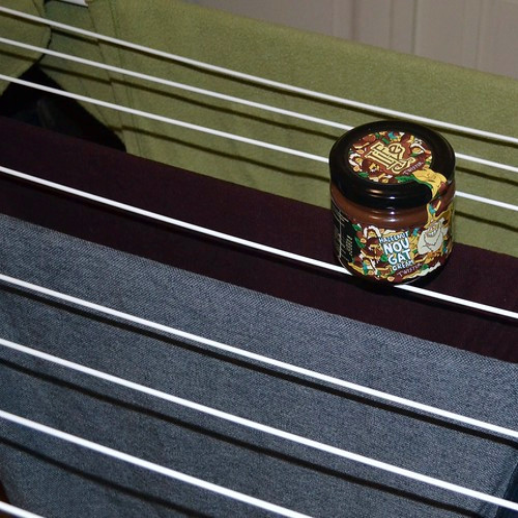} &
\posstack{3.89}{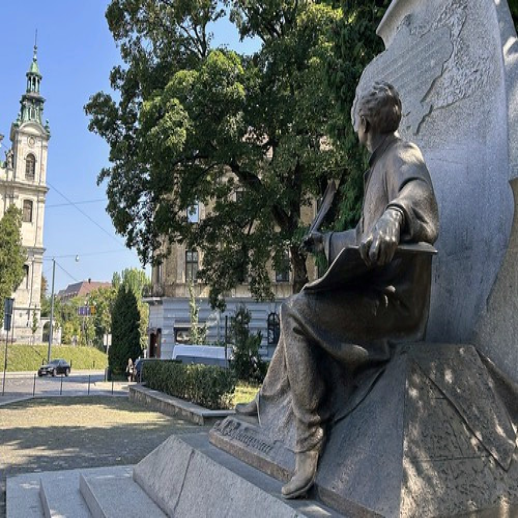} &
\posstack{1.86}{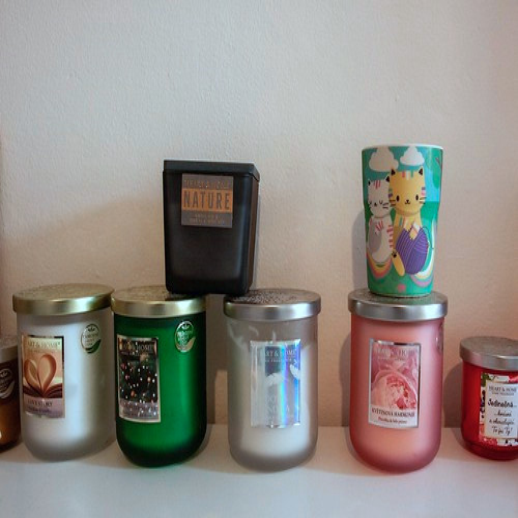} &
\posstack{2.77}{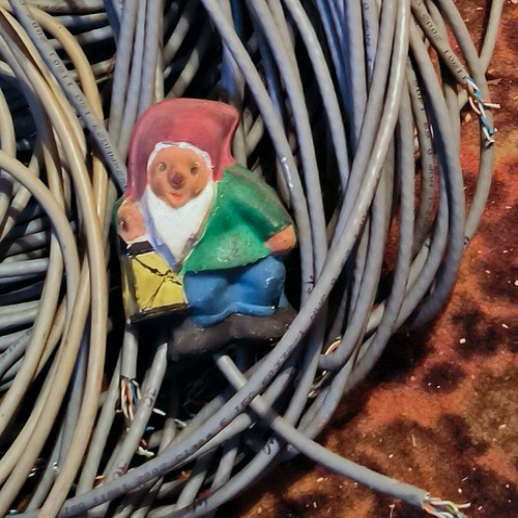} \\
[\gapPonePtwo]

&
\posstack{1.5}{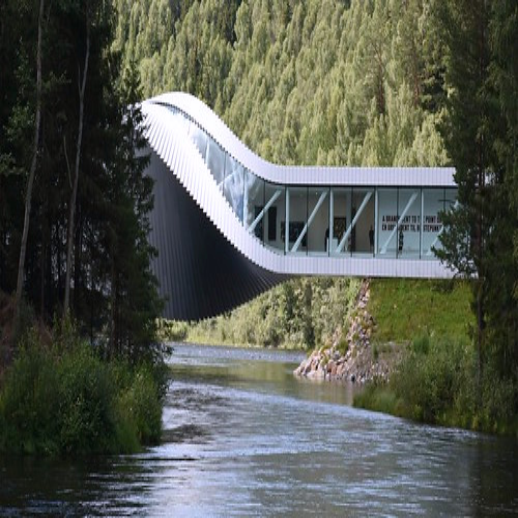} &
\posstack{0.84}{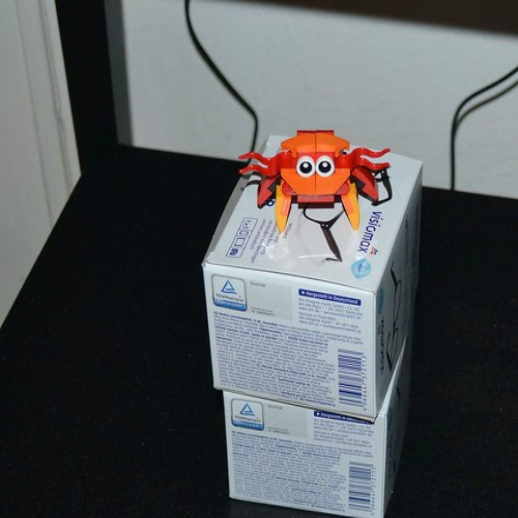} &
\posstack{2.41}{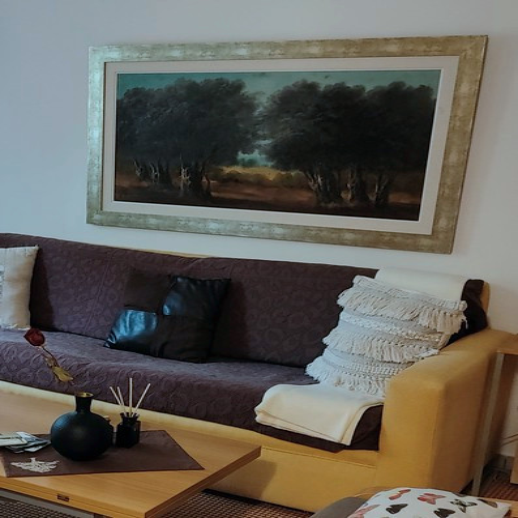} &
\posstack{0.33}{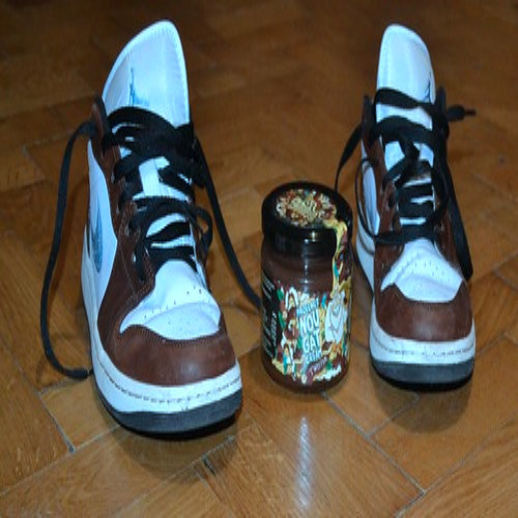} &
\posstack{3.62}{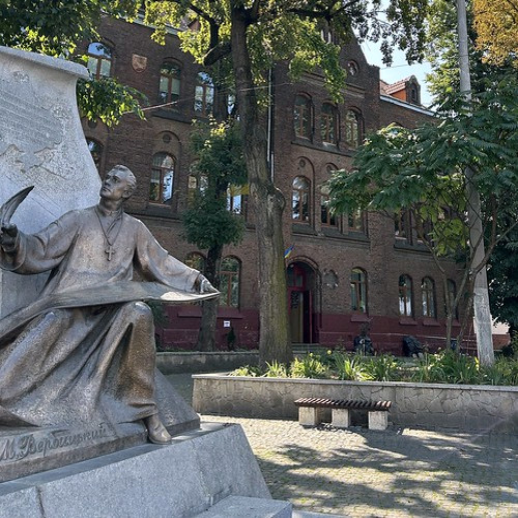} &
\posstack{1.18}{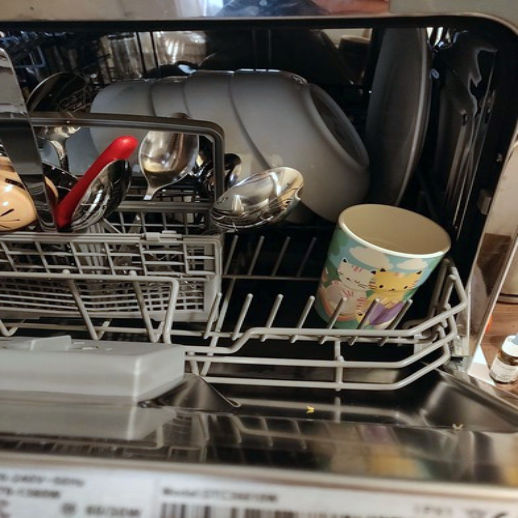} &
\posstack{2.61}{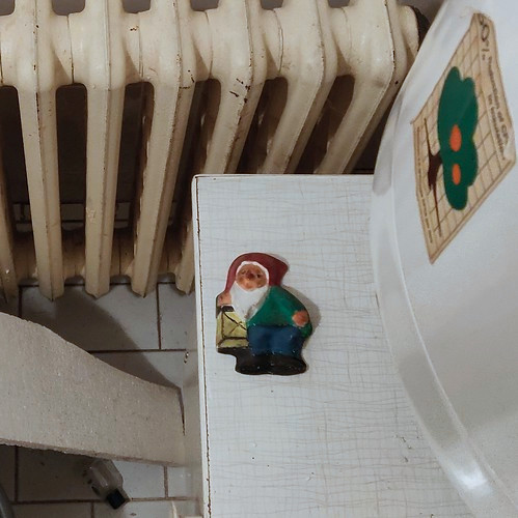} \\
[\gapPtoNone]

\multirow{2}{*}[\legendVShift]{%
\legendoverlayC{\legendHNeg}{labelNegBg}{\bfseries\Large Hard \\ \bfseries\Large Negatives}%
} &
\negstack{-0.49}{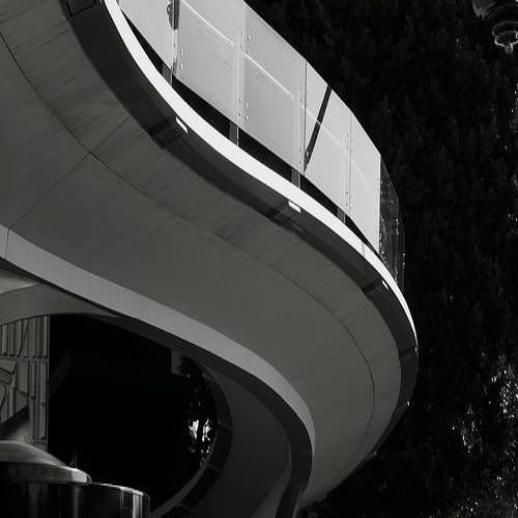} &
\negstack{-0.61}{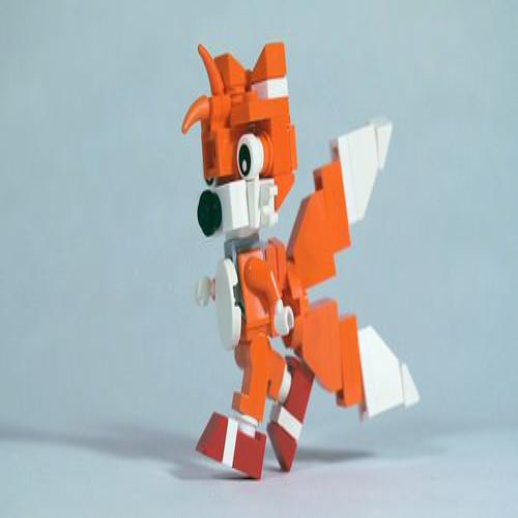} &
\negstack{-0.09}{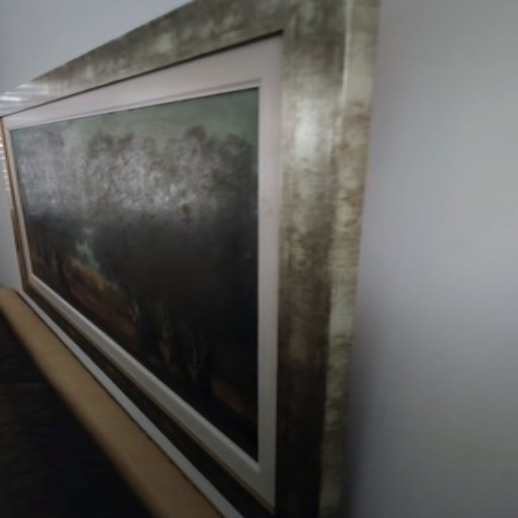} &
\negstack{-0.44}{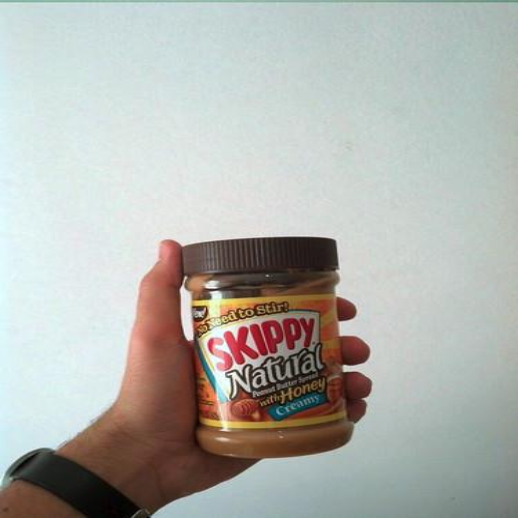} &
\negstack{-0.84}{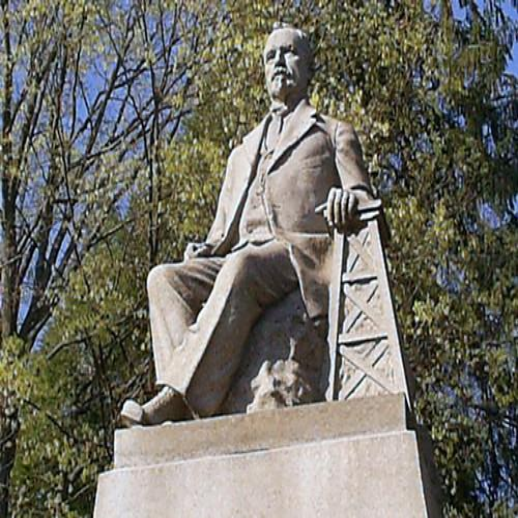} &
\negstack{-2.59}{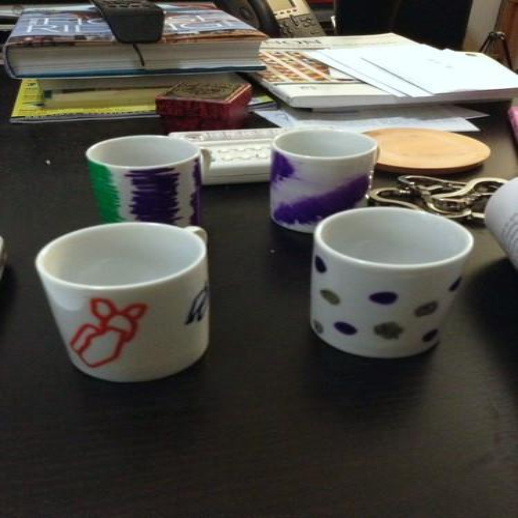} &
\negstack{-0.95}{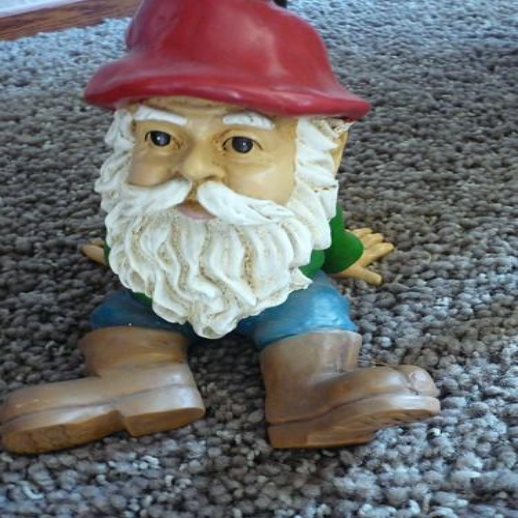} \\
[\gapNoneNtwo]

&
\negstack{-0.83}{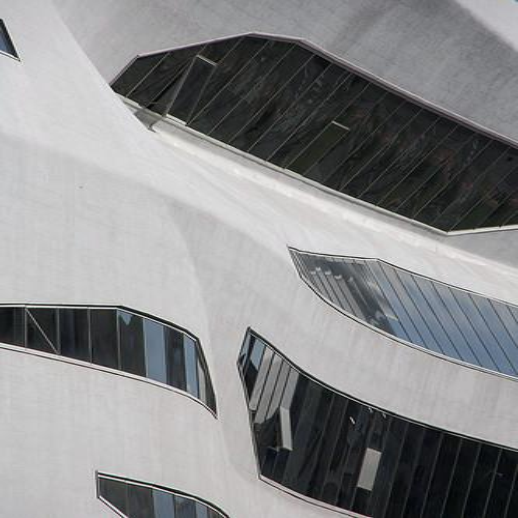} &
\negstack{-0.71}{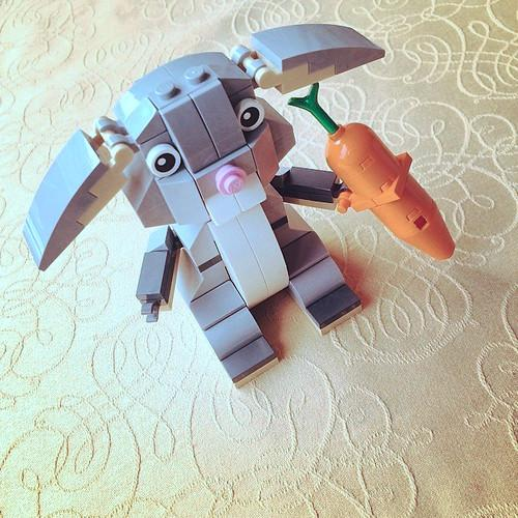} &
\negstack{-0.94}{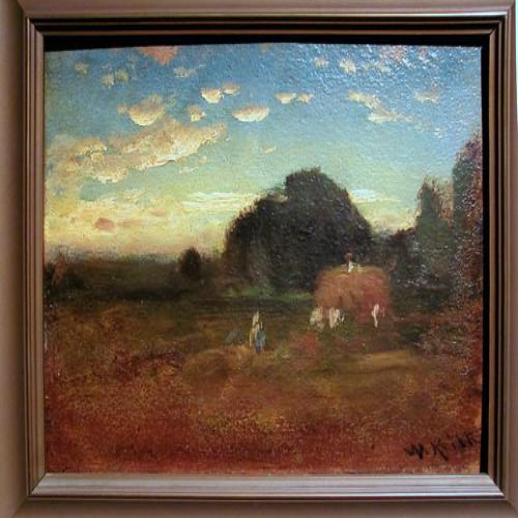} &
\negstack{-0.69}{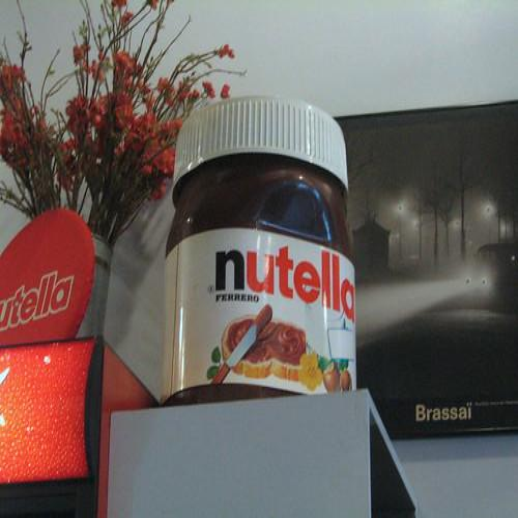} &
\negstack{-0.85}{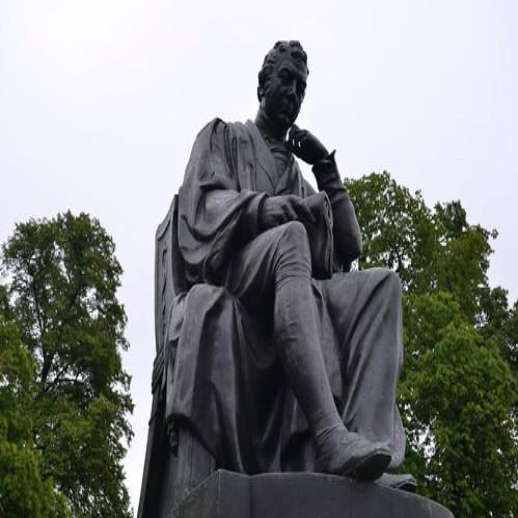} &
\negstack{-2.60}{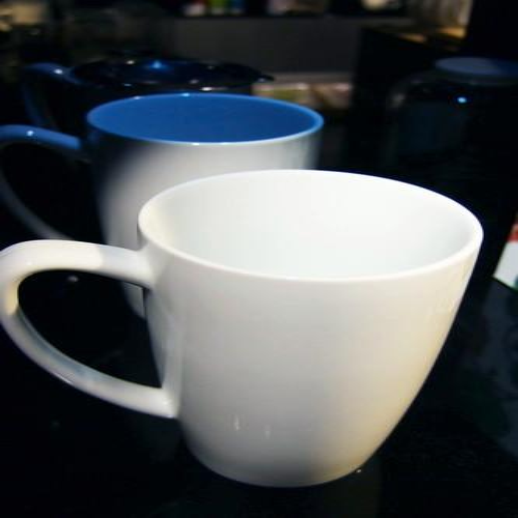} &
\negstack{-1.42}{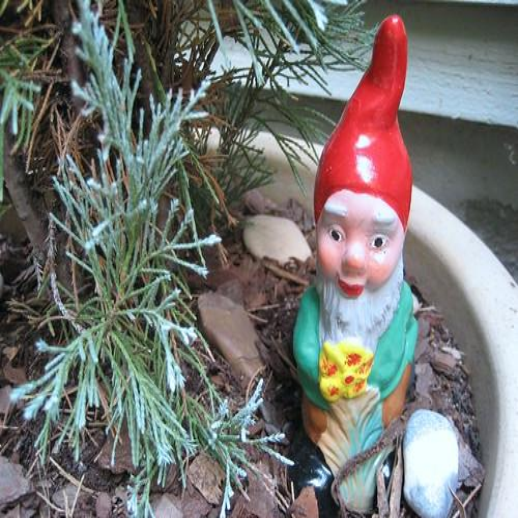} \\
[\gapNtoSeg]

\legendoverlayC{\legendHSeg}{labelSegBg}{\bfseries\Large Top-1\\ \bfseries\Large segmented} &
\segstack{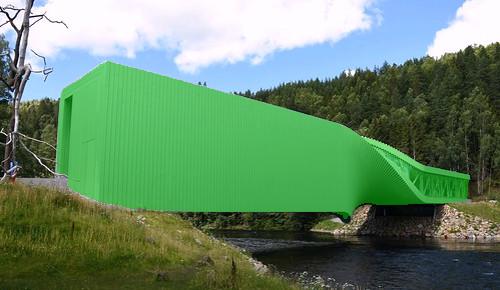} &
\segstack{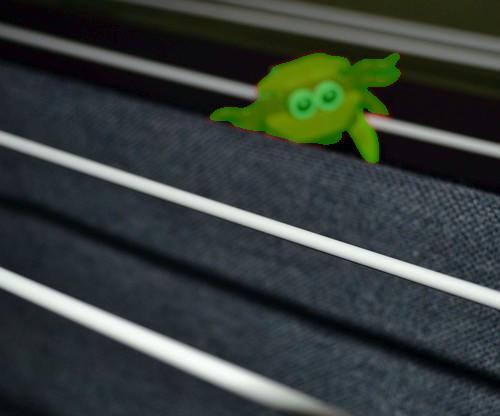} &
\segstack{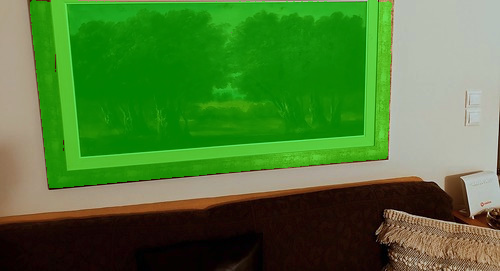} &
\segstack{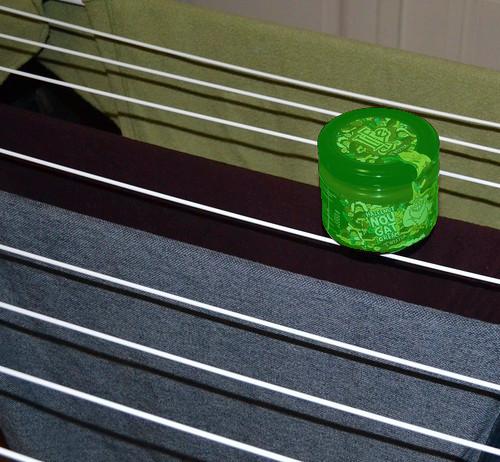} &
\segstack{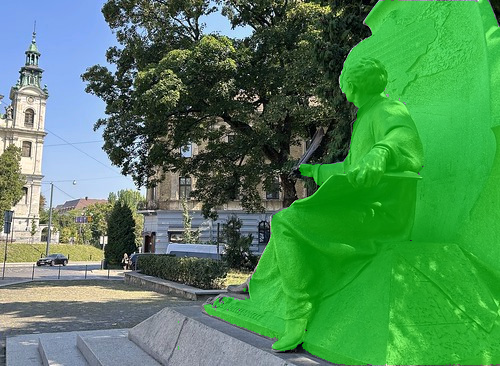} &
\segstack{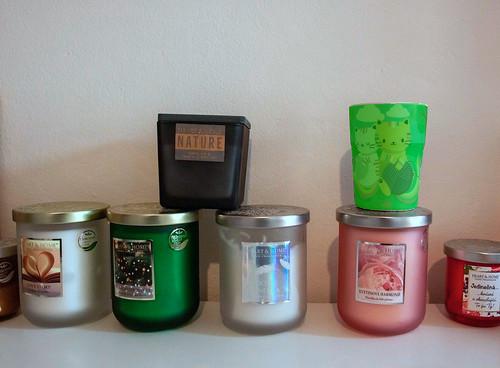} &
\segstack{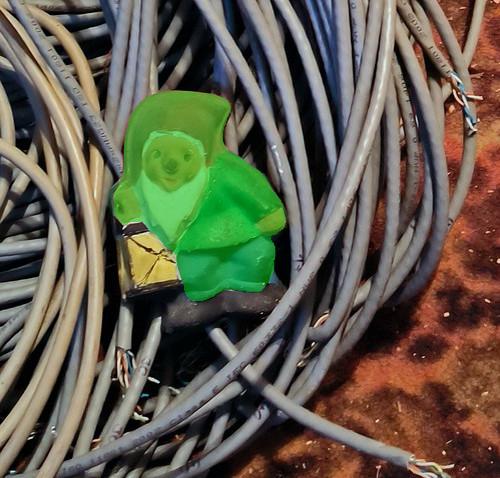} \\

\end{tabular}%
}
\vspace{-0.1cm}
\caption{\textbf{Personalized retrieval and segmentation examples.}
For each reference image (row 1), we show top-2 retrieved candidates (rows 2–3), along with hard negatives examples (rows 4–5) of visually similar but non-matching objects. The retrieval score \inlinebadge{$s_t$} is reported for each candidate. The last row shows personalized segmentation results on the top-1 retrieved image, using the same reference (row 1) as visual prompt.}
\label{fig:rerank_grid_rank_above}
\vspace{-0.3cm}
\end{figure*}

\myparagraph{Mask-based self-distillation.}
Our adaptation relaxes the spatio-temporal continuity prior implicitly encoded in \sam{}, enabling instance matching across independent images where scene context varies. However, training the adapter with an image-level objective degrades the dense representations required for accurate mask prediction (\cf ablations in \cref{sec:ablation}). To preserve this spatial structure, we introduce a self-distillation objective on that regularizes the adapted model to maintain the segmentation capabilities of the frozen backbone. During training, we compute teacher (\ie SAM 2) and student (\ie FoundYou) mask logits on a given image using \emph{i)} the frozen \sam{} model and \emph{ii)} our adapted model, respectively.
We denote with $P^{\mathrm{T}} \in \mathbb{R}^{H\times W}$ and 
$P^{\mathrm{S}} \in \mathbb{R}^{H\times W}$ the corresponding foreground probability maps obtained by applying a sigmoid to the predicted mask logits. We minimize a pixel-wise Kullback–Leibler divergence:
\begin{equation}
\mathcal{L}_{\text{dist}}
=
\frac{1}{HW}
\sum_{h,w}
\mathrm{KL}
\!\left(
P_r^{\mathrm{T}}(h,w)
\;\|\;
P_r^{\mathrm{S}}(h,w)
\right).
\end{equation}
The final training objective combines the personalization and distillation losses:
\begin{equation}
\mathcal{L}
=
\mathcal{L}_{\text{ret}}
+
\lambda_{\text{dist}} \mathcal{L}_{\text{dist}} .
\end{equation}

\section{Experiments}
We evaluate \ours{} against unified approaches and task-specific baselines on:\\
\emph{(1) \textbf{Personalized Segmentation.}} Given a visual prompt and a reference mask of an instance, the model is tasked with \textit{segmenting} the same object in a target image. We further introduce a \textit{promptable} setting in which we relax the mask assumption, and rely on box or point prompts to refer the object of interest. \\
\emph{(2) \textbf{Personalized Retrieval.}} Given a visual prompt depicting an instance or a landmark, and a gallery of candidates, the model is tasked with \textit{retrieving}, \ie, ranking the candidates according to the probability that they contain the object of interest. Furthermore, we propose a \textit{few-shot} retrieval setting, in which multiple examples are provided to the model.

\myparagraph{Datasets.}
For \emph{(1) \textbf{Personalized Segmentation}}, we evaluate on \textit{PerSeg}~\cite{Zhang:2023:PerSAM} and \textit{PerMIS}~\cite{Samuel:2024:Waldo}. \textit{PerSeg} contains 40 personal objects captured under substantial variations in pose, scale, and background. \textit{PerMIS} extends this setting with over 432 query–target pairs, captured in more challenging crowded scenes.
For \emph{(2) \textbf{Personalized Retrieval}}, we evaluate on \textit{PerMIR}~\cite{Samuel:2024:Waldo} and \textit{ILIAS}~\cite{Kordopatis:2025:ILIAS}. \textit{PerMIR} provides a baseline benchmark, but remains limited in scale, with a gallery of 432 images and 216 queries. \textit{ILIAS} introduces a significantly more challenging setting, comprising 1,232 instances with curated positives, alongside 100M distractors from YFCC100M~\cite{Thomee:2016:YFCC100M}, spanning diverse domains (\eg, landmarks, fashion, products, artworks). To assess generalization beyond strict instance matching, we evaluate on class-level retrieval benchmarks, \textit{Cars196}, \textit{iNaturalist}, \textit{RP2K}, \textit{Stanford Online Products}, \textit{Food2K}, \textit{MET}, and the landmark datasets \textit{Google Landmarks v2}, \textit{Revisited Paris}, and \textit{Oxford}. Such datasets provide a complementary setting, where the objective shifts from retrieving the \textit{same physical instance} to retrieving the \textit{same fine-grained category}. More details in the Supp.

\myparagraph{Implementation Details.}
We use the Small version of \sam{} (\SI{46}{M} parameters), and keep it entirely frozen. We add and train only  \SI{5.9}{M} parameters, \ie the adapter layers and retrieval head. The model is trained with Adam (LR $1\times10^{-4}$, batch size 16) for 150k iterations, using 4 positive and 8 negative examples per query. $\lambda_{dist}$ is set to $0.5$. 
We train FoundYou on UnED~\cite{Ypsilantis:2023:UNED}, following~\cite{Ypsilantis:2023:UNED, Kordopatis:2025:ILIAS}.

\myparagraph{Evaluation protocol.} For segmentation, we follow prior work~\cite{Zhang:2023:PerSAM, Samuel:2024:Waldo} and report mIoU (\%) and bIoU (\%).
On PerMIR, we perform retrieval over the entire gallery of 432 images. For ILIAS, which contains a gallery of 100M images, we follow~\cite{Kordopatis:2025:ILIAS}: the top-1k most similar candidates are first retrieved from the gallery using SigLIP global similarity~\cite{Zhai:2023:SigLIP}, and then re-ranked by each method. Following~\cite{Suma:2024:Ames, Kordopatis:2025:ILIAS, Ypsilantis:2023:UNED, Tan:2021:RRT, Samuel:2024:Waldo}, we evaluate retrieval performance using mAP (\%).

\subsection{Unified Personalized Segmentation and Retrieval.}

\myparagraph{Baselines.} In \cref{tab:seg_retrieval} we compare against state-of-the-art methods for personalized segmentation and retrieval. For segmentation, we compare with generalist models such as SEEM \cite{Zou:2023:SEEM}, SegGPT \cite{Wang:2023:SegGPT}, GF-SAM \cite{Zhang:2024:GF-SAM} and personalized specialist PerSAM \cite{Zhang:2023:PerSAM}, of which we also consider fine-tuned version (PerSAM-F).
Concerning retrieval, the most relevant comparison are image matching methods \cite{Sun:2021:LOFTR}, which can be used to verify and re-rank retrieved candidates \cite{Berton:2024:EarthMatch, Barbarani:2023:Local, Tan:2021:RRT, Zhu:2023:R2Former}. Among them we consider recent methods such as RoMav2 \cite{Edstedt:2025:Romav2}, MatchAnything \cite{He:2025:MatchAnything}, LightGlue \cite{Lindenberger:2023:Lightglue} and MASt3R \cite{Leroy:2024:Mast3r}. We use the number of inliers as similarity score, following standard practice \cite{Barbarani:2023:Local, Sferrazza:2025:Match}. We also evaluate re-ranking specialist methods such as RRT \cite{Tan:2021:RRT} and the recent state-of-the-art AMES \cite{Suma:2024:Ames}.
Moreover, we compare with the only prior work that supports both tasks, PDM \cite{Samuel:2024:Waldo}, which relies on StableDiffusion \cite{Rombach:2022:SD} features, and SAM \cite{Kirillov:2023:SAM} for producing masks.

\newcommand{\na}{\textcolor{black!40}{\textit{n.a.}}}

\newcommand{\pubsize}{\fontsize{6}{7}\selectfont}
\newcommand{\pub}[1]{%
  \textcolor{black!60}{{\pubsize\,#1}}%
}
\newcommand{\zwheader}[1]{\makebox[0pt][c]{#1}}
\newcommand{\smheader}[1]{\makebox[20pt][c]{#1}}
\newcommand{\mdheader}[1]{\makebox[104pt][c]{#1}}

\begin{table*}[t]
\caption{\textbf{Personalized segmentation and retrieval} (mIoU, bIoU, and mAP in \%, $\uparrow$). For image-matching methods, we use the open-source toolbox from \cite{Berton:2024:EarthMatch}. For PDM, a training-free method, we use the official code. On ILIAS, we use each method to re-rank the same top-1k candidates retrieved by SigLIP; SigLIP alone achieves 19.6 mAP.  
\ours{} achieves the best performance across benchmarks while using a compact \SI{52}{M}-parameter model running at 90 FPS on an RTX 4090. It is $75\times$ faster than PDM, faster than segmentation methods, and comparable to the fastest retrieval model.}

\vspace{-0.2cm}
\centering
\tablefontsize

\setlength{\tabcolsep}{2pt}
\begin{tabularx}{\linewidth}{@{}X|cc|cc|c|c|c@{\hspace{1pt}}c@{}}
\toprule
\multicolumn{1}{c}{} & \multicolumn{4}{c}{{\textbf{Segmentation}}}
& \multicolumn{2}{c}{\zwheader{\textbf{Retrieval}}}
& \multicolumn{2}{c}{{\textbf{Efficiency}}} \\
\cmidrule(lr){2-5}\cmidrule(lr){6-7}\cmidrule(lr){8-9}

\textbf{Model}
& \multicolumn{2}{c}{\textbf{PerSeg}}
& \multicolumn{2}{c|}{\textbf{PerMIS}}
& \multicolumn{1}{c}{\textbf{PerMIR}}
& \textbf{ILIAS}
& \textbf{FPS}
& {\textbf{Params}}  \\

& mIoU  & bIoU  
& mIoU  & bIoU  
& mAP 
& mAP
& img/s $\uparrow$ & M $\downarrow$ \\
\midrule
\multicolumn{9}{l}{\emph{Segmentation-only}} \\

~SEEM~\cite{Zou:2023:SEEM}\pub{NeurIPS'23} 
& \SI{87.1}{} & \SI{55.7}{} 
& \SI{34.2}{} & \SI{31.8}{} 
& \na & \na & 8.5 & 341 \\

~SegGPT~\cite{Wang:2023:SegGPT}\pub{ICCV'23} 
& \SI{94.3}{} & \SI{76.5}{} 
& \SI{38.7}{} & \SI{35.5}{} 
& \na & \na & 12.7 & 354 \\

~PerSAM~\cite{Zhang:2023:PerSAM}\pub{ICLR'24} 
& \SI{89.3}{} & \SI{71.7}{} 
& \SI{42.1}{} & \SI{34.4}{} 
& \na & \na & 2.5 & 945 \\

~PerSAM-F~\cite{Zhang:2023:PerSAM}\pub{ICLR'24} 
& \underline{\SI{95.3}{}} & \underline{\SI{77.9}{}} 
& \underline{\SI{47.8}{}} & \SI{34.5}{} 
& \na & \na & 2.5 & 945 \\

~GF-SAM~\cite{Zhang:2024:GF-SAM}\pub{NeurIPS'24} 
& \SI{91.7}{} & \SI{74.2}{}
& \SI{45.6}{} & \underline{\SI{38.1}{}}
& \na & \na & 1.5 & 945 \\

\midrule
\multicolumn{9}{l}{\emph{Retrieval-only}} \\

~RRT~\cite{Tan:2021:RRT} \pub{ICCV'21} 
& \na & \na & \na & \na 
& 41.7 & 9.2 & 31.0 & \textbf{50} \\

~MatchAnything~\cite{He:2025:MatchAnything}\pub{ArXiv'25} 
& \na & \na & \na & \na 
& 19.7 & 23.2 & 3.4 & 111 \\

~MASt3R~\cite{Leroy:2024:Mast3r}\pub{ECCV'24} 
& \na & \na & \na & \na 
& 37.7 & 25.3 & 1.7 & 688 \\

~RoMav2~\cite{Edstedt:2025:Romav2} \pub{ArXiv'25} 
& \na & \na & \na & \na 
& 32.3 & \underline{26.6} & 2.7 & 425 \\

~AMES~\cite{Suma:2024:Ames} \pub{ECCV'24} 
& \na & \na & \na & \na 
& 38.2 & 26.4 & \textbf{90.9} & {89} \\

\midrule
\multicolumn{9}{l}{\emph{Segmentation + Retrieval}} \\

~PDM~\cite{Samuel:2024:Waldo} \pub{NeurIPS'24} 
& 89.5 & 72.0 & 44.2 & 38.0 
& 65.3 & 14.7 & 1.2 & 1200 \\

~\textbf{\ours{} (ours)} 
& \textbf{96.4} & \textbf{85.6} 
& \textbf{62.6} & \textbf{57.4} 
& \textbf{92.1}
& \textbf{32.5}
& \underline{90.2}
& \underline {52} \\

\bottomrule
\end{tabularx}%
\label{tab:seg_retrieval}
\vspace{-0.2cm}
\end{table*}

\myparagraph{Discussion.}
In personalized segmentation, \ours{} achieves the best performance across datasets. On the challenging PerMIS benchmark, which contains crowded scenes with visually similar distractors, \ours{} surpasses PerSAM-F by +14.8 mIoU and PDM by +18.4 mIoU. The improvement in bIoU is even larger (+22.9 over PerSAM-F and +19.4 over PDM), indicating accurate boundary localization. On PerSeg, \ours{} improves over PerSAM-F by +1.1 mIoU and PDM by +6.9 mIoU, showing gains also in simpler scenes. For retrieval, \ours{} achieves 92.1 mAP on PerMIR, outperforming PDM by +26.8. On the ILIAS benchmark, \ours{} reaches 32.5 mAP@1k, exceeding PDM by +17.8 and surpassing strong task-specific methods by +6.1 mAP@1k (AMES). Importantly, \ours{} achieves these results with a single \SI{52}{M}-parameter model running at 90 FPS, over $75\times$ faster than PDM, faster than specialized segmentation methods, and comparable to the fastest retrieval model (AMES), while keeping \sam{} frozen and training only \SI{5.9}{M} additional parameters.

\subsection{Few-shot Retrieval}
We evaluate \ours{} in the proposed \textbf{few-shot retrieval} setting, which reflects realistic scenarios in which users
can provide multiple examples of the same instance at inference. \ours{} naturally supports this extension: each reference is stored in the memory bank, allowing candidate images to attend jointly to multiple instance representations via memory attention, enabling aggregation of complementary evidence across views. Competing methods lack a unified multi-view representation; we therefore extend them by computing similarity for each reference and averaging the scores. Results in \Cref{tab:n_shot} show that increasing the number of references consistently improves performance, progressively reducing the gap to the Oracle (38.9 mAP). Notably, \ours{} benefits substantially more from additional views than prior approaches, improving from 30.1 mAP in the 1-shot setting to 34.2 mAP with four references (+4.1). In comparison, AMES improves by +2.6 mAP, whereas RoMa v2 yields marginal gains (+0.2).

\definecolor{niceblue}{RGB}{52,120,190}
\definecolor{niceorange}{RGB}{230,140,60}
\definecolor{nicegreen}{RGB}{80,160,120}
\definecolor{baseline1}{RGB}{180,60,120}
\definecolor{baseline2}{RGB}{180,160,20}

\begin{figure*}[t]
\centering
\begin{minipage}[t]{0.5\linewidth}
\vspace{-7pt}
\centering
\captionof{table}{\textbf{Proposed few-shot retrieval on ILIAS, mAP@1k.}
Multiple reference instance examples are provided at inference time. All methods re-rank the same top-1k candidates retrieved with SigLIP \cite{Zhai:2023:SigLIP}.}
\label{tab:n_shot}
\setlength{\tabcolsep}{2.2pt}
\renewcommand{\arraystretch}{1.08}
\tablefontsize
\begin{tabularx}{\linewidth}{@{}X|cccc@{}}
\toprule
 & \multicolumn{4}{c}{N-shot inference} \\
\cmidrule(lr){2-5}
\textbf{Model} & \textbf{1} & \textbf{2} & \textbf{3} & \textbf{4} \\
\midrule
Baseline & 22.9 & 22.9 & 22.9 & 22.9 \\
\midrule
XFeat~\cite{Potje:2024:XFeat} & 15.5 & 16.5 & 17.4 & 17.1 \\
RoMa~\cite{Edstedt:2024:Roma} & 22.0 & 21.9 & 22.1 & 21.9 \\
SP-LightGlue~\cite{Lindenberger:2023:Lightglue} & 22.9 & 23.0 & 23.3 & 22.4 \\
ELoFTR~\cite{Wang:2024:ELoFTR} & 23.8 & 24.6 & 25.7 & 25.0 \\
MASt3R~\cite{Leroy:2024:Mast3r} & 24.3 & 26.0 & 27.2 & 27.5 \\
MatchAnything~\cite{He:2025:MatchAnything} & 25.2 & 26.3 & 27.2 & 26.8 \\
RoMa v2~\cite{Edstedt:2025:Romav2} & 27.5 & \underline{28.2} & 28.4 & 27.7 \\
AMES~\cite{Suma:2024:Ames} & \underline{27.6} & \underline{28.2} & \underline{29.0} & \underline{30.2} \\
\textbf{\ours{} (ours)} & \textbf{30.1} & \textbf{31.8} & \textbf{32.5} & \textbf{34.2} \\
\midrule
\fade{Oracle} & \fade{38.9} & \fade{38.9} & \fade{38.9} & \fade{38.9} \\
\bottomrule
\end{tabularx}
\end{minipage}
\hspace{0.04\linewidth}
\begin{minipage}[t]{0.43\linewidth}
\vspace{1pt}
\centering

\begin{tikzpicture}
\hspace{-0.1cm}
\begin{axis}[
    width=0.8\linewidth,
    height=.73\linewidth,
    scale only axis,
    xlabel={top-$K$ rerank},
    ylabel={mAP (in \%)},
    ylabel near ticks,
    xlabel style={font=\scriptsize, yshift=2pt},
    ylabel style={font=\scriptsize, xshift=3pt, yshift=-5pt},
    xmin=0, xmax=1050,
    ymin=20.5, ymax=33,
    axis x line=bottom,
    axis y line=left,
    axis line style={draw=black!60, line width=0.6pt},
    tick style={draw=black!60, line width=0.6pt},
    tick align=outside,
    grid=major,
    major grid style={draw=black!12, line width=0.3pt},
    tick label style={font=\scriptsize},
    xtick={0, 500, 1000},
    xticklabels={0, 500, 1000},
    line width=0.9pt,
    mark size=2pt,
]
\addplot+[color=niceblue, mark=square*, densely dashed, mark options={solid}] coordinates {
(10,23.4) (50,27.0) (100,28.4) (200,29.5) (500,31.6) (750,32.0) (1000,32.4)
};

\node[font=\footnotesize, text=niceblue]  
at (axis cs:800,31) {\textbf{\ours{}}};
\node[font=\scriptsize, text=niceorange] at (axis cs:800,28) {\textbf{RoMa v2}};
\node[font=\scriptsize, text=nicegreen]  at (axis cs:800,26) {\textbf{AMES}};
\node[font=\scriptsize, text=baseline1]  at (axis cs:800,24) {\textbf{MASt3R}};

\addplot+[color=niceorange, mark=*, mark options={solid}] coordinates {
(10,23.1) (50,26.3) (100,27.3) (200,27.2) (500,27.4) (750, 27.1) (1000,26.9)
};

\addplot+[color=nicegreen, mark=triangle*, mark options={solid}] coordinates {
(10,22.1)  (50,24.2) (100,25.1) (200,25.5) (500,26.3) (750, 26.8) (1000,26.4)
};

\addplot+[color=baseline1, mark=triangle*, mark options={solid}] coordinates {
(10,19.8)  (50,21.3) (100,22.8) (200,24.0) (500,24.7) (750, 25.1) (1000,25.3)
};

\end{axis}
\end{tikzpicture}
\vspace{-0.7cm}
\captionof{figure}{\textbf{Re-ranking vs. candidate pool size (ILIAS), mAP (\%, $\uparrow$).}
As the size of the candidate pool ($K$) increases, more distractors are introduced. \ours{} scales more effectively than prior methods, with a widening margin as K grows.}
\label{fig:scale_k}
\end{minipage}
\vspace{-0.4cm}
\end{figure*}

\subsection{Scaling the Re-Ranking Set}
In \cref{fig:scale_k}, we analyze how re-ranking performance evolves as the candidate pool grows on ILIAS. Larger $K$ introduces more positives but also many more distractors. \ours{} improves from 23.4 mAP at $K{=}10$ to 32.5 at $K{=}1000$ (+9.1), showing strong scalability to large candidate pools. In contrast, RoMa v2 peaks at 27.3 (at $K{=}100$) and drops to 26.9 at $K{=}1000$, a known issue with dense matchers that notoriously suffer from perceptual aliasing \cite{Sferrazza:2025:Match}. AMES increases more gradually (22.1→26.4), remaining below \ours{} across all $K$. Crucially, the advantage of \ours{} amplifies with $K$: the margin over RoMa v2 grows from +0.3 mAP at $K{=}10$ to +6.5 at $K{=}1000$. This trend shows that \ours{} effectively scales with the candidate pool, remaining \textbf{the most robust to the increasing prevalence of distractors as the candidate set grows}.

\subsection{Category-Level Retrieval}
Table~\ref{tab:retrieval_uned_datasets} reports performance across diverse domain-specific retrieval benchmarks. Unlike personalized retrieval, which targets the same physical instance, these benchmarks evaluate retrieval of fine-grained semantic categories, \eg, cars of the same model or birds of the same species. This setting can be viewed as a superset of instance retrieval, as images of the same instance necessarily belong to the same category. Since \ours{} ranks candidates by visual similarity with the query, it naturally prioritizes objects from the same category. Even though category-level retrieval is not the primary focus of our model, \ours{} performs on par or better than prior methods. On Cars196 we improve the baseline by +1.1. On iNat, \ours{} reaches 65.2 (+1.1 over AMES). On product benchmarks, we obtain 97.4 on RP2K (+0.9) and 66.2 on SOP (+0.2). For landmarks, \ours{} achieves 76.1 on RParis (+0.2) and 49.8 on ROxford (+0.7). Overall, results highlight the generalization of \ours{} to category-level retrieval. 

\begin{table*}[t]

\caption{\textbf{Comparison of \ours{} on category-level retrieval and landmark recognition datasets, grouped by domain.} Most methods underperform the baseline (\ie no re-ranking), whereas FoundYou, despite being optimized for instance retrieval, always outperforms the baseline and is competitive across domains. For RParis/ROxford, we follow \cite{Radenovic:2018:ROP}, re-ranking the top-100 candidates and compute mAP (\%, $\uparrow$). On the remaining datasets we follow \cite{Ypsilantis:2023:UNED}, re-rank the top-10 candidates, and evaluate with mAP@5 (\%, $\uparrow$). The retrieval baseline is SigLIP, following \cite{Kordopatis:2025:ILIAS}.}
\label{tab:retrieval_uned_datasets}
\vspace{-0.2cm}
\centering

\tablefontsize
\setlength{\tabcolsep}{2.2pt}
\renewcommand{\arraystretch}{1.05}

\begin{tabularx}{\linewidth}{@{}X|cc|c|c|c|c|ccc@{}}
\toprule
\multicolumn{1}{l}{\multirow{2}{*}{\textbf{Method}}} & \multicolumn{2}{c}{\textbf{Products}} & \multicolumn{1}{c}{\textbf{Cars}} & \multicolumn{1}{c}{\textbf{Animals}} & \multicolumn{1}{c}{\textbf{Food}} & \multicolumn{1}{c}{\textbf{Art}} & \multicolumn{3}{c}{\textbf{Landmarks}} \\
\cmidrule(lr){2-3} \cmidrule(lr){4-4} \cmidrule(lr){5-5} \cmidrule(lr){6-6} \cmidrule(lr){7-7} \cmidrule(lr){8-10}
& {RP2K} & {SOP} & {Cars196} & {iNat} & {Food2K} & {MET} & {GLDv2} & {RPar} & {ROxf} \\
\midrule
Baseline & 94.8 & 65.1 & \underline{90.5} & 63.9 & 93.4 & 77.7 & 87.5 & 74.4 & 39.2 \\
\midrule

XFeat~\cite{Potje:2024:XFeat}          & 96.2 & 54.9 & 89.0 & 58.5 & 92.6 & 65.1 & 84.9 & 74.7 & 43.6 \\
SP-LightGlue~\cite{Lindenberger:2023:Lightglue}   & 95.3 & 58.6 & 89.8 & 60.1 & 92.9 & 70.8 & 88.4 & 75.5 & 48.6 \\
ELoFTR~\cite{Wang:2024:ELoFTR}         & 96.2 & 58.9 & 89.3 & 60.0 & 92.6 & 74.3 & 89.0 & \underline{75.9} & \underline{49.1} \\
MatchAnything~\cite{He:2025:MatchAnything}  & 95.0 & 56.3 & 89.2 & 58.4 & 92.3 & 73.1 & 88.0 & 75.7   & 45.6 \\
RoMa v2~\cite{Edstedt:2025:Romav2}     & 96.1 & 58.9 & 90.0 & 59.2 & 92.4 & 78.6 & 88.0 & \underline{75.9} & 46.2 \\
MASt3R~\cite{Leroy:2024:Mast3r}        & 96.1 & 62.5 & 89.7 & 59.6 & 92.4 & 80.3 & 89.3 & 75.8 & 45.2 \\
AMES~\cite{Suma:2024:Ames}             & \underline{96.5} & \underline{66.0} & 89.8 & \underline{64.1} & \underline{93.6} & \textbf{80.8} & \textbf{91.8} & 75.6 & 48.7 \\
\textbf{\ours{} (ours)}                & \textbf{97.4} & \textbf{66.2} & \textbf{91.6} & \textbf{65.2} & \textbf{94.4} & \underline{80.4} & \underline{91.5} & \textbf{76.1} & \textbf{49.8} \\
\midrule
\fade{Oracle} & \fade{98.0} & \fade{76.3} & \fade{98.3} & \fade{81.2} & \fade{98.2} & \fade{88.4} & \fade{92.9} & \fade{76.3} & \fade{50.3} \\
\bottomrule
\end{tabularx}
\vspace{-0.3cm}
\end{table*}
\newcommand{\vlabel}[1]{\rotatebox[origin=c]{90}{\tablefontsize\textbf{#1}}}

\newcommand{\pairimg}[2]{%
  \begingroup
  \setlength{\fboxrule}{0.4pt}
  \setlength{\fboxsep}{0.5pt}
  \fcolorbox{black!20}{white}{%
    \includegraphics[width=0.59\linewidth,height=0.59\linewidth]{#1}%
    \hspace{1pt}%
    \includegraphics[width=0.59\linewidth,height=0.59\linewidth]{#2}%
  }%
  \endgroup
}

\begin{table*}[t]
\centering
\caption{\textbf{Promptable personalized segmentation.} \textit{Left}: mIoU (\%, $\uparrow$) for different prompts types (point, box, mask). \textit{Right}: qualitative results of \ours{}.}
\label{tab:prompt_seg_with_imgs}
\vspace{-0.2cm}
\begin{tabular}{@{}p{0.51\textwidth}@{\hspace{40pt}}p{0.66\textwidth}@{}}
{\centering
\begingroup
\tablefontsize
\setlength{\tabcolsep}{1pt}
\begin{tabularx}{\linewidth}{@{}l >{\raggedright\arraybackslash}X@{\hspace{2pt}}|@{\hspace{2pt}}cc@{}}
\toprule
Prompt & Method & \tablefontsize \textbf{PerSeg} & \tablefontsize \textbf{PerMIS} \\
\midrule

\multirow{3}{*}{\tablefontsize Point}
& \tablefontsize PDM \cite{Samuel:2024:Waldo}        & \tablefontsize 79.7 & \tablefontsize 34.8 \\
& \tablefontsize PerSAM-F \cite{Zhang:2023:PerSAM}   & \tablefontsize 89.8 & \tablefontsize 41.6 \\
& \tablefontsize \textbf{FoundYou (ours)}                   & \tablefontsize \textbf{95.2} & \tablefontsize \textbf{58.6} \\
\midrule

\multirow{3}{*}{\tablefontsize Box}
& \tablefontsize PDM \cite{Samuel:2024:Waldo}        & \tablefontsize 84.1 & \tablefontsize 39.7 \\
& \tablefontsize PerSAM-F \cite{Zhang:2023:PerSAM}   & \tablefontsize 93.9 & \tablefontsize 46.2 \\
& \tablefontsize \textbf{FoundYou (ours)}                   & \tablefontsize \textbf{95.9} & \tablefontsize \textbf{61.5} \\
\midrule

\multirow{3}{*}{\tablefontsize Mask}
& \tablefontsize PDM \cite{Samuel:2024:Waldo}        & \tablefontsize 89.5 & \tablefontsize 44.2 \\
& \tablefontsize PerSAM-F \cite{Zhang:2023:PerSAM}   & \tablefontsize 95.3 & \tablefontsize 47.8 \\
& \tablefontsize \textbf{FoundYou (ours)}                   & \tablefontsize \textbf{96.4} & \tablefontsize \textbf{62.6} \\
\bottomrule
\end{tabularx}

\endgroup
}
&
{\hspace{-29pt}\centering
\begin{tabular}{@{}c@{\hspace{3pt}}c@{\hspace{15pt}}c@{}}

\vlabel{Point}
&
\begin{minipage}[c]{0.27\linewidth}
\centering
\pairimg{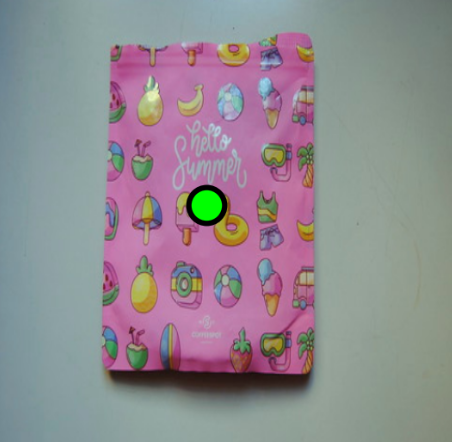}
        {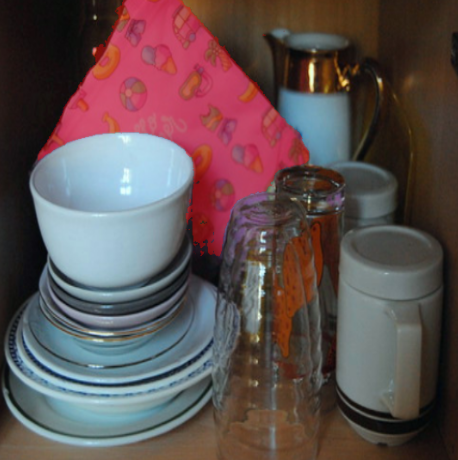}
\end{minipage}
&
\begin{minipage}[c]{0.27\linewidth}
\centering
\pairimg{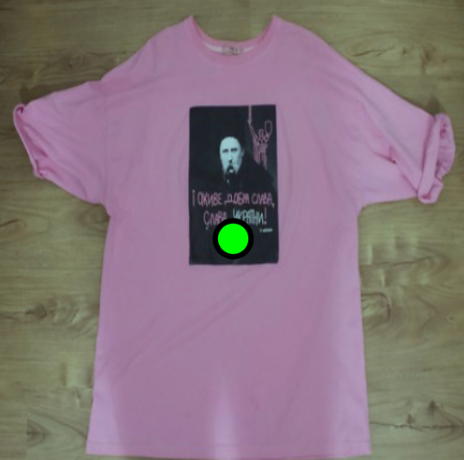}
        {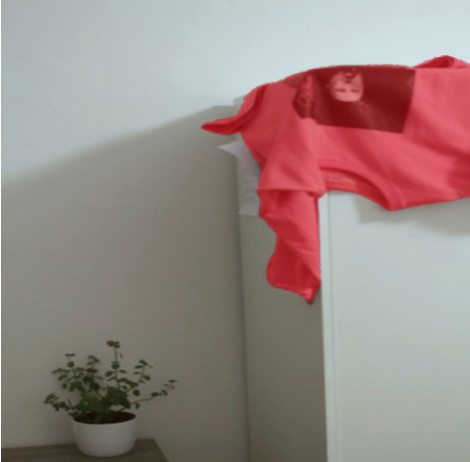}
\end{minipage}
\\[3pt]

\vlabel{Box}
&
\begin{minipage}[c]{0.27\linewidth}
\centering
\pairimg{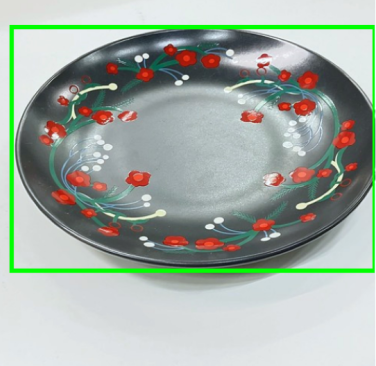}
        {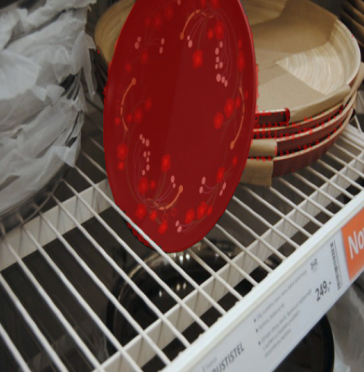}
\end{minipage}
&
\begin{minipage}[c]{0.27\linewidth}
\centering
\pairimg{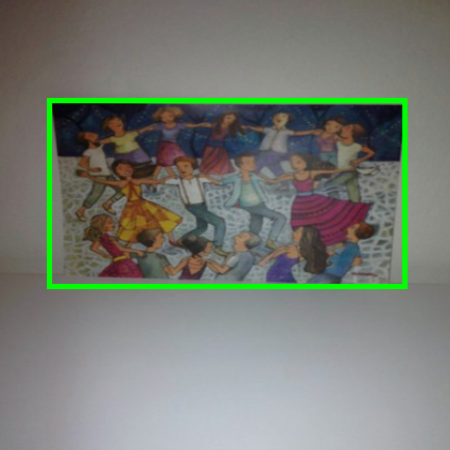}
        {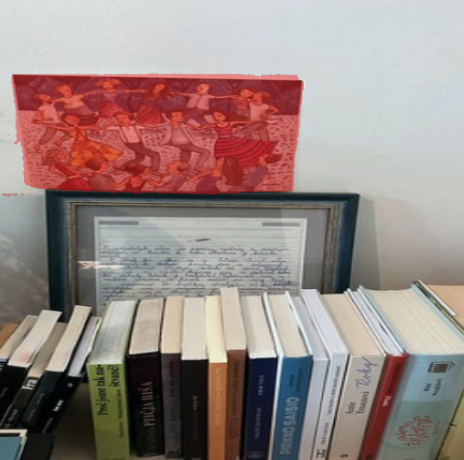}
\end{minipage}
\\[3pt]

\vlabel{Mask}
&
\begin{minipage}[c]{0.27\linewidth}
\centering
\pairimg{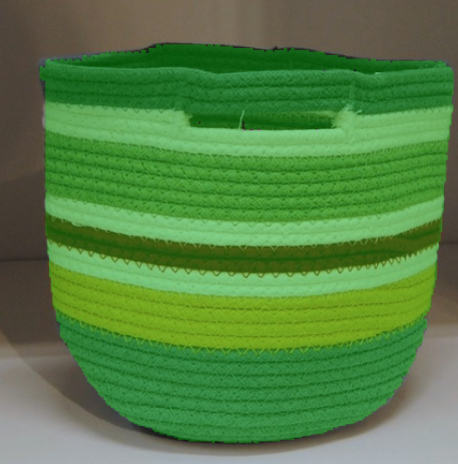}
        {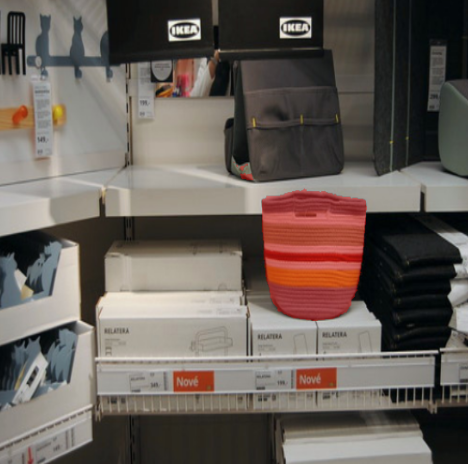}
\end{minipage}
&
\begin{minipage}[c]{0.27\linewidth}
\centering
\pairimg{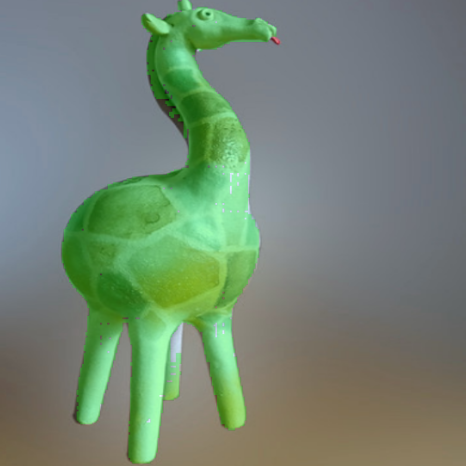}
        {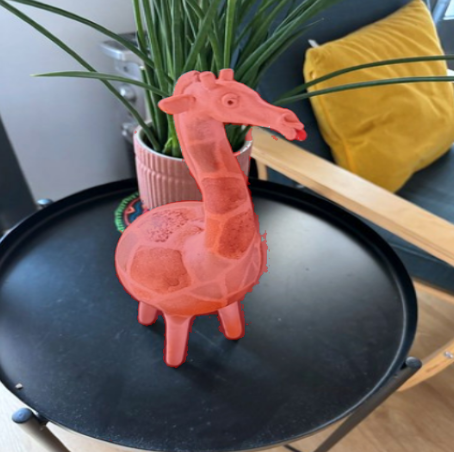}
\end{minipage}

\end{tabular}
}%
\\

\end{tabular}

\end{table*}
\vspace{-0.15cm}

\subsection{Personalized Segmentation with Flexible Prompts}

By keeping the \sam{} decoder frozen, our framework retains the ability to generate masks from flexible prompts, \ie a simple point or bounding box suffices to specify the reference object. \cref{tab:prompt_seg_with_imgs}-left reports results under different prompts. On PerSeg, \ours{} is nearly unaffected by weaker supervision, achieving 96.4 mIoU with masks, 95.9 with boxes, and 95.2 with points. On the more challenging PerMIS benchmark, the gap between mask and point prompts is 4.0 mIoU for \ours{} (62.6→58.6), smaller than PerSAM-F’s 6.2 (47.8→41.6) and PDM’s 9.4 (44.2→34.8). Qualitative examples in \cref{tab:prompt_seg_with_imgs}-right illustrate consistent segmentation across prompt types. Our unified formulation enables feature reuse, lowering annotation cost while improving practicality for real-world deployment.

\begin{table}[b]
\centering
\caption{\textbf{Ablations}. 
\textbf{Left:} parameter-efficient adaptation outperforms full and partial fine-tuning (FT).
\textbf{Middle:} AdaptFormer \cite{Chen:2022:Adaptformer} performs best with a $0.5\times$ bottleneck.
\textbf{Right:} adapting only the last two Image Encoder blocks yields the best performance.
We report mIoU (\%, $\uparrow$) on PerMIS and mAP (\%, $\uparrow$) on ILIAS.}
\vspace{-0.2cm}
\label{tab:ablations}
\tablefontsize
\setlength{\tabcolsep}{3pt}

\begin{minipage}[t]{0.39\linewidth}
\centering
\begin{tabularx}{\linewidth}{@{}Xcc@{}}
\toprule
\multicolumn{3}{c}{\textbf{Fine-tuning strategies}} \\
\cmidrule(lr){1-3}
\textbf{Method} & \textbf{mIoU} & \textbf{mAP} \\
\midrule
Frozen \sam{}  & 60.2 & 3.5 \\
Full FT & 27.8 & 13.8 \\
Backbone FT & 57.2 & 20.5 \\
QKV FT & 59.8 & 27.2 \\
\textbf{Adapt} & \textbf{62.6} & \underline{32.5} \\
Adapt w/o Distill & 56.7 & \textbf{32.7} \\
\bottomrule
\end{tabularx}
\end{minipage}
\hfill
\begin{minipage}[t]{0.3\linewidth}
\centering
\begin{tabularx}{\linewidth}{@{}Xc@{}}
\toprule
\multicolumn{2}{c}{\textbf{Adaptation strategies}} \\
\cmidrule(lr){1-2}
\textbf{Method} & \textbf{mAP} \\
\midrule
LoRA \cite{Hu:2022:LoRa} & 28.9 \\
Adapter \cite{Houlsby:2019:Adapter} & 30.2 \\
AdaptFormer \cite{Chen:2022:Adaptformer} &  \\
\;\; $0.8\times$ channel dim & 30.5 \\
\;\; $\mathbf{0.5\times}$ channel dim & \textbf{32.5} \\
\;\; $0.3\times$ channel dim & 29.8 \\
\bottomrule
\end{tabularx}
\end{minipage}
\hfill
\begin{minipage}[t]{0.25\linewidth}
\centering
\begin{tabularx}{\linewidth}{@{}Xcc@{}}
\toprule
\multicolumn{3}{c}{\textbf{Adapter placement}} \\
\cmidrule(lr){1-3}
\textbf{Stages} &  & \textbf{mAP} \\
\midrule
\\
None & – & 20.4 \\
Early & 0--1 & 21.7 \\
Middle & 1--2 & 29.6 \\
\textbf{Late} & 2--3 & \textbf{32.5} \\
All & 0--3 & 31.8 \\
\bottomrule
\end{tabularx}
\end{minipage}

\vspace{-0.15cm}
\end{table}
\subsection{Ablation Studies}
\label{sec:ablation}
We first ablate the adaptation strategy, adapter design, and adapter placement on personalized segmentation (PerMIS) and retrieval (ILIAS) in \cref{tab:ablations}. We then analyze how adaptation changes the underlying representation in \cref{fig:spatiotemporal_bias}. Additional ablations on the distillation loss weight $\lambda_{\text{dist}}$ and the number of positive and negative training examples are provided in the Supplementary Material.

\myparagraph{Adaptation vs. Fine-tuning.} As shown in \cref{tab:ablations}, applying \sam{} yields strong segmentation (60.2 mIoU) but poor retrieval performance (3.5 mAP), when using the \textit{occlusion score} as the retrieval score. Full or partial fine-tuning shows limited gains, suggesting disruption of the representations learned during pretraining. Lightweight adaptation achieves the best trade-off, reaching 62.6 mIoU and 32.5 mAP. Notably, self-distillation preserves segmentation quality (+5.9 \wrt Adapt w/o Distill) while maintaining strong retrieval performance.

\myparagraph{Adapter Design and Placement.} Among adapter variants, AdaptFormer performs best, with a $0.5\times$ bottleneck achieving 32.5 mAP. Either smaller or larger bottlenecks reduce performance (-2.0 mAP and -2.7 mAP, respectively). Adapting only the final two stages of the Image Encoder (blocks 2–3) yields the highest accuracy. This suggests that instance discrimination benefits from adapting high-level semantic features while preserving low-level spatial representations.

\myparagraph{Mitigating the spatio-temporal continuity bias.}
\sam{} is pretrained to preserve object identity across video frames, encouraging its features to maintain spatio-temporal continuity. As a result, these features may be sensitive to changes in an object position and surrounding context. In \cref{fig:spatiotemporal_bias}, we analyze this behavior on the Object-Placement dataset~\cite{Liu:2022:OPA} by measuring same-instance feature similarity under object translations, background changes, and their combination. Frozen \sam{} features become progressively less similar as these changes increase, whereas the features adapted by \ours{} remain substantially more invariant. This shows that our adaptation reduces the continuity bias, yielding features that are more reliable for matching objects across independent images.

\usepgfplotslibrary{fillbetween}

\definecolor{samTwoBlue}{HTML}{4E79A7}
\definecolor{foundYouOrange}{HTML}{D55E00}
\definecolor{oracleGray}{HTML}{595959}
\definecolor{gridGray}{HTML}{D0D0D0}

\pgfplotsset{
    rebuttal plot/.style={
        width=.78\linewidth,
        height=.42\linewidth,
        scale only axis,
        axis line style={line width=0.8pt},
        axis x line*=bottom,
        axis y line*=left,
        xlabel style={font=\scriptsize, yshift=5pt},
        ylabel style={font=\scriptsize, yshift=-5pt},
        tick label style={font=\tiny},
        ticklabel style={font=\tiny},
        tick style={line width=0.6pt},
        major tick length=1.6pt,
        minor tick length=0pt,
        tick align=outside,
        xticklabel style={
            yshift=2.5pt,
            /pgf/number format/fixed,
            /pgf/number format/precision=1,
            /pgf/number format/zerofill=false
        },
        yticklabel style={
            xshift=1.5pt,
            /pgf/number format/fixed,
            /pgf/number format/precision=1,
            /pgf/number format/zerofill=false
        },
        label style={font=\scriptsize},
        legend image post style={xscale=0.42},
        legend image code/.code={
            \draw[mark repeat=2,mark phase=2,##1]
                plot coordinates {(0cm,0cm) (0.18cm,0cm) (0.36cm,0cm)};
        },
        legend style={
            draw=none,
            fill=none,
            font=\scriptsize,
            row sep=0pt,
            nodes={inner sep=1pt},
            /tikz/every even column/.append style={column sep=1pt}
        },
        legend cell align={left},
        grid=both,
        grid style={gridGray, line width=0.45pt, opacity=0.45},
        major grid style={gridGray, line width=0.45pt, opacity=0.45},
        minor tick num=0
    }
}

\newcommand{\RebuttalSharedLegend}{%
\begingroup
\setlength{\fboxsep}{3.5pt}%
\setlength{\fboxrule}{0.4pt}%
\fcolorbox{gridGray}{white}{%
\makebox[0.90\textwidth][c]{%
\scriptsize
\textbf{Legend:}\hspace{0.8em}%
\tikz[baseline=-0.55ex]
\draw[samTwoBlue,very thick] (0,0)--(0.42,0);%
\hspace{0.22em}SAM~2 frozen features
\hspace{1.4em}%
\tikz[baseline=-0.55ex]
\draw[foundYouOrange,very thick] (0,0)--(0.42,0);%
\hspace{0.22em}\ours{} adapted features
\hspace{1.4em}%
\tikz[baseline=-0.55ex]
\draw[
oracleGray,
opacity=0.72,
line width=1.0pt,
dash pattern=on 4pt off 2.4pt
] (0,0)--(0.42,0);%
\hspace{0.22em}Oracle
}%
}%
\endgroup
}

\newcommand{\RebuttalPlotDiffPosition}{%
\begin{tikzpicture}
\begin{axis}[
    rebuttal plot,
    xmin=0, xmax=0.78,
    ymin=0.4, ymax=1.02,
    xlabel={Displacement $\Delta p$},
    ylabel={Cosine similarity},
]
\addplot[
    draw=none,
    fill=samTwoBlue,
    fill opacity=0.18,
    forget plot
] coordinates {
    (0.0826,0.9464)
    (0.1979,0.8568)
    (0.3330,0.7825)
    (0.4725,0.7171)
    (0.6023,0.6669)
    (0.7274,0.6674)
    (0.7274,0.6015)
    (0.6023,0.6300)
    (0.4725,0.6914)
    (0.3330,0.7677)
    (0.1979,0.8448)
    (0.0826,0.9375)
} \closedcycle;

\addplot[samTwoBlue, very thick] coordinates {
    (0.0826,0.9420)
    (0.1979,0.8508)
    (0.3330,0.7751)
    (0.4725,0.7042)
    (0.6023,0.6484)
    (0.7274,0.6345)
};

\addplot[
    draw=none,
    fill=foundYouOrange,
    fill opacity=0.18,
    forget plot
] coordinates {
    (0.0826,0.9738)
    (0.1979,0.9336)
    (0.3330,0.9136)
    (0.4725,0.8827)
    (0.6023,0.8791)
    (0.7274,0.8833)
    (0.7274,0.8586)
    (0.6023,0.8582)
    (0.4725,0.8658)
    (0.3330,0.9051)
    (0.1979,0.9256)
    (0.0826,0.9684)
} \closedcycle;

\addplot[foundYouOrange, very thick] coordinates {
    (0.0826,0.9711)
    (0.1979,0.9296)
    (0.3330,0.9093)
    (0.4725,0.8743)
    (0.6023,0.8686)
    (0.7274,0.8709)
};

\addplot[
    oracleGray,
    opacity=0.72,
    line width=1.0pt,
    dash pattern=on 4pt off 2.4pt
] coordinates {
    (0,1)
    (0.78,1)
};

\end{axis}
\end{tikzpicture}%
}

\newcommand{\RebuttalPlotDiffBackground}{%
\begin{tikzpicture}
\begin{axis}[
    rebuttal plot,
    xmin=0.4, xmax=1.0,
    ymin=0, ymax=6.1,
    xlabel={Cosine similarity},
    ylabel={Probab. density},
]
\addplot[samTwoBlue, very thick] coordinates {(0.3000,0.0262) (0.3018,0.0271) (0.3035,0.0279) (0.3053,0.0288) (0.3070,0.0296) (0.3088,0.0304) (0.3105,0.0311) (0.3123,0.0319) (0.3140,0.0325) (0.3158,0.0332) (0.3175,0.0338) (0.3193,0.0344) (0.3211,0.0350) (0.3228,0.0355) (0.3246,0.0360) (0.3263,0.0365) (0.3281,0.0369) (0.3298,0.0374) (0.3316,0.0378) (0.3333,0.0382) (0.3351,0.0386) (0.3368,0.0391) (0.3386,0.0395) (0.3404,0.0400) (0.3421,0.0405) (0.3439,0.0410) (0.3456,0.0416) (0.3474,0.0423) (0.3491,0.0430) (0.3509,0.0438) (0.3526,0.0446) (0.3544,0.0456) (0.3561,0.0466) (0.3579,0.0478) (0.3596,0.0491) (0.3614,0.0504) (0.3632,0.0519) (0.3649,0.0536) (0.3667,0.0553) (0.3684,0.0572) (0.3702,0.0592) (0.3719,0.0614) (0.3737,0.0637) (0.3754,0.0661) (0.3772,0.0686) (0.3789,0.0713) (0.3807,0.0741) (0.3825,0.0770) (0.3842,0.0800) (0.3860,0.0831) (0.3877,0.0862) (0.3895,0.0895) (0.3912,0.0928) (0.3930,0.0962) (0.3947,0.0996) (0.3965,0.1031) (0.3982,0.1066) (0.4000,0.1100) (0.4018,0.1135) (0.4035,0.1170) (0.4053,0.1204) (0.4070,0.1238) (0.4088,0.1272) (0.4105,0.1305) (0.4123,0.1337) (0.4140,0.1369) (0.4158,0.1399) (0.4175,0.1429) (0.4193,0.1458) (0.4211,0.1486) (0.4228,0.1513) (0.4246,0.1539) (0.4263,0.1564) (0.4281,0.1588) (0.4298,0.1611) (0.4316,0.1633) (0.4333,0.1654) (0.4351,0.1674) (0.4368,0.1694) (0.4386,0.1712) (0.4404,0.1731) (0.4421,0.1749) (0.4439,0.1766) (0.4456,0.1783) (0.4474,0.1800) (0.4491,0.1817) (0.4509,0.1834) (0.4526,0.1851) (0.4544,0.1869) (0.4561,0.1887) (0.4579,0.1905) (0.4596,0.1925) (0.4614,0.1945) (0.4632,0.1966) (0.4649,0.1988) (0.4667,0.2010) (0.4684,0.2035) (0.4702,0.2060) (0.4719,0.2086) (0.4737,0.2114) (0.4754,0.2144) (0.4772,0.2174) (0.4789,0.2206) (0.4807,0.2239) (0.4825,0.2274) (0.4842,0.2310) (0.4860,0.2348) (0.4877,0.2386) (0.4895,0.2426) (0.4912,0.2467) (0.4930,0.2509) (0.4947,0.2552) (0.4965,0.2596) (0.4982,0.2641) (0.5000,0.2687) (0.5018,0.2733) (0.5035,0.2780) (0.5053,0.2828) (0.5070,0.2875) (0.5088,0.2924) (0.5105,0.2972) (0.5123,0.3021) (0.5140,0.3070) (0.5158,0.3118) (0.5175,0.3167) (0.5193,0.3216) (0.5211,0.3265) (0.5228,0.3314) (0.5246,0.3363) (0.5263,0.3412) (0.5281,0.3462) (0.5298,0.3511) (0.5316,0.3560) (0.5333,0.3610) (0.5351,0.3660) (0.5368,0.3710) (0.5386,0.3761) (0.5404,0.3813) (0.5421,0.3865) (0.5439,0.3918) (0.5456,0.3972) (0.5474,0.4028) (0.5491,0.4084) (0.5509,0.4142) (0.5526,0.4201) (0.5544,0.4263) (0.5561,0.4325) (0.5579,0.4390) (0.5596,0.4457) (0.5614,0.4526) (0.5632,0.4598) (0.5649,0.4672) (0.5667,0.4748) (0.5684,0.4827) (0.5702,0.4909) (0.5719,0.4993) (0.5737,0.5080) (0.5754,0.5171) (0.5772,0.5264) (0.5789,0.5360) (0.5807,0.5459) (0.5825,0.5562) (0.5842,0.5667) (0.5860,0.5775) (0.5877,0.5887) (0.5895,0.6002) (0.5912,0.6119) (0.5930,0.6240) (0.5947,0.6364) (0.5965,0.6492) (0.5982,0.6622) (0.6000,0.6755) (0.6018,0.6892) (0.6035,0.7031) (0.6053,0.7174) (0.6070,0.7320) (0.6088,0.7469) (0.6105,0.7621) (0.6123,0.7776) (0.6140,0.7934) (0.6158,0.8096) (0.6175,0.8261) (0.6193,0.8429) (0.6211,0.8601) (0.6228,0.8776) (0.6246,0.8955) (0.6263,0.9136) (0.6281,0.9322) (0.6298,0.9510) (0.6316,0.9703) (0.6333,0.9898) (0.6351,1.0097) (0.6368,1.0300) (0.6386,1.0506) (0.6404,1.0715) (0.6421,1.0927) (0.6439,1.1142) (0.6456,1.1360) (0.6474,1.1580) (0.6491,1.1803) (0.6509,1.2028) (0.6526,1.2255) (0.6544,1.2484) (0.6561,1.2714) (0.6579,1.2945) (0.6596,1.3177) (0.6614,1.3409) (0.6632,1.3641) (0.6649,1.3872) (0.6667,1.4103) (0.6684,1.4333) (0.6702,1.4561) (0.6719,1.4787) (0.6737,1.5011) (0.6754,1.5233) (0.6772,1.5452) (0.6789,1.5668) (0.6807,1.5880) (0.6825,1.6090) (0.6842,1.6295) (0.6860,1.6498) (0.6877,1.6696) (0.6895,1.6891) (0.6912,1.7083) (0.6930,1.7271) (0.6947,1.7456) (0.6965,1.7638) (0.6982,1.7817) (0.7000,1.7994) (0.7018,1.8168) (0.7035,1.8341) (0.7053,1.8513) (0.7070,1.8683) (0.7088,1.8854) (0.7105,1.9024) (0.7123,1.9195) (0.7140,1.9366) (0.7158,1.9540) (0.7175,1.9715) (0.7193,1.9892) (0.7211,2.0072) (0.7228,2.0255) (0.7246,2.0441) (0.7263,2.0631) (0.7281,2.0825) (0.7298,2.1022) (0.7316,2.1224) (0.7333,2.1430) (0.7351,2.1641) (0.7368,2.1855) (0.7386,2.2073) (0.7404,2.2295) (0.7421,2.2521) (0.7439,2.2750) (0.7456,2.2981) (0.7474,2.3216) (0.7491,2.3452) (0.7509,2.3690) (0.7526,2.3929) (0.7544,2.4169) (0.7561,2.4409) (0.7579,2.4648) (0.7596,2.4886) (0.7614,2.5122) (0.7632,2.5356) (0.7649,2.5588) (0.7667,2.5815) (0.7684,2.6040) (0.7702,2.6259) (0.7719,2.6474) (0.7737,2.6684) (0.7754,2.6888) (0.7772,2.7086) (0.7789,2.7277) (0.7807,2.7462) (0.7825,2.7641) (0.7842,2.7812) (0.7860,2.7976) (0.7877,2.8133) (0.7895,2.8283) (0.7912,2.8426) (0.7930,2.8562) (0.7947,2.8690) (0.7965,2.8812) (0.7982,2.8927) (0.8000,2.9035) (0.8018,2.9137) (0.8035,2.9233) (0.8053,2.9323) (0.8070,2.9408) (0.8088,2.9488) (0.8105,2.9562) (0.8123,2.9632) (0.8140,2.9698) (0.8158,2.9760) (0.8175,2.9819) (0.8193,2.9874) (0.8211,2.9927) (0.8228,2.9977) (0.8246,3.0025) (0.8263,3.0072) (0.8281,3.0117) (0.8298,3.0161) (0.8316,3.0203) (0.8333,3.0246) (0.8351,3.0288) (0.8368,3.0330) (0.8386,3.0372) (0.8404,3.0414) (0.8421,3.0457) (0.8439,3.0500) (0.8456,3.0544) (0.8474,3.0589) (0.8491,3.0635) (0.8509,3.0681) (0.8526,3.0728) (0.8544,3.0775) (0.8561,3.0824) (0.8579,3.0872) (0.8596,3.0921) (0.8614,3.0970) (0.8632,3.1018) (0.8649,3.1067) (0.8667,3.1114) (0.8684,3.1161) (0.8702,3.1206) (0.8719,3.1249) (0.8737,3.1290) (0.8754,3.1329) (0.8772,3.1365) (0.8789,3.1398) (0.8807,3.1427) (0.8825,3.1452) (0.8842,3.1472) (0.8860,3.1488) (0.8877,3.1498) (0.8895,3.1503) (0.8912,3.1502) (0.8930,3.1494) (0.8947,3.1480) (0.8965,3.1458) (0.8982,3.1429) (0.9000,3.1392) (0.9018,3.1348) (0.9035,3.1295) (0.9053,3.1233) (0.9070,3.1162) (0.9088,3.1083) (0.9105,3.0994) (0.9123,3.0895) (0.9140,3.0786) (0.9158,3.0667) (0.9175,3.0538) (0.9193,3.0398) (0.9211,3.0247) (0.9228,3.0085) (0.9246,2.9911) (0.9263,2.9725) (0.9281,2.9528) (0.9298,2.9318) (0.9316,2.9095) (0.9333,2.8859) (0.9351,2.8611) (0.9368,2.8348) (0.9386,2.8072) (0.9404,2.7782) (0.9421,2.7477) (0.9439,2.7158) (0.9456,2.6824) (0.9474,2.6476) (0.9491,2.6112) (0.9509,2.5734) (0.9526,2.5340) (0.9544,2.4932) (0.9561,2.4509) (0.9579,2.4071) (0.9596,2.3620) (0.9614,2.3154) (0.9632,2.2674) (0.9649,2.2182) (0.9667,2.1677) (0.9684,2.1160) (0.9702,2.0633) (0.9719,2.0095) (0.9737,1.9547) (0.9754,1.8992) (0.9772,1.8429) (0.9789,1.7859) (0.9807,1.7284) (0.9825,1.6706) (0.9842,1.6124) (0.9860,1.5541) (0.9877,1.4958) (0.9895,1.4376) (0.9912,1.3795) (0.9930,1.3219) (0.9947,1.2647) (0.9965,1.2081) (0.9982,1.1522) (1.0000,1.0971)};
\addplot[foundYouOrange, very thick] coordinates {(0.3000,0.0000) (0.3018,0.0000) (0.3035,0.0000) (0.3053,0.0000) (0.3070,0.0000) (0.3088,0.0000) (0.3105,0.0000) (0.3123,0.0000) (0.3140,0.0000) (0.3158,0.0000) (0.3175,0.0000) (0.3193,0.0000) (0.3211,0.0000) (0.3228,0.0000) (0.3246,0.0000) (0.3263,0.0000) (0.3281,0.0000) (0.3298,0.0000) (0.3316,0.0000) (0.3333,0.0000) (0.3351,0.0000) (0.3368,0.0000) (0.3386,0.0000) (0.3404,0.0000) (0.3421,0.0000) (0.3439,0.0000) (0.3456,0.0000) (0.3474,0.0000) (0.3491,0.0000) (0.3509,0.0000) (0.3526,0.0000) (0.3544,0.0000) (0.3561,0.0000) (0.3579,0.0000) (0.3596,0.0000) (0.3614,0.0000) (0.3632,0.0000) (0.3649,0.0000) (0.3667,0.0000) (0.3684,0.0000) (0.3702,0.0000) (0.3719,0.0000) (0.3737,0.0000) (0.3754,0.0000) (0.3772,0.0000) (0.3789,0.0000) (0.3807,0.0000) (0.3825,0.0000) (0.3842,0.0000) (0.3860,0.0000) (0.3877,0.0000) (0.3895,0.0000) (0.3912,0.0000) (0.3930,0.0000) (0.3947,0.0001) (0.3965,0.0001) (0.3982,0.0001) (0.4000,0.0001) (0.4018,0.0002) (0.4035,0.0002) (0.4053,0.0003) (0.4070,0.0004) (0.4088,0.0004) (0.4105,0.0006) (0.4123,0.0007) (0.4140,0.0008) (0.4158,0.0010) (0.4175,0.0012) (0.4193,0.0015) (0.4211,0.0018) (0.4228,0.0021) (0.4246,0.0025) (0.4263,0.0030) (0.4281,0.0035) (0.4298,0.0041) (0.4316,0.0047) (0.4333,0.0054) (0.4351,0.0062) (0.4368,0.0070) (0.4386,0.0079) (0.4404,0.0089) (0.4421,0.0100) (0.4439,0.0111) (0.4456,0.0122) (0.4474,0.0134) (0.4491,0.0146) (0.4509,0.0159) (0.4526,0.0172) (0.4544,0.0184) (0.4561,0.0196) (0.4579,0.0208) (0.4596,0.0220) (0.4614,0.0231) (0.4632,0.0241) (0.4649,0.0250) (0.4667,0.0259) (0.4684,0.0266) (0.4702,0.0272) (0.4719,0.0277) (0.4737,0.0281) (0.4754,0.0284) (0.4772,0.0286) (0.4789,0.0287) (0.4807,0.0287) (0.4825,0.0287) (0.4842,0.0286) (0.4860,0.0285) (0.4877,0.0285) (0.4895,0.0285) (0.4912,0.0285) (0.4930,0.0286) (0.4947,0.0289) (0.4965,0.0293) (0.4982,0.0298) (0.5000,0.0305) (0.5018,0.0315) (0.5035,0.0326) (0.5053,0.0339) (0.5070,0.0355) (0.5088,0.0373) (0.5105,0.0393) (0.5123,0.0414) (0.5140,0.0438) (0.5158,0.0463) (0.5175,0.0490) (0.5193,0.0518) (0.5211,0.0546) (0.5228,0.0576) (0.5246,0.0605) (0.5263,0.0634) (0.5281,0.0663) (0.5298,0.0690) (0.5316,0.0717) (0.5333,0.0742) (0.5351,0.0766) (0.5368,0.0788) (0.5386,0.0808) (0.5404,0.0826) (0.5421,0.0842) (0.5439,0.0856) (0.5456,0.0868) (0.5474,0.0878) (0.5491,0.0887) (0.5509,0.0894) (0.5526,0.0900) (0.5544,0.0905) (0.5561,0.0910) (0.5579,0.0915) (0.5596,0.0919) (0.5614,0.0924) (0.5632,0.0930) (0.5649,0.0937) (0.5667,0.0945) (0.5684,0.0954) (0.5702,0.0965) (0.5719,0.0977) (0.5737,0.0991) (0.5754,0.1007) (0.5772,0.1025) (0.5789,0.1043) (0.5807,0.1064) (0.5825,0.1085) (0.5842,0.1108) (0.5860,0.1131) (0.5877,0.1155) (0.5895,0.1179) (0.5912,0.1203) (0.5930,0.1227) (0.5947,0.1251) (0.5965,0.1275) (0.5982,0.1297) (0.6000,0.1319) (0.6018,0.1340) (0.6035,0.1360) (0.6053,0.1379) (0.6070,0.1396) (0.6088,0.1413) (0.6105,0.1428) (0.6123,0.1442) (0.6140,0.1455) (0.6158,0.1467) (0.6175,0.1477) (0.6193,0.1486) (0.6211,0.1494) (0.6228,0.1501) (0.6246,0.1507) (0.6263,0.1511) (0.6281,0.1515) (0.6298,0.1517) (0.6316,0.1519) (0.6333,0.1520) (0.6351,0.1521) (0.6368,0.1521) (0.6386,0.1521) (0.6404,0.1522) (0.6421,0.1523) (0.6439,0.1526) (0.6456,0.1530) (0.6474,0.1535) (0.6491,0.1543) (0.6509,0.1553) (0.6526,0.1566) (0.6544,0.1582) (0.6561,0.1602) (0.6579,0.1625) (0.6596,0.1652) (0.6614,0.1683) (0.6632,0.1718) (0.6649,0.1756) (0.6667,0.1798) (0.6684,0.1843) (0.6702,0.1891) (0.6719,0.1942) (0.6737,0.1995) (0.6754,0.2049) (0.6772,0.2105) (0.6789,0.2163) (0.6807,0.2220) (0.6825,0.2278) (0.6842,0.2337) (0.6860,0.2395) (0.6877,0.2452) (0.6895,0.2510) (0.6912,0.2567) (0.6930,0.2623) (0.6947,0.2680) (0.6965,0.2736) (0.6982,0.2792) (0.7000,0.2849) (0.7018,0.2906) (0.7035,0.2964) (0.7053,0.3024) (0.7070,0.3085) (0.7088,0.3147) (0.7105,0.3211) (0.7123,0.3278) (0.7140,0.3346) (0.7158,0.3417) (0.7175,0.3491) (0.7193,0.3567) (0.7211,0.3646) (0.7228,0.3729) (0.7246,0.3815) (0.7263,0.3905) (0.7281,0.3998) (0.7298,0.4096) (0.7316,0.4199) (0.7333,0.4307) (0.7351,0.4420) (0.7368,0.4540) (0.7386,0.4667) (0.7404,0.4800) (0.7421,0.4942) (0.7439,0.5092) (0.7456,0.5250) (0.7474,0.5418) (0.7491,0.5596) (0.7509,0.5783) (0.7526,0.5981) (0.7544,0.6190) (0.7561,0.6409) (0.7579,0.6639) (0.7596,0.6880) (0.7614,0.7132) (0.7632,0.7395) (0.7649,0.7669) (0.7667,0.7953) (0.7684,0.8248) (0.7702,0.8554) (0.7719,0.8869) (0.7737,0.9195) (0.7754,0.9530) (0.7772,0.9875) (0.7789,1.0229) (0.7807,1.0592) (0.7825,1.0964) (0.7842,1.1345) (0.7860,1.1734) (0.7877,1.2130) (0.7895,1.2535) (0.7912,1.2946) (0.7930,1.3365) (0.7947,1.3791) (0.7965,1.4224) (0.7982,1.4663) (0.8000,1.5109) (0.8018,1.5562) (0.8035,1.6022) (0.8053,1.6489) (0.8070,1.6965) (0.8088,1.7450) (0.8105,1.7945) (0.8123,1.8452) (0.8140,1.8970) (0.8158,1.9503) (0.8175,2.0051) (0.8193,2.0617) (0.8211,2.1201) (0.8228,2.1805) (0.8246,2.2431) (0.8263,2.3081) (0.8281,2.3754) (0.8298,2.4453) (0.8316,2.5177) (0.8333,2.5926) (0.8351,2.6701) (0.8368,2.7499) (0.8386,2.8320) (0.8404,2.9161) (0.8421,3.0021) (0.8439,3.0894) (0.8456,3.1779) (0.8474,3.2672) (0.8491,3.3567) (0.8509,3.4460) (0.8526,3.5346) (0.8544,3.6221) (0.8561,3.7079) (0.8579,3.7915) (0.8596,3.8726) (0.8614,3.9505) (0.8632,4.0250) (0.8649,4.0957) (0.8667,4.1623) (0.8684,4.2246) (0.8702,4.2823) (0.8719,4.3353) (0.8737,4.3837) (0.8754,4.4273) (0.8772,4.4665) (0.8789,4.5012) (0.8807,4.5317) (0.8825,4.5583) (0.8842,4.5813) (0.8860,4.6011) (0.8877,4.6181) (0.8895,4.6328) (0.8912,4.6456) (0.8930,4.6571) (0.8947,4.6677) (0.8965,4.6780) (0.8982,4.6885) (0.9000,4.6996) (0.9018,4.7118) (0.9035,4.7256) (0.9053,4.7413) (0.9070,4.7594) (0.9088,4.7800) (0.9105,4.8035) (0.9123,4.8301) (0.9140,4.8599) (0.9158,4.8930) (0.9175,4.9295) (0.9193,4.9694) (0.9211,5.0125) (0.9228,5.0587) (0.9246,5.1079) (0.9263,5.1598) (0.9281,5.2142) (0.9298,5.2706) (0.9316,5.3289) (0.9333,5.3885) (0.9351,5.4491) (0.9368,5.5100) (0.9386,5.5709) (0.9404,5.6312) (0.9421,5.6902) (0.9439,5.7474) (0.9456,5.8020) (0.9474,5.8533) (0.9491,5.9007) (0.9509,5.9433) (0.9526,5.9804) (0.9544,6.0111) (0.9561,6.0346) (0.9579,6.0500) (0.9596,6.0566) (0.9614,6.0535) (0.9632,6.0400) (0.9649,6.0153) (0.9667,5.9788) (0.9684,5.9299) (0.9702,5.8682) (0.9719,5.7933) (0.9737,5.7049) (0.9754,5.6031) (0.9772,5.4878) (0.9789,5.3594) (0.9807,5.2181) (0.9825,5.0647) (0.9842,4.8999) (0.9860,4.7245) (0.9877,4.5396) (0.9895,4.3464) (0.9912,4.1461) (0.9930,3.9402) (0.9947,3.7301) (0.9965,3.5173) (0.9982,3.3033) (1.0000,3.0895)};
\addplot[oracleGray, opacity=0.72, line width=1.0pt, dash pattern=on 4pt off 2.4pt] coordinates {(0.3000,0.0000) (0.3018,0.0000) (0.3035,0.0000) (0.3053,0.0000) (0.3070,0.0000) (0.3088,0.0000) (0.3105,0.0000) (0.3123,0.0000) (0.3140,0.0000) (0.3158,0.0000) (0.3175,0.0000) (0.3193,0.0000) (0.3211,0.0000) (0.3228,0.0000) (0.3246,0.0000) (0.3263,0.0000) (0.3281,0.0000) (0.3298,0.0000) (0.3316,0.0000) (0.3333,0.0000) (0.3351,0.0000) (0.3368,0.0000) (0.3386,0.0000) (0.3404,0.0000) (0.3421,0.0000) (0.3439,0.0000) (0.3456,0.0000) (0.3474,0.0000) (0.3491,0.0000) (0.3509,0.0000) (0.3526,0.0000) (0.3544,0.0000) (0.3561,0.0000) (0.3579,0.0000) (0.3596,0.0000) (0.3614,0.0000) (0.3632,0.0000) (0.3649,0.0000) (0.3667,0.0000) (0.3684,0.0000) (0.3702,0.0000) (0.3719,0.0000) (0.3737,0.0000) (0.3754,0.0000) (0.3772,0.0000) (0.3789,0.0000) (0.3807,0.0000) (0.3825,0.0000) (0.3842,0.0000) (0.3860,0.0000) (0.3877,0.0000) (0.3895,0.0000) (0.3912,0.0000) (0.3930,0.0000) (0.3947,0.0000) (0.3965,0.0000) (0.3982,0.0000) (0.4000,0.0000) (0.4018,0.0000) (0.4035,0.0000) (0.4053,0.0000) (0.4070,0.0000) (0.4088,0.0000) (0.4105,0.0000) (0.4123,0.0000) (0.4140,0.0000) (0.4158,0.0000) (0.4175,0.0000) (0.4193,0.0000) (0.4211,0.0000) (0.4228,0.0000) (0.4246,0.0000) (0.4263,0.0000) (0.4281,0.0000) (0.4298,0.0000) (0.4316,0.0000) (0.4333,0.0000) (0.4351,0.0000) (0.4368,0.0000) (0.4386,0.0000) (0.4404,0.0000) (0.4421,0.0000) (0.4439,0.0000) (0.4456,0.0000) (0.4474,0.0000) (0.4491,0.0000) (0.4509,0.0000) (0.4526,0.0000) (0.4544,0.0000) (0.4561,0.0000) (0.4579,0.0000) (0.4596,0.0000) (0.4614,0.0000) (0.4632,0.0000) (0.4649,0.0000) (0.4667,0.0000) (0.4684,0.0000) (0.4702,0.0000) (0.4719,0.0000) (0.4737,0.0000) (0.4754,0.0000) (0.4772,0.0000) (0.4789,0.0000) (0.4807,0.0000) (0.4825,0.0000) (0.4842,0.0000) (0.4860,0.0000) (0.4877,0.0000) (0.4895,0.0000) (0.4912,0.0000) (0.4930,0.0000) (0.4947,0.0000) (0.4965,0.0000) (0.4982,0.0000) (0.5000,0.0000) (0.5018,0.0000) (0.5035,0.0000) (0.5053,0.0000) (0.5070,0.0000) (0.5088,0.0000) (0.5105,0.0000) (0.5123,0.0000) (0.5140,0.0000) (0.5158,0.0000) (0.5175,0.0000) (0.5193,0.0000) (0.5211,0.0000) (0.5228,0.0000) (0.5246,0.0000) (0.5263,0.0000) (0.5281,0.0000) (0.5298,0.0000) (0.5316,0.0000) (0.5333,0.0000) (0.5351,0.0000) (0.5368,0.0000) (0.5386,0.0000) (0.5404,0.0000) (0.5421,0.0000) (0.5439,0.0000) (0.5456,0.0000) (0.5474,0.0000) (0.5491,0.0000) (0.5509,0.0000) (0.5526,0.0000) (0.5544,0.0000) (0.5561,0.0000) (0.5579,0.0000) (0.5596,0.0000) (0.5614,0.0000) (0.5632,0.0000) (0.5649,0.0000) (0.5667,0.0000) (0.5684,0.0000) (0.5702,0.0000) (0.5719,0.0000) (0.5737,0.0000) (0.5754,0.0000) (0.5772,0.0000) (0.5789,0.0000) (0.5807,0.0000) (0.5825,0.0000) (0.5842,0.0000) (0.5860,0.0000) (0.5877,0.0000) (0.5895,0.0000) (0.5912,0.0000) (0.5930,0.0000) (0.5947,0.0000) (0.5965,0.0000) (0.5982,0.0000) (0.6000,0.0000) (0.6018,0.0000) (0.6035,0.0000) (0.6053,0.0000) (0.6070,0.0000) (0.6088,0.0000) (0.6105,0.0000) (0.6123,0.0000) (0.6140,0.0000) (0.6158,0.0000) (0.6175,0.0000) (0.6193,0.0000) (0.6211,0.0000) (0.6228,0.0000) (0.6246,0.0000) (0.6263,0.0000) (0.6281,0.0000) (0.6298,0.0000) (0.6316,0.0000) (0.6333,0.0000) (0.6351,0.0000) (0.6368,0.0000) (0.6386,0.0000) (0.6404,0.0000) (0.6421,0.0000) (0.6439,0.0000) (0.6456,0.0000) (0.6474,0.0000) (0.6491,0.0000) (0.6509,0.0000) (0.6526,0.0000) (0.6544,0.0000) (0.6561,0.0000) (0.6579,0.0000) (0.6596,0.0000) (0.6614,0.0000) (0.6632,0.0000) (0.6649,0.0000) (0.6667,0.0000) (0.6684,0.0000) (0.6702,0.0000) (0.6719,0.0000) (0.6737,0.0000) (0.6754,0.0000) (0.6772,0.0000) (0.6789,0.0000) (0.6807,0.0000) (0.6825,0.0000) (0.6842,0.0000) (0.6860,0.0000) (0.6877,0.0000) (0.6895,0.0000) (0.6912,0.0000) (0.6930,0.0000) (0.6947,0.0000) (0.6965,0.0000) (0.6982,0.0000) (0.7000,0.0000) (0.7018,0.0000) (0.7035,0.0000) (0.7053,0.0000) (0.7070,0.0000) (0.7088,0.0000) (0.7105,0.0000) (0.7123,0.0000) (0.7140,0.0000) (0.7158,0.0000) (0.7175,0.0000) (0.7193,0.0000) (0.7211,0.0000) (0.7228,0.0000) (0.7246,0.0000) (0.7263,0.0000) (0.7281,0.0000) (0.7298,0.0000) (0.7316,0.0000) (0.7333,0.0000) (0.7351,0.0000) (0.7368,0.0000) (0.7386,0.0000) (0.7404,0.0000) (0.7421,0.0000) (0.7439,0.0000) (0.7456,0.0000) (0.7474,0.0000) (0.7491,0.0000) (0.7509,0.0000) (0.7526,0.0000) (0.7544,0.0000) (0.7561,0.0000) (0.7579,0.0000) (0.7596,0.0000) (0.7614,0.0000) (0.7632,0.0000) (0.7649,0.0000) (0.7667,0.0000) (0.7684,0.0000) (0.7702,0.0000) (0.7719,0.0000) (0.7737,0.0000) (0.7754,0.0000) (0.7772,0.0000) (0.7789,0.0000) (0.7807,0.0000) (0.7825,0.0000) (0.7842,0.0000) (0.7860,0.0000) (0.7877,0.0000) (0.7895,0.0000) (0.7912,0.0000) (0.7930,0.0000) (0.7947,0.0000) (0.7965,0.0000) (0.7982,0.0000) (0.8000,0.0000) (0.8018,0.0000) (0.8035,0.0000) (0.8053,0.0000) (0.8070,0.0000) (0.8088,0.0000) (0.8105,0.0000) (0.8123,0.0000) (0.8140,0.0000) (0.8158,0.0000) (0.8175,0.0000) (0.8193,0.0000) (0.8211,0.0000) (0.8228,0.0000) (0.8246,0.0000) (0.8263,0.0000) (0.8281,0.0000) (0.8298,0.0000) (0.8316,0.0000) (0.8333,0.0000) (0.8351,0.0000) (0.8368,0.0000) (0.8386,0.0000) (0.8404,0.0000) (0.8421,0.0000) (0.8439,0.0000) (0.8456,0.0000) (0.8474,0.0000) (0.8491,0.0000) (0.8509,0.0000) (0.8526,0.0000) (0.8544,0.0000) (0.8561,0.0000) (0.8579,0.0000) (0.8596,0.0000) (0.8614,0.0000) (0.8632,0.0000) (0.8649,0.0000) (0.8667,0.0000) (0.8684,0.0000) (0.8702,0.0000) (0.8719,0.0000) (0.8737,0.0000) (0.8754,0.0000) (0.8772,0.0000) (0.8789,0.0000) (0.8807,0.0001) (0.8825,0.0001) (0.8842,0.0001) (0.8860,0.0002) (0.8877,0.0003) (0.8895,0.0003) (0.8912,0.0005) (0.8930,0.0006) (0.8947,0.0009) (0.8965,0.0011) (0.8982,0.0015) (0.9000,0.0020) (0.9018,0.0027) (0.9035,0.0035) (0.9053,0.0046) (0.9070,0.0060) (0.9088,0.0078) (0.9105,0.0100) (0.9123,0.0128) (0.9140,0.0164) (0.9158,0.0208) (0.9175,0.0263) (0.9193,0.0331) (0.9211,0.0414) (0.9228,0.0515) (0.9246,0.0638) (0.9263,0.0787) (0.9281,0.0965) (0.9298,0.1178) (0.9316,0.1431) (0.9333,0.1730) (0.9351,0.2081) (0.9368,0.2491) (0.9386,0.2967) (0.9404,0.3516) (0.9421,0.4147) (0.9439,0.4866) (0.9456,0.5683) (0.9474,0.6604) (0.9491,0.7637) (0.9509,0.8787) (0.9526,1.0062) (0.9544,1.1464) (0.9561,1.2998) (0.9579,1.4665) (0.9596,1.6464) (0.9614,1.8393) (0.9632,2.0447) (0.9649,2.2620) (0.9667,2.4899) (0.9684,2.7274) (0.9702,2.9729) (0.9719,3.2246) (0.9737,3.4803) (0.9754,3.7379) (0.9772,3.9949) (0.9789,4.2485) (0.9807,4.4961) (0.9825,4.7347) (0.9842,4.9615) (0.9860,5.1736) (0.9877,5.3682) (0.9895,5.5428) (0.9912,5.6950) (0.9930,5.8226) (0.9947,5.9238) (0.9965,5.9972) (0.9982,6.0417) (1.0000,6.0566)};
\end{axis}
\end{tikzpicture}%
}

\newcommand{\RebuttalPlotDiffBackgroundPosition}{%
\begin{tikzpicture}
\begin{axis}[
    rebuttal plot,
    xmin=0, xmax=0.90,
    ymin=0.4, ymax=1.02,
    xlabel={Displacement $\Delta p$},
    ylabel={Cosine similarity},
]
\addplot[
    draw=none,
    fill=samTwoBlue,
    fill opacity=0.18,
    forget plot
] coordinates {
    (0.0945,0.8620)
    (0.2230,0.7870)
    (0.3709,0.7071)
    (0.5148,0.6328)
    (0.6586,0.5727)
    (0.8458,0.5330)
    (0.8458,0.4890)
    (0.6586,0.5377)
    (0.5148,0.6113)
    (0.3709,0.6930)
    (0.2230,0.7752)
    (0.0945,0.8465)
} \closedcycle;

\addplot[samTwoBlue, very thick] coordinates {
    (0.0945,0.8542)
    (0.2230,0.7811)
    (0.3709,0.7001)
    (0.5148,0.6221)
    (0.6586,0.5552)
    (0.8458,0.5110)
};

\addplot[
    draw=none,
    fill=foundYouOrange,
    fill opacity=0.18,
    forget plot
] coordinates {
    (0.0945,0.9259)
    (0.2230,0.8929)
    (0.3709,0.8631)
    (0.5148,0.8377)
    (0.6586,0.8253)
    (0.8458,0.8640)
    (0.8458,0.8248)
    (0.6586,0.8090)
    (0.5148,0.8240)
    (0.3709,0.8534)
    (0.2230,0.8852)
    (0.0945,0.9158)
} \closedcycle;

\addplot[foundYouOrange, very thick] coordinates {
    (0.0945,0.9208)
    (0.2230,0.8890)
    (0.3709,0.8583)
    (0.5148,0.8308)
    (0.6586,0.8172)
    (0.8458,0.8444)
};

\addplot[
    oracleGray,
    opacity=0.72,
    line width=1.0pt,
    dash pattern=on 4pt off 2.4pt
] coordinates {
    (0,1)
    (0.90,1)
};

\end{axis}
\end{tikzpicture}%
}

\begin{figure*}[t]
\vspace{-0.6em}
\centering

\begin{minipage}[t]{0.32\textwidth}
\centering

\begin{minipage}{0.96\linewidth}
    \centering
    \RebuttalPlotDiffPosition
\end{minipage}

\vspace{-0.35em}

\begin{minipage}{0.96\linewidth}
    \centering
    \includegraphics[
        width=0.31\linewidth,
        height=0.40in,
        keepaspectratio
    ]{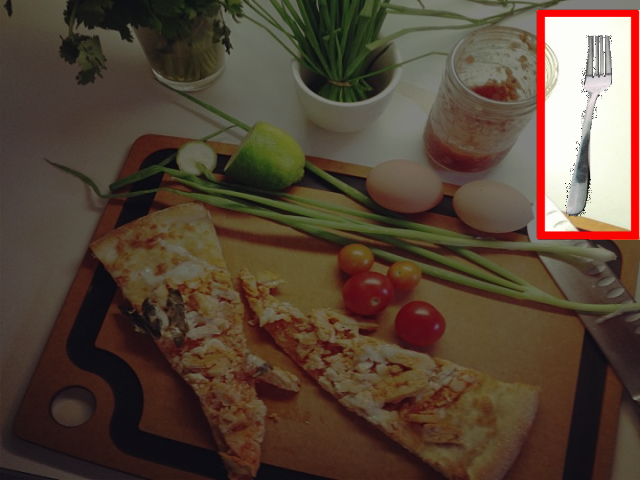}
    \hfill
    \includegraphics[
        width=0.31\linewidth,
        height=0.40in,
        keepaspectratio
    ]{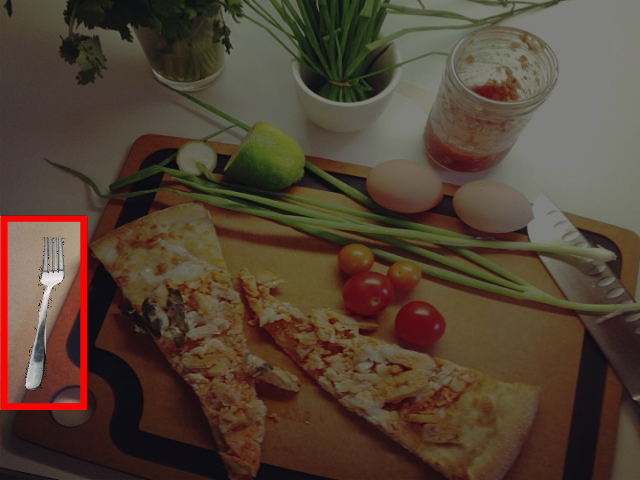}
    \hfill
    \includegraphics[
        width=0.31\linewidth,
        height=0.40in,
        keepaspectratio
    ]{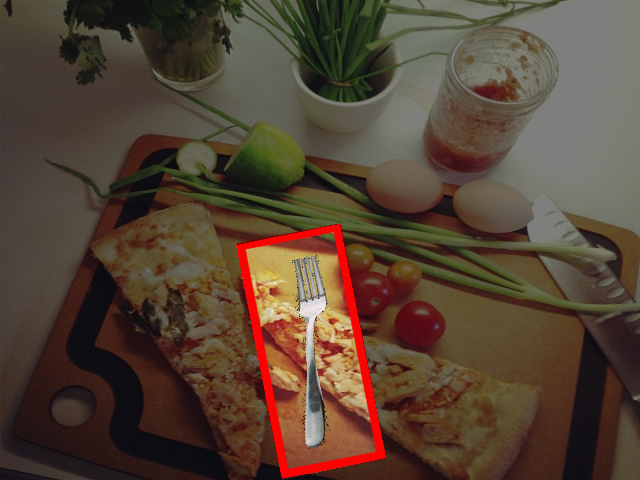}
\end{minipage}

\resizebox{0.98\linewidth}{!}{\textbf{(a)} Same background, translate object}

\end{minipage}
\hfill
\begin{minipage}[t]{0.32\textwidth}
\centering

\begin{minipage}{0.96\linewidth}
    \centering
    \RebuttalPlotDiffBackground
\end{minipage}

\vspace{-0.35em}

\begin{minipage}{0.96\linewidth}
    \centering
    \includegraphics[
        width=0.31\linewidth,
        height=0.40in,
        keepaspectratio
    ]{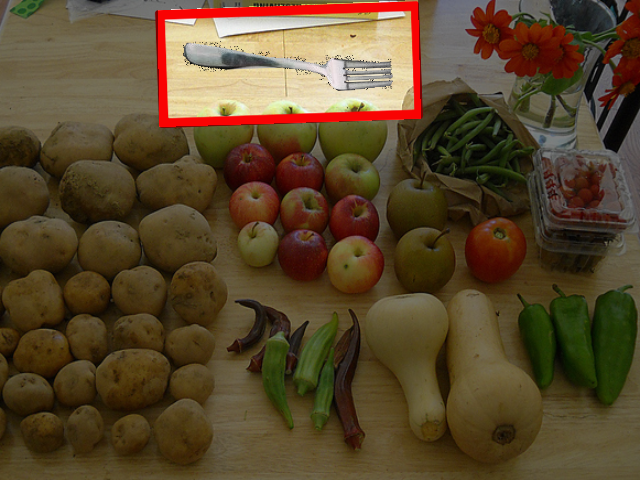}
    \hfill
    \includegraphics[
        width=0.31\linewidth,
        height=0.40in,
        keepaspectratio
    ]{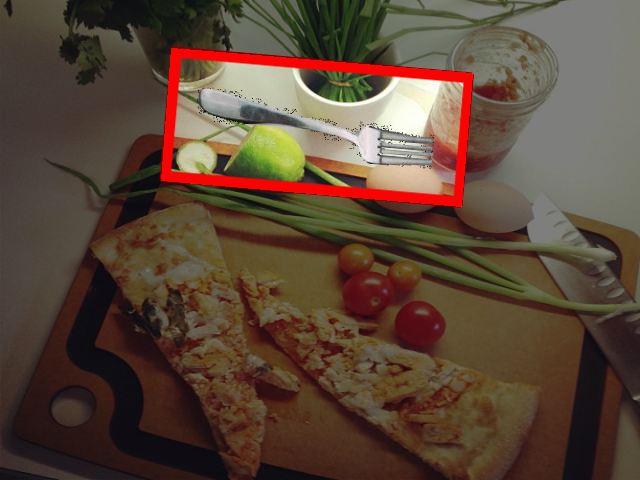}
    \hfill
    \includegraphics[
        width=0.31\linewidth,
        height=0.40in,
        keepaspectratio
    ]{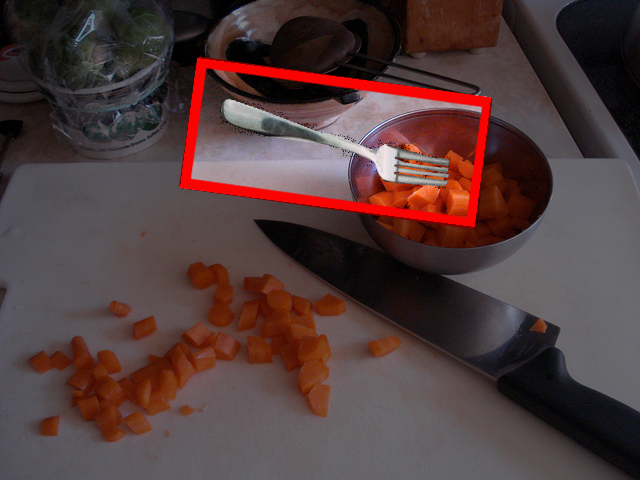}
\end{minipage}

\resizebox{0.98\linewidth}{!}{\textbf{(b)} Fixed object, varying background}

\end{minipage}
\hfill
\begin{minipage}[t]{0.32\textwidth}
\centering

\begin{minipage}{0.96\linewidth}
    \centering
    \RebuttalPlotDiffBackgroundPosition
\end{minipage}

\vspace{-0.35em}

\begin{minipage}{0.96\linewidth}
    \centering
    \includegraphics[
        width=0.31\linewidth,
        height=0.40in,
        keepaspectratio
    ]{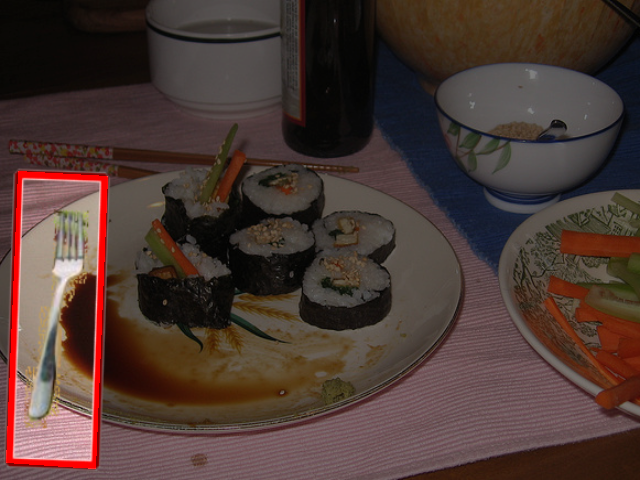}
    \hfill
    \includegraphics[
        width=0.31\linewidth,
        height=0.40in,
        keepaspectratio
    ]{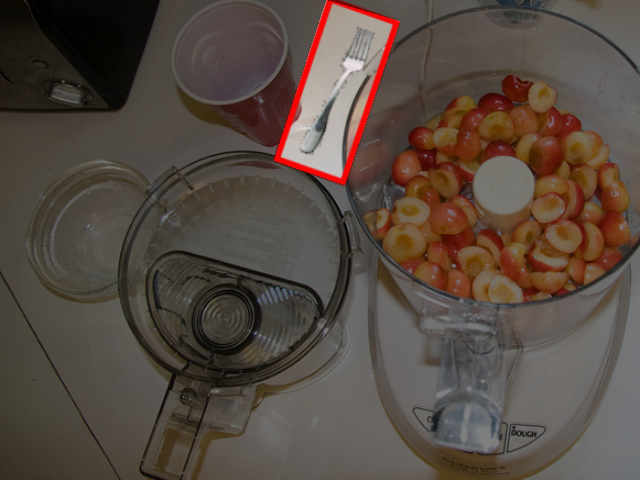}
    \hfill
    \includegraphics[
        width=0.31\linewidth,
        height=0.40in,
        keepaspectratio
    ]{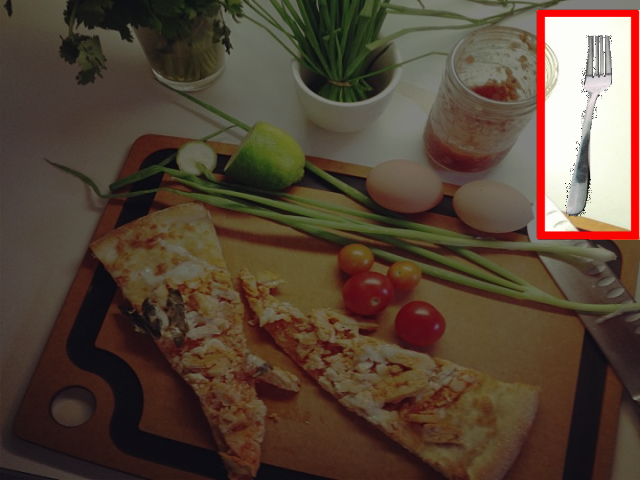}
\end{minipage}

\resizebox{0.98\linewidth}{!}{\textbf{(c)} Varying background and position}

\end{minipage}

\vspace{0.25em}

\RebuttalSharedLegend


\caption{\textbf{Adaptation mitigates SAM~2 spatio-temporal continuity bias.}
Same-instance feature similarity under \textbf{(a)} object translation, \textbf{(b)} background replacement, and \textbf{(c)} both changes. \ours{} features remain more invariant across these shifts.}

\label{fig:spatiotemporal_bias}

\vspace{-0.8em}
\end{figure*}

\section{Conclusion}

We presented \ours{}, a unified framework for personalized segmentation and retrieval that extends \sam{} from video tracking to cross-image instance matching. With lightweight adaptation layers and a compact retrieval head, while keeping \sam{} entirely frozen, we remove its temporal continuity bias and enable discrimination of a specific physical instance across independent images. The resulting \SI{52}{M}-parameter model outperforms prior unified and task-specific methods, scales effectively with additional references, and remains robust to weaker prompts. More broadly, our results suggest that large-scale video pretraining in \sam{} induces identity-aware representations that extend beyond tracking, providing a strong foundation for instance-level reasoning across tasks.

\inparagraph{Acknowledgements.} 
The Ministry of Science, Innovation and Universities of Spain has supported this work through FPU23/\allowbreak00587 (Marcos Alfaro). This research work is part of the projects PID2023-149575OB-I00, funded by MICIU/\allowbreak AEI/\allowbreak 10.13039/\allowbreak 501100011033 and by FEDER, UE, and CIPROM/2024/8, funded by Generalitat Valenciana (program PROMETEO 2025). Claudia Cuttano was supported by the Sustainable Mobility Center (CNMS), which received funding from European Union Next Generation EU (Piano Nazionale di Ripresa e Resilienza (PNRR), Missione 4 Componente 2 Investimento 1.4), grant agreement no.\ CN\_00000023. We acknowledge the CINECA award
under the ISCRA initiative, for availability of high performance computing resources.

\bibliographystyle{splncs04}
\bibliography{bibtex/short, bibtex/references}

\clearpage
\appendix
\renewcommand{\theHsection}{supp.\Alph{section}}
\renewcommand{\theHsubsection}{supp.\Alph{section}.\arabic{subsection}}
\renewcommand{\theHsubsubsection}{supp.\Alph{section}.\arabic{subsection}.\arabic{subsubsection}}
\setcounter{section}{0}
\renewcommand\thesection{\Alph{section}}
\setcounter{page}{1}
\pagenumbering{roman}

\setcounter{figure}{5}
\setcounter{table}{5}

\section*{Appendix}

\noindent{}In this appendix, we provide additional insights into our \ours{} as well as additional experiments. 
Specifically:
\begin{itemize}
    \item \textbf{Additional experiments.} \Cref{sec:supp_exp} reports additional ablation studies regarding our method.
    \item \textbf{Comparison to few-shot segmentation.} \Cref{sec_supp_few-shot-segmentation} reports a comparison with traditional few-shot segmentation methods, including experiments with similar \sam{} based methods.
    \item \textbf{Few-shot retrieval details.} \Cref{sec:supp_few-shot} provides further details on the proposed few-shot retrieval setting.
    \item \textbf{Qualitative results.} \Cref{sec:supp_qual} we report qualitative examples of our method against task-specific and unified approaches for both segmentation and retrieval tasks.
    \item \textbf{Failure cases.} \Cref{sec:failure_qual} reports failure cases of our method to provide insights for future work.
    \item \textbf{Dataset details.} \Cref{sec:supp_data} reports more information on all datasets used for evaluation.
\end{itemize}

\section{Additional Experiments}
\label{sec:supp_exp}
We provide additional ablations in \cref{tab:ablations_supp}, to better understand the design choices of our approach and the role of \sam{} representations. Specifically, we analyze three aspects: (i) the effect of the distillation weight that balances preserving segmentation-friendly spatial structure and encouraging instance-level discrimination, (ii) the impact of the number of positive and negative reference samples used during training, and (iii) the influence of the underlying feature backbone by replacing \sam{} with alternative pretrained encoders while keeping the rest of the pipeline identical. These experiments help isolate the contributions of the training objective, the training setup, and the representation itself to the final performance.

\myparagraph{Distillation weight.}
We study the impact of the distillation weight $\lambda_{\text{dist}}$, which balances preserving segmentation-relevant information from the original SAM 2 representations and enforcing instance-discriminative cues during adaptation. The distillation objective encourages the adapted features to retain the spatial structure required for accurate mask prediction, while the retrieval-oriented loss promotes instance-specific matching and reduces positional bias inherited from the video training of SAM 2. As shown in Table~\ref{tab:ablations_supp} (left), increasing the weight improves segmentation performance, with mIoU rising from 56.7 ($\lambda=0$) to 63.3 ($\lambda=10$). However, stronger distillation progressively harms retrieval performance, as mAP decreases from 32.7 to 25.4, indicating that overly constraining the representation limits the emergence of stronger instance-level discrimination. We therefore use $\lambda_{\text{dist}}=0.5$, which provides a balanced trade-off between segmentation quality (62.6 mIoU) and retrieval performance (32.5 mAP).

\myparagraph{Number of positive and negative training examples.}
We ablate the number of positive and negative candidates used in the personalization loss at each training batch, where positives correspond to images containing the same physical instance as the reference and negatives correspond to non-matching instances. Both are necessary for effective learning. Positives provide appearance variability of the target instance, while negatives supply the discriminative signal required to separate the instance from visually similar objects. Our results show that performance benefits from configurations skewed toward more negatives than positives. The best performance is obtained with 4 positives and 8 negatives (32.5 mAP), which we adopt in the main experiments. Nearby configurations perform slightly worse, \eg 4/10 yields 32.0 mAP, while balanced or positive-heavy setups such as 6/6 and 6/8 drop to 31.4 and 30.8 mAP respectively. This behavior is expected: hard negatives provide the strongest supervisory signal for instance discrimination and therefore must be sufficiently represented in each batch, while too few positives bias the model toward predicting negative scores.

\begin{table}[t]
\centering
\caption{\textbf{Additional Ablation.}
\textbf{Left:} effect of the distillation loss weight $\lambda_{\text{dist}}$.
\textbf{Middle:} impact of number of positive and negative reference samples used during training.
\textbf{Right:} effect of replacing the SAM 2 backbone with alternative pretrained feature encoders, all trained with the same lightweight adapters (AdaptFormer), to assess their instance-awareness.
We report mIoU (\%, $\uparrow$) on PerMIS and mAP (\%, $\uparrow$) on ILIAS.}
\vspace{-0.2cm}
\label{tab:ablations_supp}
\tablefontsize
\setlength{\tabcolsep}{2pt}
\begin{minipage}[t]{0.21\linewidth}
\centering
\begin{tabularx}{\linewidth}{@{}Xcc@{}}
\toprule
\multicolumn{3}{c}{\textbf{Distill. weight}} \\
\cmidrule(lr){1-3}
$\lambda_{\text{dist}}$ & \textbf{mIoU} & \textbf{mAP} \\
\midrule
0 & 56.7 & 32.7 \\
0.1 &  58.4    & 32.8 \\
0.5 &  62.6 & 32.5 \\
 1 &   62.9  &  29.1 \\
 10  & 63.3 & 25.4 \\
\bottomrule
\end{tabularx}
\end{minipage}
\hspace{0.15cm}
\begin{minipage}[t]{0.34\linewidth}
\centering
\begin{tabularx}{\linewidth}{@{}cccccc@{}}
\toprule
\multicolumn{6}{c}{\textbf{Pos/Neg number}} \\
\cmidrule(lr){1-6}
\textbf{Pos} & 2 & 4 & 6 & 8 & 10 \\
\midrule
2  & \cellcolor{red!73!green!20}26.4 & \cellcolor{red!61!green!20}27.5 & \cellcolor{red!53!green!20}28.2 & \cellcolor{red!37!green!20}29.7 & \cellcolor{red!62!green!20}27.4 \\
4  & \cellcolor{red!61!green!20}27.5 & \cellcolor{red!43!green!20}29.1 & \cellcolor{red!19!green!20}31.3 & \cellcolor{green!45!white}\textbf{32.5} & \cellcolor{red!11!green!20}32.0 \\
6  & \cellcolor{red!76!green!20}26.2 & \cellcolor{red!48!green!20}28.7 & \cellcolor{red!18!green!20}31.4 & \cellcolor{red!24!green!20}30.8 & \cellcolor{red!23!green!20}30.9 \\
8  & \cellcolor{red!92!green!20}24.7 & \cellcolor{red!79!green!20}25.9 & \cellcolor{red!69!green!20}26.8 & \cellcolor{red!29!green!20}30.4 & \cellcolor{red!28!green!20}30.5 \\
10 & \cellcolor{red!94!green!20}24.5 & \cellcolor{red!86!green!20}25.3 & \cellcolor{red!66!green!20}27.1 & \cellcolor{red!51!green!20}28.4 & \cellcolor{red!47!green!20}28.8 \\
\bottomrule
\end{tabularx}
\end{minipage}
\hspace{0.15cm}
\begin{minipage}[t]{0.35\linewidth}
\centering
\begin{tabularx}{\linewidth}{@{}X|c@{}}
\toprule
\multicolumn{1}{c|}{\textbf{Pre-trained features}} &\textbf{mAP} \\

\midrule
\textbf{\ours{}} (\sam{}) & \textbf{32.5} \\
\midrule
\sam{} from scratch & 7.2 \\
DINOv2~\cite{Oquab:2023:Dinov2}    & 26.6 \\
DINOv3~\cite{Simeoni:2025:Dinov3}    & 18.5 \\
PEspatial~\cite{Bolya:2025:PE} & 22.8 \\
SigLIP2~\cite{Tschannen:2025:SigLIP2}   & 15.2 \\
\bottomrule
\end{tabularx}
\end{minipage}
\vspace{-0.15cm}
\end{table}

\myparagraph{Pre-trained feature encoders.}
A central hypothesis of our work is that SAM 2 representations are inherently well suited for instance-level recognition, as the model is trained to preserve object identity across video frames. To validate this, we replace the SAM 2 backbone with alternative pretrained feature encoders while keeping the rest of the pipeline identical. In our standard setting, the SAM 2 encoder and memory attention remain frozen and only lightweight adapters and the decoder are trained. To ensure a fair comparison, when replacing the backbone we additionally fine-tune the memory attention to allow the architecture to adapt to the different feature space. Importantly, we also include a variant where SAM 2 is trained from scratch without loading the pretrained checkpoint. As shown in Table~\ref{tab:ablations_supp} (right), this variant performs poorly (7.2 mAP), demonstrating that finetuning alone does not solve the task and that the pretrained representations are critical. While strong visual encoders such as DINOv2 achieve competitive performance (26.6 mAP), \sam{} features yield the best results (32.5 mAP). These results support our claim that representations learned through video object tracking encode strong instance-level identity cues, which transfer naturally to personalized matching across independent images.

\section{Few-Shot vs. Personalized Segmentation}
\label{sec_supp_few-shot-segmentation}
\begin{table}[t]
\centering
\caption{\textbf{Few-shot segmentation methods evaluated on personalized segmentation benchmarks.}
We report mIoU and bIoU (\%).}
\label{tab:few_shot}
\setlength{\tabcolsep}{6pt}
\begin{tabular}{lcccc}
\toprule
& \multicolumn{2}{c}{PerSeg} & \multicolumn{2}{c}{PerMIS} \\
\cmidrule(lr){2-3}
\cmidrule(lr){4-5}
Method & mIoU & bIoU & mIoU & bIoU \\
\midrule
SANSA & 95.5 & 83.6 & 38.2 & 32.5 \\
FS-SAM2 & 94.2 & 78.7 & 50.3 & 43.0 \\
\midrule
\textbf{\ours{}} & \textbf{96.4} & \textbf{85.6} & \textbf{62.6} & \textbf{57.4} \\
\bottomrule
\end{tabular}
\end{table}

SANSA \cite{Cuttano:2025:Sansa} and FS-SAM2 \cite{Forni:2025:fs-sam2} are particularly close in spirit to \ours{}, as they also use \sam{} without external feature encoders to match objects from a reference prompt. However, they address few-shot segmentation, where the goal is to segment all objects belonging to the reference category. In contrast, personalized segmentation requires identifying only the \emph{same physical instance} while rejecting other objects of the same category. We report their results on PerSeg and PerMIS in \cref{tab:few_shot}. Interestingly, both methods achieve similarly strong performance on PerSeg. This benchmark typically contains a single salient foreground object, so category-level and instance-level predictions often coincide (\eg when only one dog is present, segmenting the dog category is equivalent to segmenting that specific dog). This limitation motivated PDM \cite{Samuel:2024:Waldo} to introduce PerMIS, which contains scenes with multiple same-category instances and therefore better exposes the distinction between the two tasks. As expected, on PerMIS, both methods degrade substantially.

\section{Few-Shot Retrieval Details}
\label{sec:supp_few-shot}

In the main paper, we describe our method assuming a single reference visual prompt $V_r$. Here we clarify how the formulation naturally extends to the \textbf{few-shot retrieval} setting, where multiple reference examples are provided.

Let a user provide a set of $N$ visual prompts:
\begin{equation}
\mathcal{V}_r = \{V_r^{(n)}\}_{n=1}^{N},
\end{equation}
where each prompt is defined as
\begin{equation}
V_r^{(n)} = (\mathcal{I}_r^{(n)}, \mathcal{A}_r^{(n)}).
\end{equation}

Here $\mathcal{I}_r^{(n)} \in \mathbb{R}^{H \times W \times 3}$ is the $n$-th reference image and $\mathcal{A}_r^{(n)}$ specifies the target instance within that image. All prompts depict the \emph{same physical instance}, potentially under different viewpoints or imaging conditions. Each reference image is processed independently by the frozen \sam{} \texttt{Image Encoder}:
\begin{equation}
\mathcal{F}_r^{(n)} =
\texttt{Image Encoder}(\mathcal{I}_r^{(n)}).
\end{equation}

A reference mask $\hat{\mathcal{M}}_r^{(n)}$ is obtained either directly from the annotation ($\hat{\mathcal{M}}_r^{(n)} = \mathcal{A}_r^{(n)}$) or predicted from a bounding box using the mask decoder. The memory encoder then produces a memory representation:
\begin{equation}
\mathcal{S}_r^{(n)} =
\mathrm{conv}(\mathcal{F}_r^{(n)}) +
\mathrm{conv}(\hat{\mathcal{M}}_r^{(n)}).
\end{equation}

Each memory tensor $\mathcal{S}_r^{(n)} \in
\mathbb{R}^{\frac{H}{16} \times \frac{W}{16} \times D}$ encodes the representation of the instance in the corresponding reference image. All reference memories are stored in the \texttt{Memory Bank}:
\begin{equation}
\texttt{Memory Bank} =
{\mathcal{S}_r^{(1)}, \mathcal{S}_r^{(2)}, \dots, \mathcal{S}_r^{(N)}}.
\end{equation}

\begin{figure*}[t]
        \centering
    \includegraphics[width=0.98\linewidth]{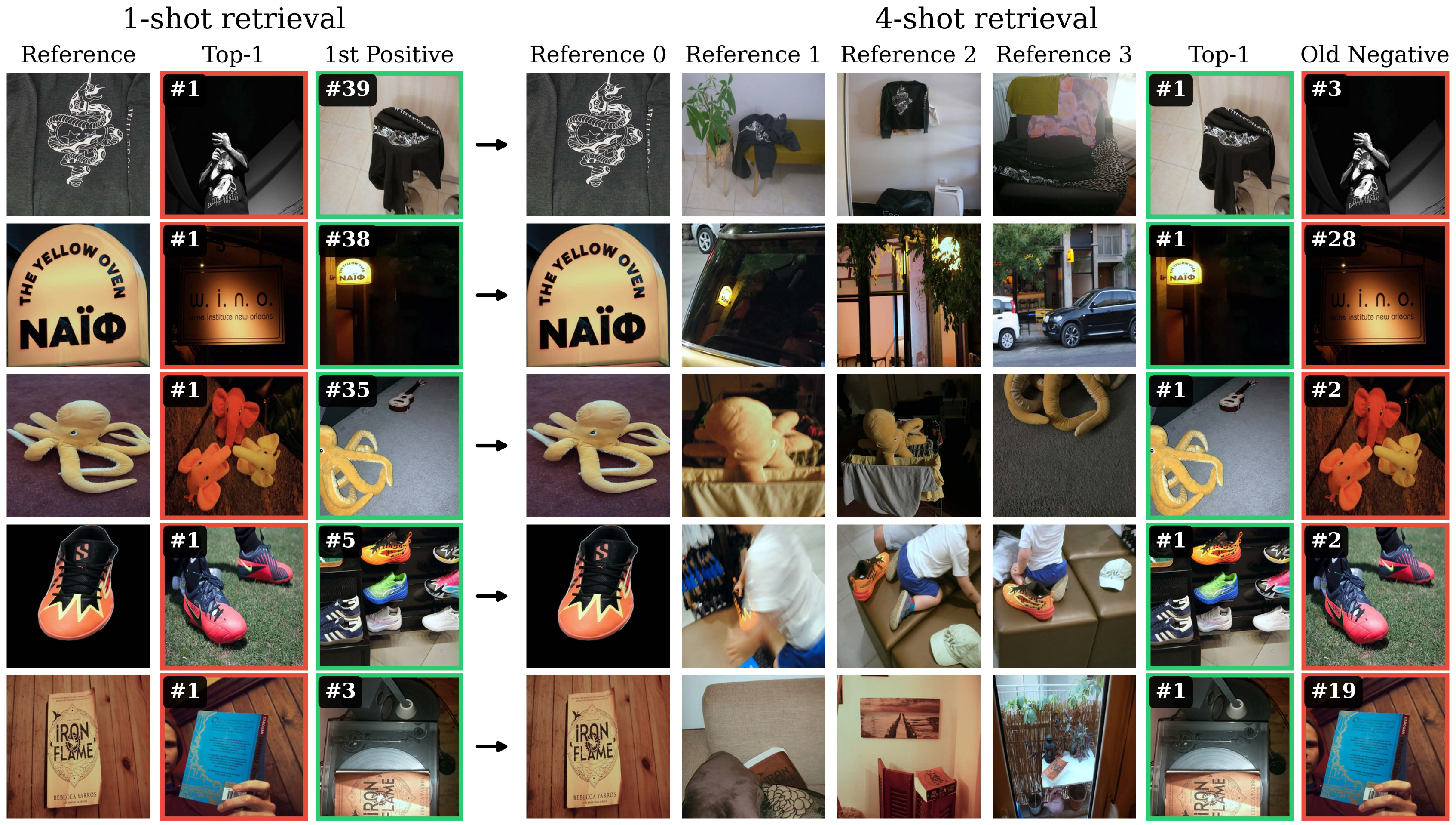}
    \caption{\textbf{Qualitative results for few-shot Personalized Retrieval}. On the left, 1-shot retrieval examples, which includes a reference image, the top-1 retrieved (negative) and the first retrieved positive with its \colorbox{black}{\textcolor{white}{\#rank}}. On the right, 4-shot retrieval for the same instances, including the four reference images fed to \ours{}, the new top-1 retrieved (now a positive) and the negative with its new \colorbox{black}{\textcolor{white}{\#rank}}.
    }
    \label{fig:nshot}
    \vspace{-0.2em}
\end{figure*}

Following the \sam{} design, we concatenate them along the spatial dimension and treat them as a single memory tensor:
\begin{equation}
\mathcal{S}_r =
\mathrm{concat}
\left(
\mathcal{S}_r^{(1)},\dots,\mathcal{S}_r^{(N)}
\right).
\end{equation}

After reshaping into a sequence, the memory contains
$L_r = N \cdot \frac{H}{16}\frac{W}{16} $
tokens.

Given a target image $\mathcal{I}_t$, the image encoder produces target features
\begin{equation}
\mathcal{F}_t =
\texttt{Image Encoder}(\mathcal{I}_t).
\end{equation}

\texttt{Memory Attention} matches target features against reference memories jointly:
\begin{equation}
\mathcal{F}_t^{\text{mem}} =
\mathrm{MHCA}
\left(
Q(\mathcal{F}_t),
K(\mathcal{S}_r),
V(\mathcal{S}_r)
\right).
\end{equation}

Because the keys and values contain tokens from all references, each location in the target image can attend to multiple views of the instance simultaneously, allowing the model to aggregate complementary evidence across viewpoints.

The memory-conditioned features $\mathcal{F}_t^{\text{mem}}$ are then processed by the retrieval decoder (\cref{sec:decoders} in the main paper) to produce the final similarity score:
\begin{equation}
s_t = f_{\text{ret}}(\mathcal{V}_r, \mathcal{I}_t).
\end{equation}

When $N=1$, the formulation reduces to the single-reference case:
\begin{equation}
\mathcal{S}_r = \mathcal{S}_r^{(1)}.
\end{equation}

Few-shot retrieval therefore corresponds to storing multiple instance representations in the \texttt{Memory Bank}, enabling joint matching across them.

\myparagraph{Adapting Competing Methods to Few-shot Retrieval.}
Competing methods evaluated in \cref{tab:n_shot} are designed for matching \emph{pairs of images} and do not provide a mechanism to aggregate multiple reference views. To evaluate them in the few-shot setting, we adapt them as follows. Given $N$ reference prompts $\mathcal{V}r = {V_r^{(n)}}{n=1}^{N}$ and a candidate image $\mathcal{I}t$, we compute an individual similarity score for each reference:
\begin{equation}
s_t^{(n)} = f_{\text{ret}}(V_r^{(n)}, \mathcal{I}_t).
\end{equation}

The final retrieval score is obtained by averaging the scores across references:
\begin{equation}
s_t =
\frac{1}{N}
\sum_{n=1}^{N}
s_t^{(n)}.
\end{equation}

\myparagraph{Qualitative results on Few-shot Personalized Retrieval.}
\Cref{fig:nshot} shows qualitative results of \ours{} under the 1-shot and 4-shot retrieval setting. These examples show how multiple reference inputs enhance the retrieval capabilities, pulling positive gallery images to top-rank positions and pushing hard negatives to low-rank positions.

\section{Qualitative Results}
\label{sec:supp_qual}

\begin{figure*}[h]
    \vspace{-2pt}
    \centering
    \includegraphics[width=0.87\linewidth]{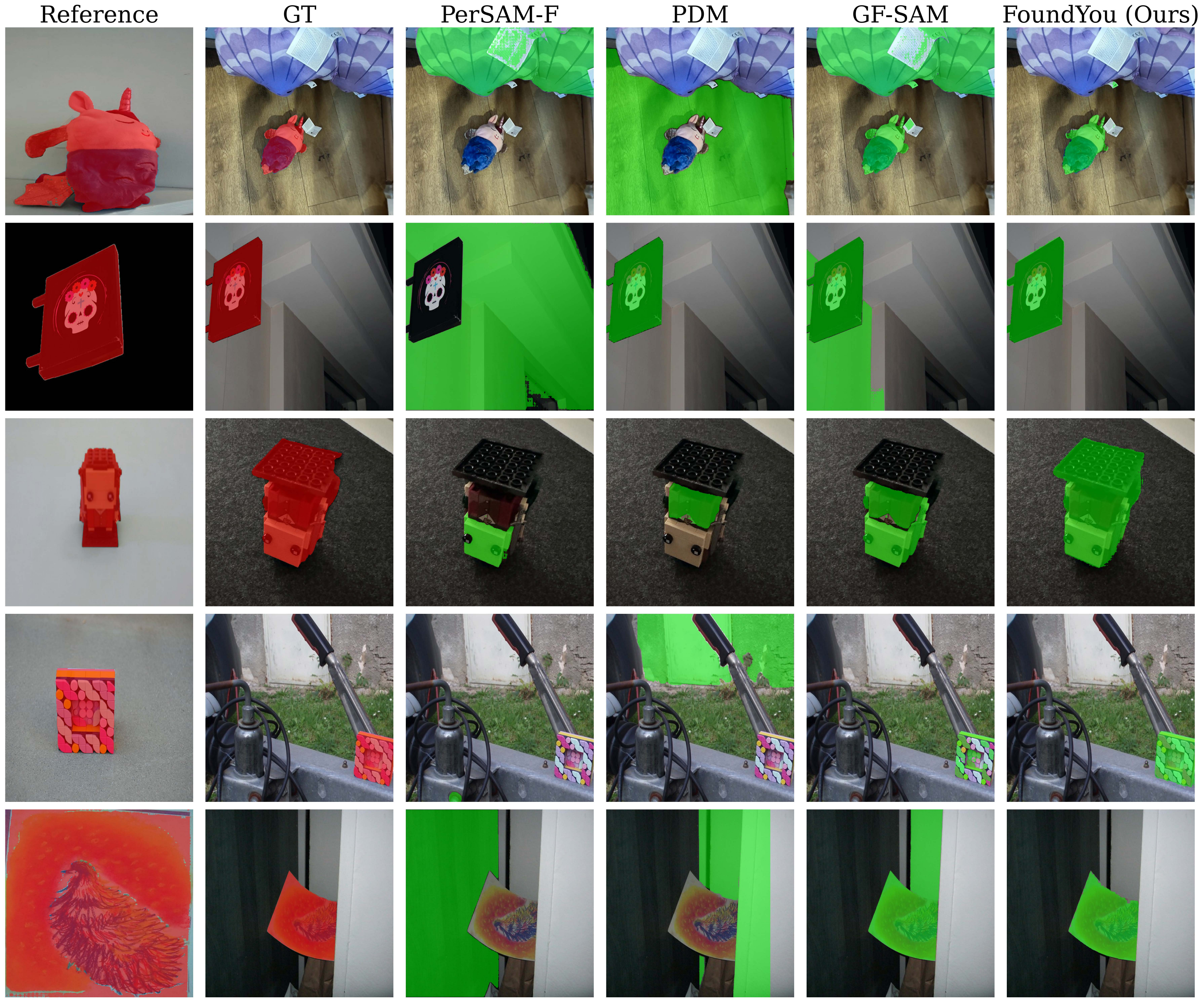}
    \caption{\textbf{Qualitative comparison for Personalized Segmentation}. From left to right: reference image with the reference mask indicating the target instance, ground-truth mask on the target image, and predictions from PerSAM-F \cite{Zhang:2023:PerSAM}, PDM \cite{Samuel:2024:Waldo}, GF-SAM \cite{Zhang:2024:GF-SAM}, and \ours{}. Our method more consistently localizes the reference instance in the target image and produces more accurate segmentation masks.
    }
    \label{fig:supp_qual_seg}
    \vspace{-0.4em}
\end{figure*}

\myparagraph{Qualitative results on Personalized Segmentation.}
\Cref{fig:supp_qual_seg} presents qualitative comparisons between \ours{} and recent personalized segmentation approaches, including PerSAM-F~\cite{Zhang:2023:PerSAM}, PDM~\cite{Samuel:2024:Waldo}, and GF-SAM~\cite{Zhang:2024:GF-SAM}. 
Each example shows a reference image with the annotated instance, the ground-truth mask in the target image, and the corresponding predictions of each method.
Overall, \ours{} more consistently localizes the correct instance and produces more accurate segmentation masks. In contrast, competing approaches often either fail to identify the correct instance in the target image or produce incomplete masks when multiple visually similar objects are present.

\begin{figure*}[h]
        \centering
    \includegraphics[width=0.85\linewidth]{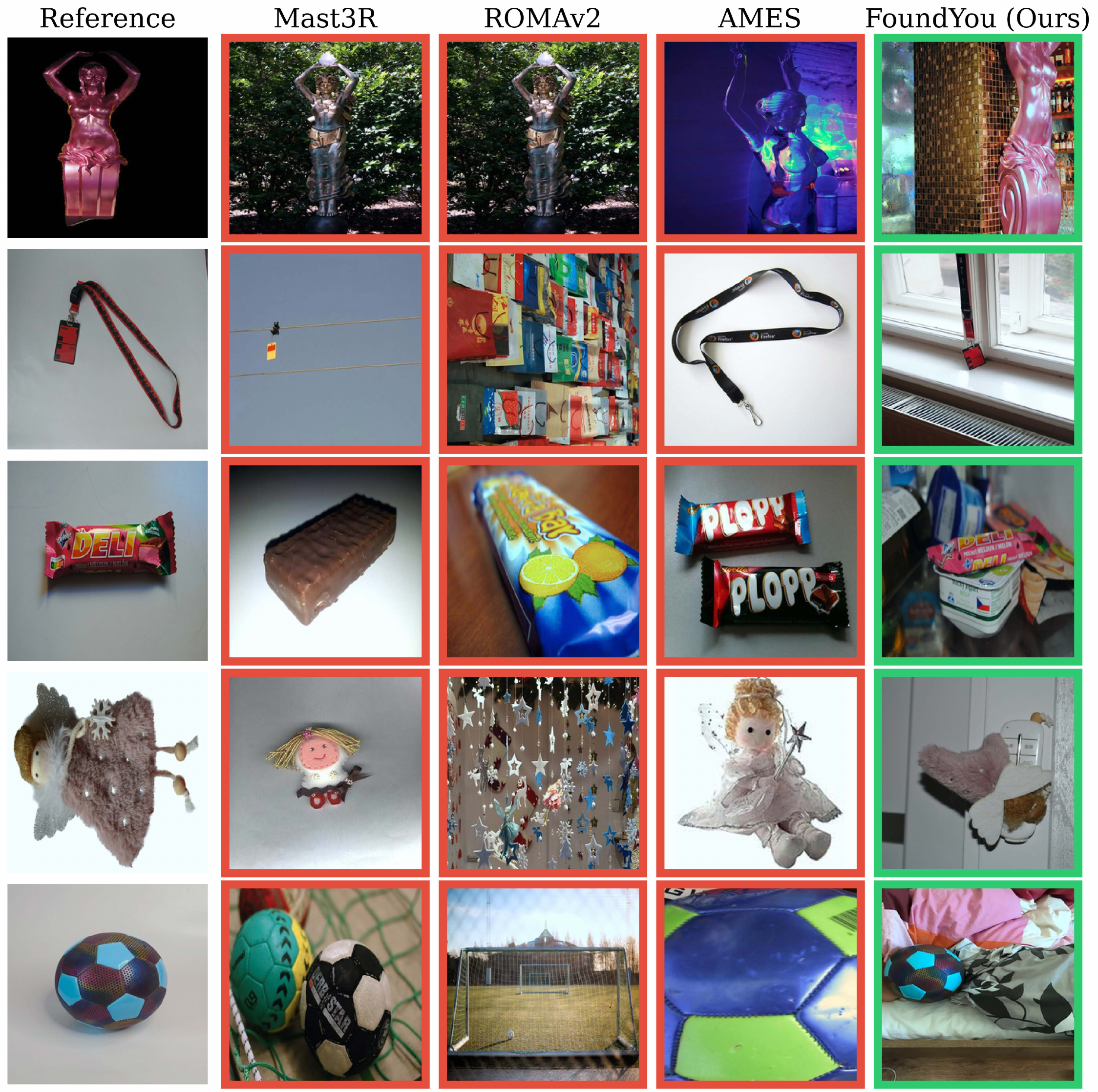}
    \caption{\textbf{Qualitative results for Personalized Retrieval}. The first column shows the reference instance. The following columns show the top-1 retrieved image for MASt3R \cite{Leroy:2024:Mast3r}, RoMa v2 \cite{Edstedt:2025:Romav2}, AMES \cite{Suma:2024:Ames}, and \ours{}. The negatives ranked first by competing methods are visually very similar to the reference instance, highlighting the difficulty of the task. \ours{} successfully retrieves the correct instance.
    }
    \label{fig:supp_qual_ret}
    \vspace{-0.2em}
\end{figure*}

\myparagraph{Qualitative results on Personalized Retrieval.}
\Cref{fig:supp_qual_ret} illustrates qualitative retrieval results. 
For each query instance, we show the top-1 retrieved image returned by MASt3R~\cite{Leroy:2024:Mast3r}, RoMa v2~\cite{Edstedt:2025:Romav2}, AMES~\cite{Suma:2024:Ames}, and \ours{}. 
These examples highlight the difficulty of the task: the negatives ranked first by competing methods are often visually very similar to the reference instance. Despite these challenging distractors, \ours{} is able to correctly retrieve the target instance, demonstrating stronger instance-level discrimination.
 
\begin{figure*}[h]
        \centering
    \includegraphics[width=\linewidth]{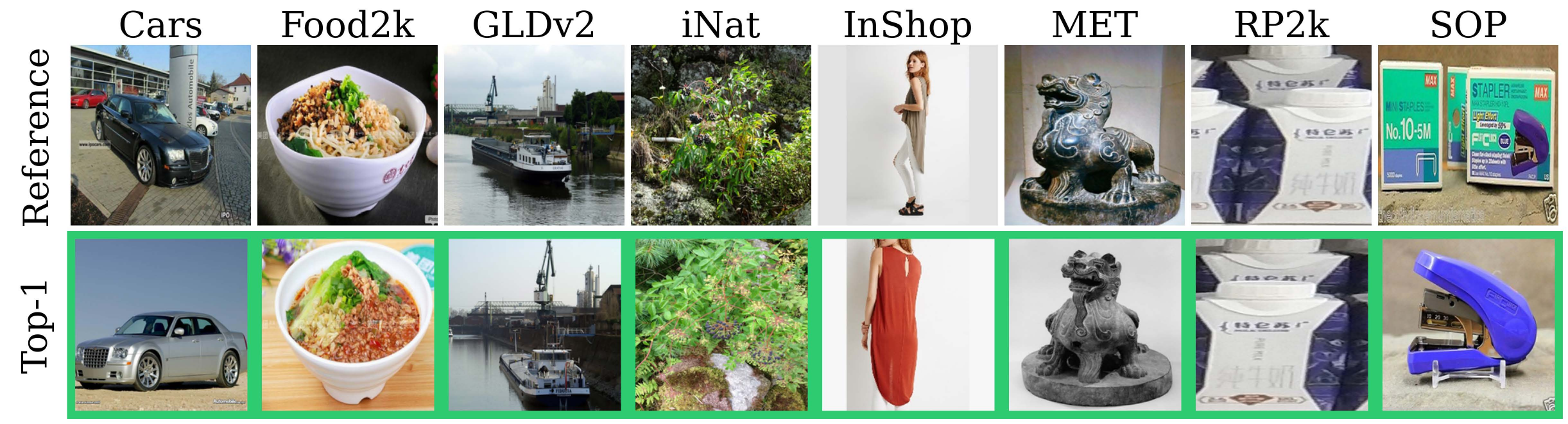}
    \caption{\textbf{Qualitative results on category-level retrieval} across multiple datasets. For each query image (top), we show the top-1 retrieved result (bottom).}
    \label{fig:supp_categ_ret}
    \vspace{-0.2em}
\end{figure*}

\myparagraph{Qualitative results on Category-level Retrieval.}
\Cref{fig:supp_categ_ret} shows qualitative examples on category-level retrieval benchmarks. 
In contrast to personalized retrieval, where the objective is to retrieve the \emph{same physical instance} (\eg, the user’s specific toy, as in \cref{fig:supp_qual_ret}), these benchmarks require retrieving images belonging to the \emph{same fine-grained category}, such as cars of the same model, plants of the same species, products of the same type, or images of the same landmark. \ours{} ranks candidates according to visual similarity with the query. Consequently, although designed for instance-level matching, it naturally retrieves images that share the similar visual attributes. These examples illustrate that the learned representation extends beyond strict instance matching.
\begin{table*}[t]
\centering
\small
\caption{Datasets used for evaluating personalized segmentation and retrieval. 
Retrieval benchmarks include both strict instance-level and category-level settings.}
\label{tab:datasets}
\vspace{-0.2cm}
\setlength{\tabcolsep}{4pt}
\adjustbox{max width=\textwidth}{%
\begin{tabular}{l|l|l|c|c|c|c}
\toprule
\textbf{Task} & \textbf{Dataset} & \textbf{Domain} & 
\textbf{Class Def.} & \textbf{\# Queries} & 
\textbf{\# Images / Gallery} & \textbf{Query Type} \\
\midrule
\multirow{2}{*}{Segmentation}
& PerSeg  & Objects & Instance & 40   & 216  & Img + mask \\
& PerMIS  & Urban & Instance & 216     & 432  & Img + mask \\
\midrule
& PerMIR   & Urban            & Instance & 216 & 432 & Img + bbox \\\
& ILIAS    & Multi-domain     & Instance & 1,232 & 100M & Img + bbox \\
\cmidrule{2-7}
\multirow{8}{*}{Retrieval}
& Cars196  & Cars        & Category & 1,000 & 1,7K & Img \\
& iNat     & Animals/Plants     & Category & 1,000 & 51K   & Img \\
& RP2K     & Products    & Mixed     & 1,000 & 17K & Img \\
& SOP      & Products    & Mixed     & 1,000 & 10K   & Img \\
& Food2K   & Food        & Category & 1,000 & 49k & Img \\
& MET      & Art         & Instance & 129 & 38K   & Img \\
& GLDv2    & Landmarks   & Instance & 1,000 & 157K   & Img \\
& RParis+1M   & Landmarks   & Instance & 1,000 & 1M  & Img + bbox \\
& ROxford+1M  & Landmarks   & Instance & 1,000 & 1M  & Img + bbox \\
\bottomrule
\end{tabular}%
}
\end{table*}

\section{Failure Cases}
\label{sec:failure_qual}
\begin{figure*}[t]
\centering

\begin{minipage}[t]{0.4\linewidth}
\centering
\includegraphics[width=\linewidth]{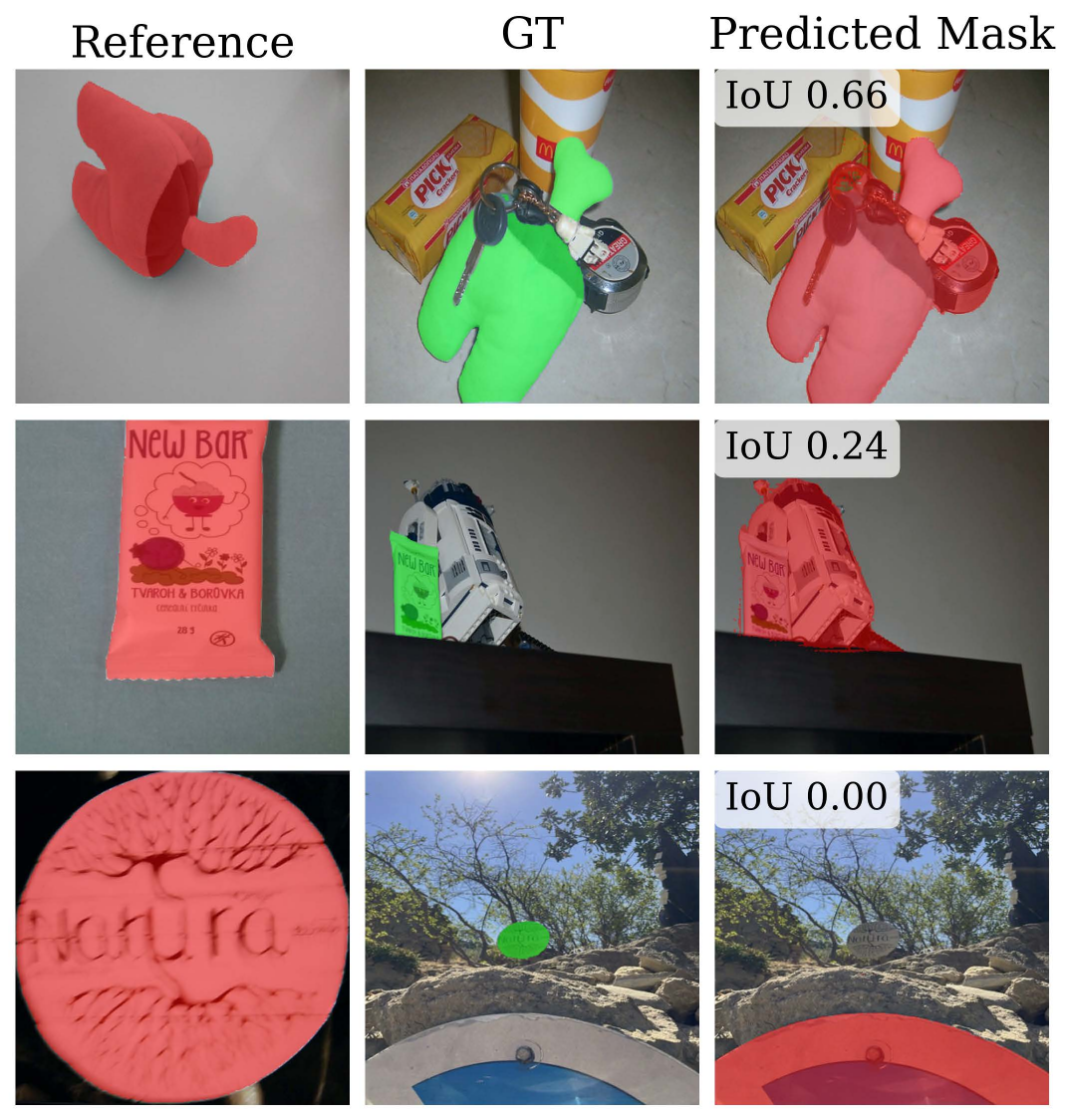}
\caption{\textbf{Failure cases for Personalized Segmentation.} 
From left to right: reference image with the mask indicating the target instance, ground-truth mask on the target image, and our prediction. 
Failure cases occur when the model either produces an inaccurate segmentation mask (\eg, overshooting the object boundaries) or fails to localize the instance entirely.}
\label{fig:failure_seg}
\end{minipage}%
\hspace{0.04\linewidth}
\begin{minipage}[t]{0.53\linewidth}
\centering
\includegraphics[width=\linewidth]{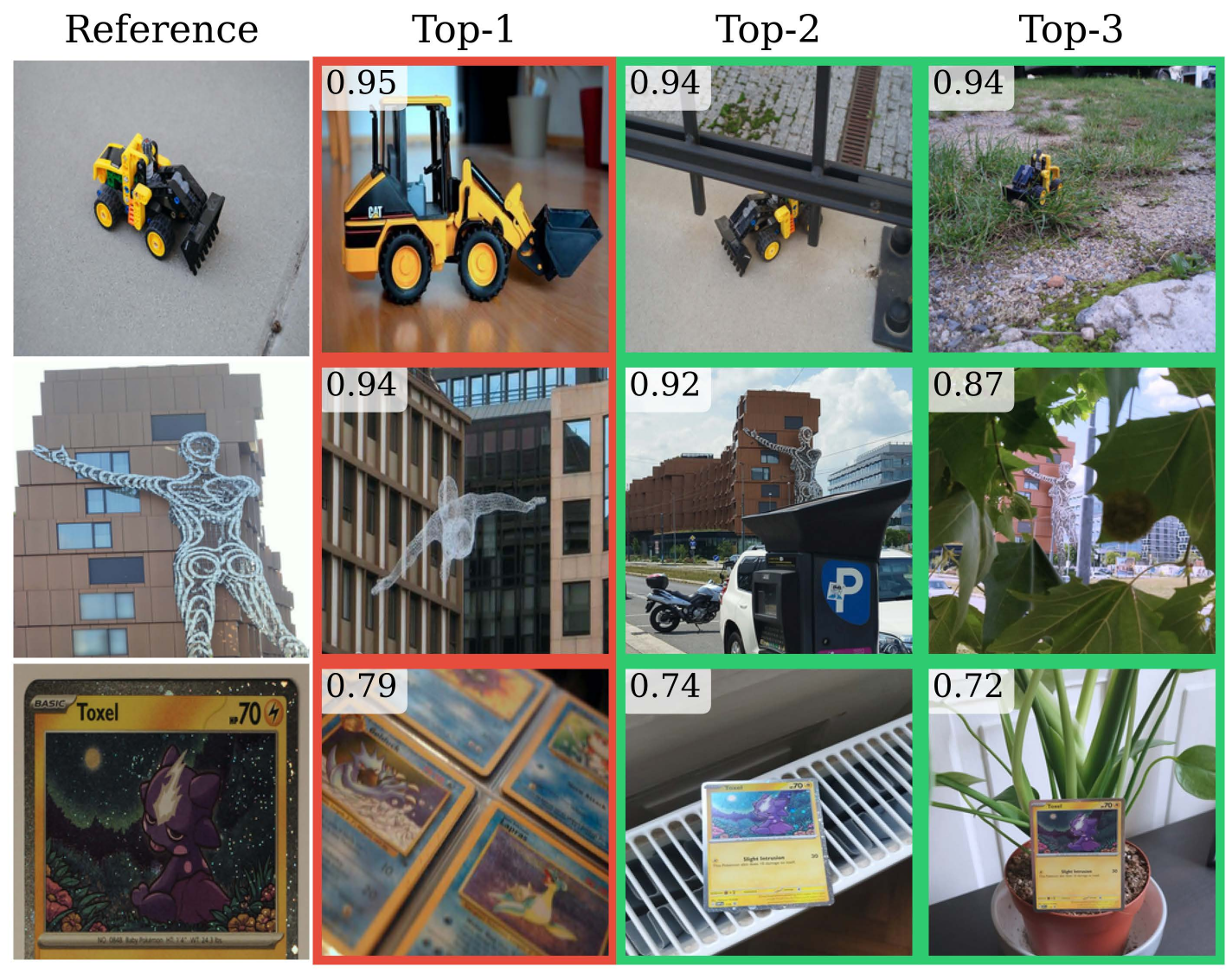}
\caption{\textbf{Failure cases for Personalized Retrieval.} 
The first column shows the reference instance. The following columns display the top-1, top-2, and top-3 retrieved images returned by \ours{}, together with their retrieval scores. 
In these examples, the top-1 result is an incorrect retrieval corresponding to a visually very similar distractor, illustrating the difficulty of instance-level discrimination. The correct positive images are instead ranked at positions two and three.}
\label{fig:failure_ret}
\end{minipage}

\end{figure*}

While our approach achieves strong performance on both personalized segmentation and personalized retrieval, challenging scenarios still lead to failure cases. ~\Cref{fig:failure_seg} and~\Cref{fig:failure_ret} present representative examples.

For \textbf{personalized segmentation}, errors mainly arise when the model struggles to accurately localize or precisely delineate the target instance in the target image. In some cases, the predicted mask extends beyond the object boundaries and includes surrounding regions with similar visual appearance, leading to over-segmentation. In other cases, the model fails to correctly identify the instance altogether and predicts an incorrect region.

For \textbf{personalized retrieval}, the main challenge arises from the presence of visually similar distractor instances in the retrieval database. As illustrated in ~\cref{fig:failure_ret}, the model may assign the highest similarity score to an incorrect instance that closely resembles the reference object in shape, color, or texture. In several cases, the correct instance still appears among the top retrieved results but is ranked slightly lower (\eg, second or third), suggesting that the model captures the relevant features but is not always able to fully distinguish between highly similar candidates. These examples highlight the inherent difficulty of instance-level retrieval in scenarios where subtle visual differences separate correct matches from distractors.

\section{Dataset details}
\label{sec:supp_data}

\Cref{tab:datasets} summarizes the datasets used in our evaluation, including their domain, query modality, and scale.

\myparagraph{Personalized Segmentation.}
We evaluate on \textit{PerSeg} and \textit{PerMIS}. \textit{PerSeg}, introduced with PerSAM~\cite{Zhang:2023:PerSAM}, contains 40 personal objects and 216 annotated images used to form query–target segmentation pairs. Visual prompts are defined by a reference image with a mask specifying the target instance.
\textit{PerMIS}, introduced in PDM~\cite{Samuel:2024:Waldo}, is constructed from 150 video sequences from BURST~\cite{Athar:2023:Burst} with dense instance masks for all frames. From these sequences, the benchmark extracts 216 reference queries and 432 target images containing the same instance. Compared to PerSeg, PerMIS is more challenging because scenes frequently contain multiple instances of the same category.

\myparagraph{Personalized Retrieval.}
We evaluate on \textit{PerMIR} and \textit{ILIAS}. \textit{PerMIR}~\cite{Samuel:2024:Waldo} provides a small-scale personalized retrieval benchmark with 216 queries and a gallery of 432 images, where each query is defined by a reference image and bounding box specifying the instance of interest.
\textit{ILIAS}~\cite{Kordopatis:2025:ILIAS} evaluates retrieval at significantly larger scale, with 1,232 query instances spanning multiple domains (\eg, fashion, products, artworks, and landmarks) and a gallery containing 4,715 positives and 100M distractor images sampled from YFCC100M~\cite{Thomee:2016:YFCC100M}. Queries are specified by an image and bounding box identifying the instance to retrieve.

To evaluate \textbf{generalization beyond strict instance matching}, we additionally report results on established retrieval benchmarks. These include fine-grained category-level datasets such as \textit{Cars196} (1,000 queries, 1.7K gallery images) and \textit{iNaturalist} (1,000 queries, 51K gallery images), as well as product retrieval datasets such as \textit{RP2K} (1,000 queries, 17K gallery images) and \textit{Stanford Online Products (SOP)} (1,000 queries, 10K gallery images). We include \textit{Food2K} (1,000 queries, 49K gallery images) for fine-grained food retrieval. Finally, we evaluate on instance-level recognition benchmarks including \textit{MET} (129 queries, 38K gallery images) and landmark retrieval datasets \textit{GLDv2}, \textit{RParis+1M}, and \textit{ROxford+1M}, which contain up to one million distractor images.

\end{document}